\documentclass[10pt,twocolumn,letterpaper]{article}

\usepackage{cvpr}
\usepackage[utf8]{inputenc}
\usepackage{times}
\usepackage{epsfig}
\usepackage{graphicx}
\usepackage{amsmath}
\usepackage{amssymb}

\makeatletter
\@namedef{ver@everyshi.sty}{}
\makeatother

\usepackage{tabularx}
\usepackage{multirow} %
\usepackage{booktabs} %
\usepackage{adjustbox}
\usepackage{placeins}
\usepackage{afterpage}

\usepackage{caption}
\usepackage{subcaption}
\usepackage{stmaryrd}
\usepackage[ruled,vlined,linesnumbered,procnumbered,longend]{algorithm2e}
\usepackage{enumitem}
\setitemize{itemsep=0pt,topsep=0pt,leftmargin=*}
\usepackage[numbers,square,sort]{natbib}
\usepackage{chngcntr}
\usepackage{multirow}

\usepackage{microtype} %

\usepackage{tikz}
\usepackage{float}
\usetikzlibrary{calc} %
\usetikzlibrary{arrows} %
\usetikzlibrary{positioning} %
\usetikzlibrary{fit} %
\usepackage{mathrsfs}

\newcommand{\bpl}{BP-Layer\xspace} 
\newcommand{\bpls}{BP-Layers\xspace} 

\usepackage{xcolor}
\usepackage{mycommands}

\usepackage[pagebackref=false,breaklinks=true,letterpaper=true,colorlinks,bookmarks=false]{hyperref}

\newcommand{\Real}{\mathbb{R}}

\newcommand{\V}{\mathcal{V}}

\renewcommand{\L}{\mathcal{L}}
\newcommand{\E}{\mathcal{E}}

\newcommand{\G}{\mathcal{G}}
\newcommand{\N}{\mathcal{N}}

\def\mathrlap{\mathpalette\mathrlapinternal} 
\def\mathllap{\mathpalette\mathllapinternal}
\def\mathllapinternal#1#2{\llap{$\mathsurround=0pt#1{#2}$}}
\def\mathrlapinternal#1#2{\rlap{$\mathsurround=0pt#1{#2}$}}

\newcommand{\revisit}[1][]{%
\ifthenelse{\equal{#1}{}}{%
\ensuremath{\red \triangle}\xspace}{%
{\ensuremath{\red \rhd}\xspace}%
{\gray #1}%
{\ensuremath{\red \lhd}\xspace}%
}%
}

\def\anchor [#1]#2{%
\phantomsection{}#1\label{#2}%
\def\arga{#2}%
\global\expandafter\def\csname#2\endcsname{%
\hyperref[#2]{#1}\xspace%
}%
}%

\def\codefunction [#1]#2{%
\phantomsection{}\label{#2}{\ttfamily #1\xspace}%
\def\arga{#2}%
\global\expandafter\def\csname#2\endcsname{%
\hyperref[#2]{\ttfamily #1}\xspace%
}%
}

\newcommand{\dbar}{d\hspace*{-0.08em}\bar{}\hspace*{0.1em}}
\newcommand*{\dv}[1]{\dbar #1}

\newcommand{\gray}{\color[rgb]{0.5,0.5,0.5}}
\newcommand{\red}{\color[rgb]{1,0,0}}

\DeclareMathOperator*{\smax}{{\widetilde{\max}}}
\DeclareMathOperator*{\softmax}{softmax}
\DeclareMathOperator*{\argmax}{argmax}

\SetAlgoSkip{2pt}
\SetAlgoInsideSkip{2pt}

\makeatletter
\newcommand{\removelatexerror}{\let\@latex@error\@gobble}
\makeatother

\usepackage[capitalise]{cleveref} %

\cvprfinalcopy %

\def\cvprPaperID{5455} %

\ifcvprfinal\fi
\addtocontents{toc}{\protect\setcounter{tocdepth}{-1}}
\begin{document}

\title{Belief Propagation Reloaded: Learning BP-Layers for Labeling Problems}

\author{Patrick Knöbelreiter$^1$\\
{\tt\small knoebelreiter@icg.tugraz.at}%
\and
Christian Sormann$^1$\\
{\tt\small christian.sormann@icg.tugraz.at}
\and
Alexander Shekhovtsov$^2$\\
{\tt\small shekhovtsov@gmail.com}
\and
Friedrich Fraundorfer$^1$\\
{\tt\small fraundorfer@icg.tugraz.at}
\and
Thomas Pock$^1$\\
{\tt\small pock@icg.tugraz.at}
\and
$^1$ 
Graz University of Technology \\
$^2$Czech Technical University in Prague
}

\maketitle

\begin{abstract}
    It has been proposed by many researchers that combining deep neural networks with graphical models can create more efficient and better regularized composite models. The main difficulties in implementing this in practice are associated with a discrepancy in suitable learning objectives as well as with the necessity of approximations for the inference. %
In this work we take one of the simplest inference methods, a truncated max-product Belief Propagation, and add what is necessary to make it a proper component of a deep learning model: We connect it to learning formulations with losses on marginals and compute the backprop operation. This \bpl can be used as the final or an intermediate block in convolutional neural networks (CNNs), allowing us to design a hierarchical model composing BP inference and CNNs at different scale levels. The model is applicable to a range of dense prediction problems, is well-trainable and provides parameter-efficient and robust solutions in stereo, optical flow and semantic segmentation.

\end{abstract}

\section{Introduction}
\label{sec:intro}

\begin{figure}
    \centering
    \includegraphics[width=0.95\columnwidth]{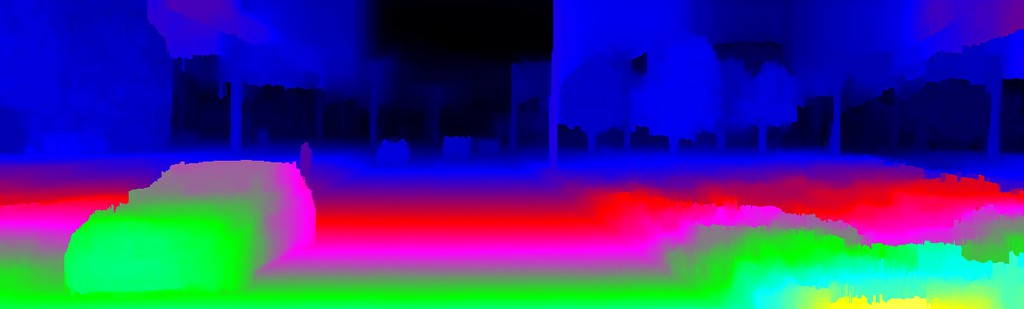}
    \includegraphics[width=0.95\columnwidth, clip, trim={0 130 0 70}]{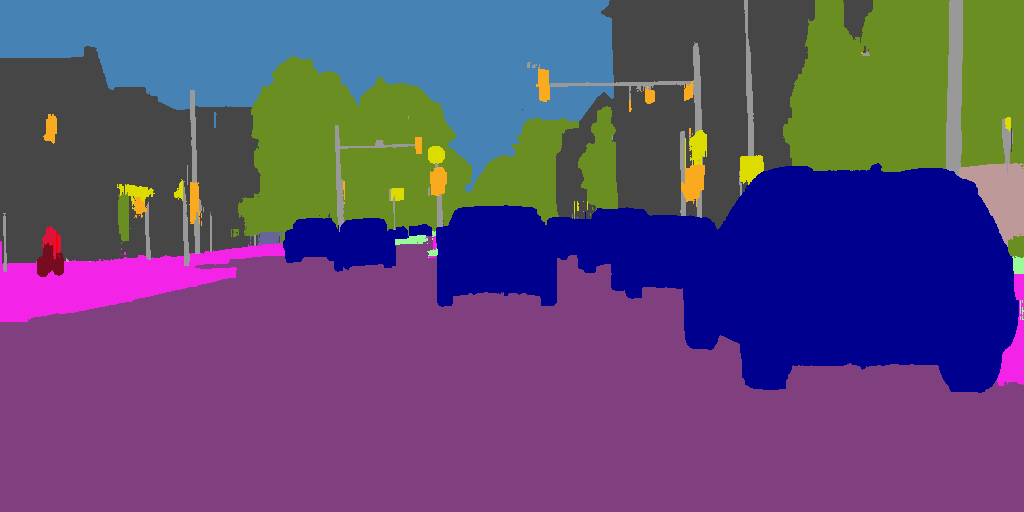}
    \includegraphics[width=0.95\columnwidth, clip, trim={100, 100, 0, 60}]{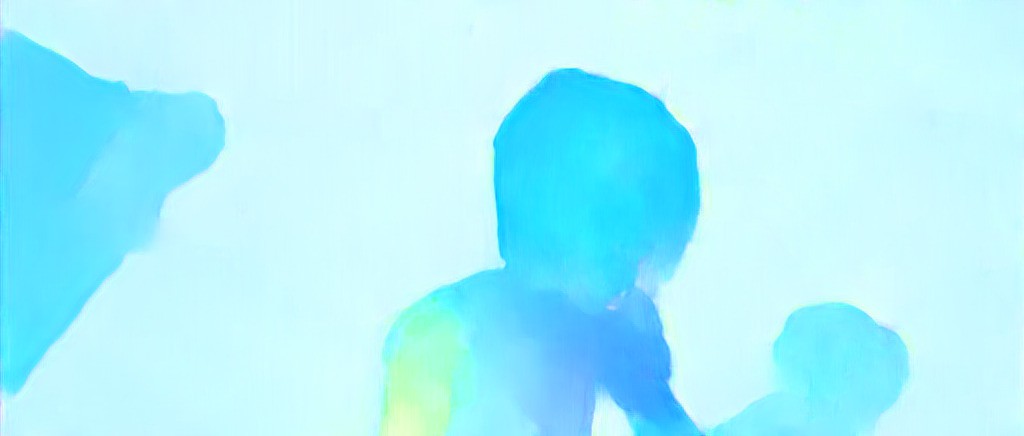}
    \caption{\bpl in action. The \bpl can be used for dense prediction problems such as stereo (top) semantic segmentation (middle) or optical flow (bottom). Note the sharp and precise edges for all three tasks. Input images are from Kitti, Cityscapes and Sintel benchmarks.}
    \label{fig:my_label}
\end{figure}
We consider dense prediction tasks in computer vision that can be formulated as assigning a categorical or real value to every pixel. Of particular interest are the problems of semantic segmentation, stereo depth reconstruction and optical flow. The importance of these applications is indicated by the active development of new methods and intense competition on common benchmarks.

Convolutional Neural Networks (CNNs) have significantly pushed the limits in dense prediction tasks. However, composing only CNN blocks, though a general solution, becomes inefficient if we want to increase robustness and accuracy: with the increase of the number of blocks and respectively parameters the computational complexity and the training data required grow significantly. The limitations are in particular in handling long-range spatial interactions and structural constraints, for which Conditional Random Fields (CRFs) are much more suitable. Previous work has shown that a combination of CNN+CRF models can offer an increased performance, but incorporating inference in the stochastic gradient training poses some difficulties. %

In this work we consider several simple inference methods for CRFs: A variant of Belief Propagation (BP)~\cite{tappen03}, tree-structured dynamic programming~\cite{Bleyer08simplebut} and semi-global matching~\cite{Hirschmuller:2008}. %
We introduce a general framework, where we view all these methods as specific schedules of max-product BP updates and propose how to use such BP inference as a layer in neural networks fully compatible with deep learning.
The layer takes categorical probabilities on the input and produces refined categorical probabilities on the output, associated with marginals of the CRF. 
This allows for direct training of the truncated inference method by propagating gradients through the layer. 
The proposed \bpl can have an associated loss function on its output probabilities, which we argue to be more practical than other variants of CRF training. %
Importantly, it can be also used as an inner layer of the network. %
We propose a multi-resolution model in which \bpls are combined in a hierarchical fashion and feature both, associated loss functions as well as dependent further processing blocks. %

We demonstrate the effectiveness of our \bpl on three dense prediction tasks. The \bpl performs a global spatial integration of the information on the pixel-level and is able to accurately preserve object boundaries as highlighted in~\cref{fig:my_label}. Deep models with this layer have the following beneficial properties: (i) they contain much fewer parameters, (ii) have a smaller computation cost than the SoTA fully CNN alternatives, (iii) they are better interpretable (for example we can visualize and interpret CRF pairwise interaction costs) and (iv) lead to robust accuracy rates. In particular, in the high-resolution stereo Middlebury benchmark, amongst the models that run in less than 10 seconds, our model achieves the second best accuracy. The CRF for stereo is particularly efficient in handling occlusions, explicitly favoring slanted surfaces and in modelling a variable disparity range. In contrast, many CNN techniques have the disparity range hard-coded in the architecture.

\subsection*{Related Work}\label{sec:relatedWork}
We discuss the related work from the points of view of the learning formulation, gradient computation and application in dense prediction tasks.

{\bf CRF Learning}
CRFs can be learned by the maximum margin approach (\eg, \cite{Keshet-14,Knobelreiter_2017_CVPR}) or the maximum likelihood approach and its variants~(\eg, \cite{KirillovSFZ0TR15,Alahari10a,Christopher-12-learnig,LinSRH15}). In the former, the loss depends on the optimal (discrete) solution and is hard to optimize. In the latter, the gradient of the likelihood is expressed via marginals and approximate marginals can be used. However, it must be ensured that during learning enough iterations are performed, close to convergence of the approximation scheme~\cite{Domke2011ParameterLW}, which is prohibitive in large-scale learning settings. Instead, several works advocate truncated inference and a loss function directly formulated on the approximate marginals~\cite{Kakade02analternate,Domke2011ParameterLW,Domke-learning}. This gives a tighter connection between learning and inference, is better corresponding to the empirical loss minimization with the Hamming loss and is easy to apply with incomplete ground truth labelings.
Experimental comparison of multiple learning approaches for CRFs~\cite{Domke-learning} suggest that marginalization-based learning performs better than likelihood-based approximations on difficult problems where the model being fit is approximate in nature. Our framework follows this approach.

{\bf Differentiable CRF Inference}
For learning with losses on marginals~\citet{Domke-learning} introduced Back-Mean Field and Back-TRW algorithms allowing back-propagation in the respective inference methods. 
Back-Belief Propagation~\cite{eaton2009choosing} is an efficient method applicable at a fixed point of BP, originally applied in order to improve the quality of inference, and not suitable for truncated inference.
While the methods~\cite{eaton2009choosing,Domke-learning,Domke2011ParameterLW} consider the sum-product algorithms and back-propagate their elementary message passing updates, our method back-propagates the sequence of max-product BP updates on a chain at once.  %
Max-product BP is closely related with the Viterbi algorithm and Dynamic Programming (DP). However, DP is primarily concerned with finding the optimal configuration. The smoothing technique~\cite{mensch2018differentiable} addresses differentiating the optimal solution itself and its cost. In difference, we show the back propagation of max-marginals. %

The {\em mean field inference} in fully connected CRFs 
for semantic segmentation~\cite{chen14semantic,zheng2015conditional} like our method maps label probabilities to label probabilities, is well-trainable and gives improvements in semantic segmentation. 
However, the model does not capture accurate boundaries~\cite{Marin_2019_CVPR} and cannot express constraints needed for stereo/flow such as non-symmetric and anisotropic context dependent potentials.

{\em Gaussian CRFs} (GCRFs) use quadratic costs, which is restrictive and not robust if the solution is represented by one variable per pixel. If $K$ variables are used per pixel~\cite{vemulapalli2016gaussianCRF}, a solution of a linear system of size $K \times K$ is needed per each pairwise update and the propagation range is only proportional to the number of iterations.

{\em Semi-Global Matching} (SGM)~\cite{Hirschmuller:2008} is a very popular technique adopted by many works on stereo due to its simplicity and effectiveness. However, its training has been limited either to learning only a few global parameters~\cite{mensch2018differentiable} or to indirect training via auxiliary loss functions~\cite{Seki-2017-CVPR} avoiding backpropagating SGM.
Although we focus on a different inference method, our framework allows for a simple implementation of SGM and its end-to-end learning.

{\bf Non-CRF Propagation} Many methods train continuous optimization algorithms used inside neural networks by unrolling their iterations~\cite{riegler2016atgv,Knobelreiter_2019_GCPR,vogel2018learning}. Spatial propagation networks~\cite{Liu-17-SPN}, their convolutional variant~\cite{Xinjing-18-CSPN} and guided propagation~\cite{zhang2019ga} apply linear spatial propagation models in particular in stereo reconstruction. 
In difference, we train an inference algorithm that applies non-linear spatial propagation.
From this point of view it becomes related to recurrent non-linear processing methods such PixelCNN~\cite{VanDenOord-16}.

\section{Belief Propagation}
\label{sec:lbp}

In this section we give an overview of sum-product and max-product belief propagation (BP) algorithms and argue that max-marginals can be viewed as approximation to marginals. This allows to connect learning with losses on marginals~\cite{Domke-learning} and the max-product inference in a non-standard way, where the output is not simply the approximate MAP solution, but the whole volume of max-marginals.

Let $\G  = (\V, \E)$ be an undirected graph and $\L$ a discrete set of labels. 
A pairwise Markov Random Field (MRF)~\cite{Lauritzen96} over $\G$ with state space $\V^\L$ is a probabilistic graphical model $p\colon \V^\L \to \Real_+$ that can be written in the form
\begin{align}\label{eq:mrfEnergy}
    p(x) = \frac{1}{Z} \exp
    \Big(
    \sum_{i \in \mathcal{V}} g_i(x_i) + \sum_{(i,j) \in \mathcal{E}} f_{ij}(x_i, x_j)
    \Big),
\end{align}
where $Z$ is the normalization constant, functions $g_i \colon \L \rightarrow \R$ are the {\em unary scores}\footnote{The negative scores are called {\em costs} in the context of minimization.}, typically containing data evidence; and functions $f_{ij} \colon \mathcal{L}^2 \rightarrow \R$ are {\em pairwise scores} measuring the compatibility of labels at nodes $i$ and $j$.
A CRF $p(x|y)$ is a MRF model~\eqref{eq:mrfEnergy} with scores depending on the inputs $y$.

Belief Propagation~\cite{Pearl:1982} was proposed to compute marginal probabilities of a MRF~\eqref{eq:mrfEnergy} when the graph $\G$ is a tree. 
BP iteratively sends {\em messages} $M_{ij} \in \Real^L_+$ from node $i$ to node $j$ with the update:
\begin{align}\label{eq:messageUpdates}
    M_{ij}^{k+1}(t) \propto \sum_s
    e^{g_i(s)} e^{f_{ij}(s, t)} \prod_{n \in \mathcal{N}(i)\setminus j} M_{ni}^{k}(s),
\end{align}
where $\N(i)$ is the set of neighboring nodes of a node $i$ and $k$ is the iteration number. %
In a tree graph a message $M_{ij}$ is proportional to the marginal probability that a configuration of a tree branch ending with $(i,j)$ selects label $t$ at $j$. %
Updates of all messages are iterated until the messages have converged. Then the marginals, or in a general graph {\em beliefs}, are defined as 
\begin{align}
B_i(x_i) \propto e^{g_i(x_i)} \prod_{n \in \mathcal{N}(i)} M_{ni}(x_i),
\label{eq:beliefReadout}
\end{align}
where the proportionality constant ensures $\sum_s B_i(s) = 1$. 

The above {\em sum-product} variant of BP can be restated in the log domain, where the connection to max-product BP becomes apparent.
We denote $\smax$ the operation $\Real^n \to \Real$ that maps $(a_1, \dots a_n)$ to $\log \sum_{i} e^{a_i}$, known as log-sum-exp or {\em smooth maximum}.
The update of the sum-product BP~\eqref{eq:messageUpdates} can be expressed as
\begin{align}\label{eq:messageUpdates-log}
    m_{ij}^{k+1}(t) := \smax_{s}
    \Big(g_i(s) + f_{ij}(s, t) + \!\!\!\!\!\! \sum_{n \in \mathcal{N}(i)\setminus j} \!\!\!\!\! m_{ni}^{k}(s)\Big),
\end{align}
where $m$ are the log domain messages, defined up to an additive constant. The {\em log-beliefs} are respectively
\begin{align}\label{log-beliefs}
b_i(x_i) = g_i(x_i) + \sum_{n \in \mathcal{N}(i)} m_{ni}(x_i).
\end{align}
The {\em max-product BP} in the log domain takes the same form as~\eqref{eq:messageUpdates-log} but with the hard $\max$ operation. 
Max-product solves the problem of finding the configuration $x$ of the maximum probability (MAP solution) and computes {\em max-marginals} via~\eqref{log-beliefs}. It can be viewed as an approximation to the marginals problem since there holds
\begin{align}\label{lse-max-ineq}
\max_i a_i \leq \smax_i a_i \leq \max_i a_i + \log n
\end{align}
for any tuple $(a_1 \dots a_n)$. Preceding work has noticed that max-marginals can in practice be used to assess uncertainty~\cite{Kohli:2006}, \ie, they can be viewed as approximation to marginals. The perturb and MAP technique~\cite{PaYu14} makes the relation even more precise.
In this work we apply max-marginal approximation to marginals as a practical and fast inference method for both, prediction time and learning. We rely on deep learning to make up for the approximation. In particular the learning can tighten~\eqref{lse-max-ineq} by scaling up all the inputs. 

To summarize, the approximation to marginals that we construct is obtained by running the updates~\eqref{eq:messageUpdates-log} with hard $\max$ and then computing beliefs from log-beliefs~\eqref{log-beliefs} as
\begin{align}\label{eq:beliefs}
B_i(x_i{=}s) = \softmax_{s} b_i(s),
\end{align}
where $\softmax_s b_i(s) = e^{b_i(s)}/\sum_{s} e^{b_i(s)}$. %
Beliefs constructed in this way may be used in the loss functions on the marginal or as an input to subsequent layers, similarly to how simple logistic regression models are composed to form a sigmoid neural network. This approach is akin to previous work that used the regularized cost volume in a subsequent refinement step~\cite{khamis2018stereonet}, but is better interpretable and learnable with our methods.

\section{Sweep \bpl}
\begin{figure}[t!]
\centering
\scalebox{0.7}{
\begin{tikzpicture}
\begin{scope}[ultra thick]
    \foreach \x in {0,...,4} {
        \foreach \y in {0,...,4} {
            \node[draw, circle, fill=gray!50] (\x\y) at (\x, \y) {};
            \node[draw, circle, xshift=5.5cm] (\x\y) at (\x, \y) {};
        }
    }
    
    \begin{scope}[shorten <= 7pt, shorten >= 7pt,->]
    \foreach \y in {0,...,4} {
        \foreach \x in {0,...,1} {
            \draw[] (\x, \y) -- (\x+1, \y);
        }
        \foreach \x in {2,...,3} {
            \draw[] (\x+1, \y) -- (\x, \y);
        }
    }

    \foreach \y in {0,...,1} {
        \draw[] (2, \y) -- (2, \y + 1);
        \draw[xshift=5.5cm] (2, \y) -- (2, \y + 1);
        
        \draw[xshift=5.5cm] (\y, 2) -- (\y + 1, 2);
        
        \node[draw, circle, fill=gray!50, xshift=5.5cm] at (2, \y) {};
        \node[draw, circle, fill=gray!50, xshift=5.5cm] at (\y, 2) {};
    }
    \foreach \y in {2,...,3} {
        \draw[] (2, \y + 1) -- (2, \y);
        \draw[xshift=5.5cm] (2, \y + 1) -- (2, \y);
        
        \draw[xshift=5.5cm] (\y + 1, 2) -- (\y, 2);
        
        \node[draw, circle, fill=gray!50, xshift=5.5cm] at (2, \y + 1) {};
        \node[draw, circle, fill=gray!50, xshift=5.5cm] at (\y + 1, 2) {};
    }
    \end{scope}
    
    \begin{scope}[minimum size=0.0cm, inner sep=2pt]
        \node[draw, circle, minimum size=0.0cm, fill=gray!80] at (2, 2) {$p$};
        \node[draw, circle, minimum size=0.0cm, fill=gray!80, xshift=5.5cm] at (2, 2) {$p$};
    \end{scope}
    
    \begin{scope}[minimum size=4.8cm, thick]
        \node[draw] at (2,2) {};
        \node[draw, xshift=5.5cm] at (2,2) {};
    \end{scope}
\end{scope}
\end{tikzpicture}
}
\caption{Max-marginal computation for node $p$ on the highlighted trees. {\em Left}: Left-right-up-down BP~\cite{tappen03} or equivalent tree DP~\cite{Bleyer08simplebut}. {\em Right}: SGM~\cite{Hirschmuller:2008} on a 4-connected graph. Note that SGM prediction for node $p$ uses much smaller trees, ignoring the evidence from out of tree nodes.%
\label{fig:tree-DP}}
\end{figure}
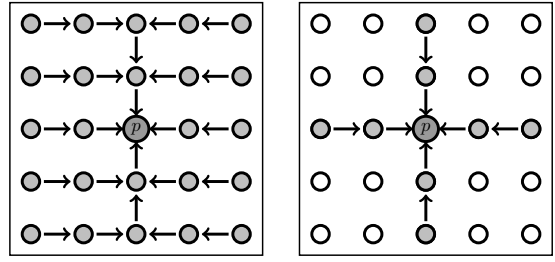

When BP is applied in general graphs, the schedule of updates becomes important. We find that the parallel synchronous update schedule~\cite{Pearl:1988} requires too many iterations to propagate information over the image and rarely converges. For application in deep learning, we found that the schedule which makes sequential sweeps in different directions as proposed by~\cite{tappen03} is more suitable. 
For a given sweep direction, we can compute the result of all sequential updates and backpropagate the gradient in a very efficient and parallel way. This allows to propagate information arbitrarily far in the sweep direction, while working on a pixel level, which makes this schedule very powerful.

Before detailing the sweep variant of BP~\cite{tappen03}, let us make clear what is needed in order to make an operation a part of an end-to-end learning framework. Let us denote the gradient of a loss function $L$ in variables $y$ as $\dv{y} := \frac{{\rm d} L}{{\rm d y}}$.
If a layer computed $y = f(x)$ in the forward pass, the gradient in $x$ is obtained as
\begin{align}
\textstyle
\dv{x}_j = \sum_{i}\frac{\partial f_i}{\partial x_j} \dv{y}_i,
\end{align}
called the {\em backprop} of layer $f$. %
For the BP-Layer the input probabilities $x$ and output beliefs $y$ are big arrays containing all pixels and all labels. It is therefore crucial to be able to compute the backprop in linear time.
\subsection{Sweep BP as Dynamic Programming}\label{sec:bpLayer}
\begin{figure}[t]
\centering
\vbox to .849\textheight{%
\begingroup
\removelatexerror%
\begin{algorithm*}[H]
\KwIn{CRF scores $g\in \Real^{\V \times \L}$, $f\in \Real^{\E \times \L^2}$\;}
\KwOut{Beliefs $B \in \Real^{\V\times \L}$\;}
\ForP{each horizontal chain subgraph $(\V', \E')$}{
	$a_{\V'} := \maxmarginals(g_{\V'}, f_{\E'})$\;
}
\ForP{each vertical chain subgraph $(\V', \E')$}{
	$b_{\V'} := \maxmarginals(a_{\V'}, f_{\E'})$\;
}
\Return beliefs $B_i(s) := \softmax_s(b_i(s))$\;
\caption{Sweep Belief Propagation\label{alg:BP}}
\end{algorithm*}
\endgroup
\vfill
\begingroup
\removelatexerror%
\begin{algorithm*}[H]
\KwIn{Directed chain $(\V,\E)$, nodes $\V$ enumerated in chain direction from $0$ to $n{=}|\V|{-}1$, \mbox{scores $g \in \Real^{\V{\times}\L}$, $f \in \Real^{\E{\times}\L^2}$}\;}
\KwOut{Messages $m \in \Real^{\V\times \L}$ in chain direction\;}
{\bf Init:} Set: $m_0(s) := 0$\tcc*{first node}
\For{$i=0 \dots n - 2$}{
	\tcc{Compute message:}
	$ m_{i+1}(t) := \max\limits_{s}\big(g_{i}(s) + m_{i}(s) + f_{i, i+1}(s,t)\big)$\;
	\tcc{Save argmax for backward:}
    ${o_{i{+}1}(t) := \argmax\limits_{s}\big(g_{i}(s) + m_{i}(s) + f_{i, i+1}(s,t)\big)}$;
    }
\Return m\;
\caption{Dynamic Programming (DP)\label{alg:DP}}
\end{algorithm*}
\endgroup
\vfill
\ \ \ \includegraphics[width=0.95\linewidth]{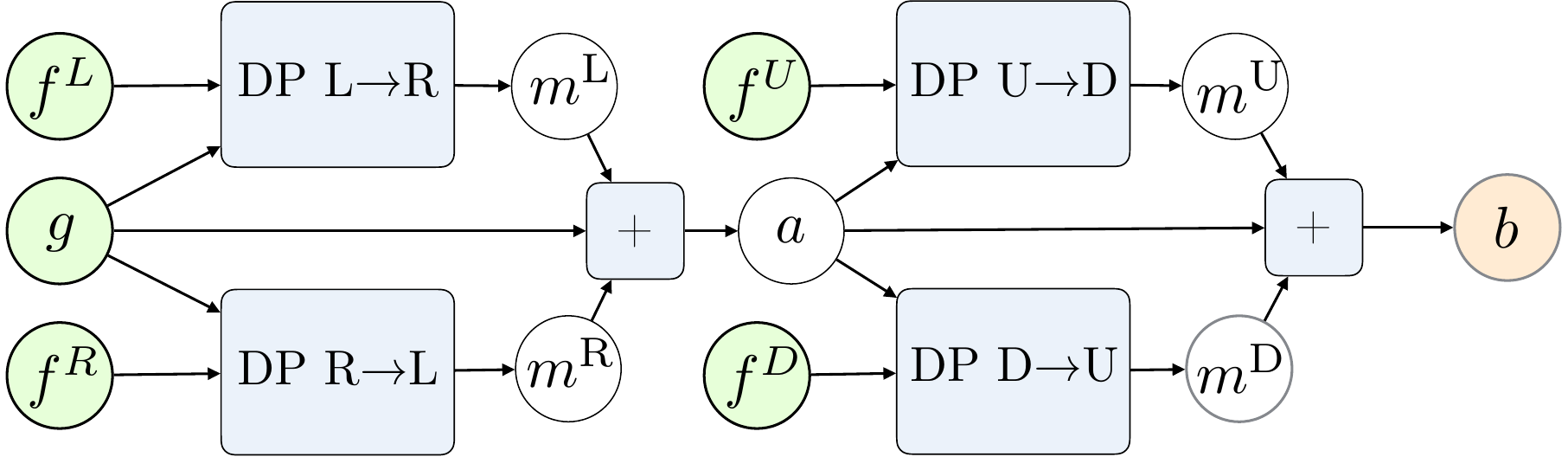}
\caption{Computation graph of BP-Layer with Sweep BP in~\cref{alg:BP} down to log-beliefs $b$. Dynamic Programming computational nodes (DP) are made differentiable with the backprop in~\cref{alg:back-DP}.
The pairwise terms $f^{\rm L}$,
$f^{\rm R}$, $f^{\rm U}$, $f^{\rm D}$ illustrate the case when pairwise scores $f_{ij}$ are different for all four directions.
\label{fig:BPL-graph}}
\vfill
\begingroup
\removelatexerror%
\begin{algorithm*}[H]
	\KwIn{$\dv{m} \in \Real^{\V \times \L}$, gradient of the loss in the messages $m$ returned by DP on chain $(\V,\E)$\;}
  \KwOut{$\dv{g} \in \Real^{\V \times \L}$, $\dv{f} \in \Real^{\E \times \L^2}$, gradients of the loss in the DP inputs $g$, $f$\;}
	{\bf Init:} $\dv{g} :=0;$ $\dv{f} := 0$\;
		\For{$i=n-2 \dots 0$}{
			\For{$t \in \L$}{
				$s := o_{i+1}(t)$\;
				$z := \dv{m}_{i+1}(t) + \dv{g}_{i+1}(t)$\label{back-DP:line:z}\;
				$\dv{g}_i(s) \: {+}{=}\: z$\;
				$\dv{f}_{i,i+1}(s,t) \: {+}{=}\: z$\;
			}
		}
		\Return $\dv{g}$, $\dv{f}$\;
\caption{Backprop DP\label{alg:back-DP}}
\end{algorithm*}
\endgroup
}
\end{figure}

\SetKwFunction{maxmarginals}{{\ttfamily max\_marginals}}
\SetKwFor{ForP}{par. for}{do}{end}
The BP variant of~\cite{tappen03} (called left-right-up-down BP there and BP-M in~\cite{Szeliski:2008}) performs sweeps in directions left$\rightarrow$right, right$\rightarrow$left, up$\rightarrow$down, down$\rightarrow$up.
For each direction only messages in that direction are updated sequentially, and the rest is kept unmodified.
We observe the following properties of this sweep BP:
(i) Left and right messages do not depend on each other and neither on the up and down messages. Therefore, their calculation can run independently in all horizontal chains.
(ii) When left-right messages are fixed, they can be combined into unary scores, which makes it possible to compute the up and down messages independently in all vertical chains in a similar manner.
These properties allow us to express left-right-up-down BP as shown in~\cref{alg:BP} and illustrated in~\cref{fig:tree-DP}~(left). %
In~\cref{alg:BP}, the notation $a_{\V'}$ means the restriction of $a$ to the nodes in $\V'$, \ie to a chain.
It is composed of dynamic programming subroutines computing max-marginals. Since individual chains in each of the loops do not interact, they can be processed in parallel (denoted as par. for).
The max-marginals $a$ of a horizontal chain are computed as
\begin{align}\label{max-marginals-a}
\textstyle
a_i(s) = g_i(s) + m^L_i(s) + m^R_i(s),
\end{align}
where $m^L_i(s)$ denotes the message to $i$ from its left neighbour and $m^R_i(s)$ from its right. The max-marginals~\eqref{max-marginals-a} are indeed the beliefs after the left-right pass. The max-marginals $b$ for vertical chains are, respectively,
\begin{align}\label{max-marginals-b}
\textstyle
b_i(s) = a_i(s) + m^U_i(s) + m^D_i(s).
\end{align}
It remains to define how the messages $m$ are computed and back-propagated.
Given a chain and the processing direction (\ie, $\rm L$-$\rm R$ for left messages $m^{\rm L}$), we order the nodes ascending in this direction and apply dynamic programming in~\cref{alg:DP}.
The Jacobian of~\cref{alg:DP} is well defined if the maximizer in each step is unique\footnote{Otherwise we take any maximizer resulting in a conditional derivative like with ReLU at 0.}. In this case we have a linear recurrent dependence in the vicinity of the input:
\begin{align}
\textstyle
m_{i+1}(t) = g_{i}(s) + m_{i}(s) + f_{i, i+1}(s,t),
\end{align}
where $s = o_{i+1}(t)$, \ie the label maximizing the message, as defined in~\cref{alg:DP}. Back-propagating this linear dependence is similar to multiplying by the transposed matrix, \eg, for the gradient in $g_i(s)$ we need to accumulate over all elements to which $g_{i}(s)$ is contributing. This can be efficiently done as proposed in~\cref{alg:back-DP}. 

Thus we have completely defined sweep BP, further on referred to as {\em \bpl}, as a composition of differential operations. The computation graph of the BP-Layer shown in~\cref{fig:BPL-graph} can be back-propagated using standard rules and our Backprop DP in order to compute the gradients in all inputs very efficiently.

\subsection{Other Inference Methods}
We show the generality of the proposed framework by mapping several other inference techniques to the same simple DP operations. This allows to make them automatically differentiable and suitable for learning with marginal losses. %

{\bf SGM } %
We can implement SGM using the same DP function we needed for BP (\cref{alg:SGM}), where for brevity we considered a 4-connected grid graph. %
As discussed in the related work, the possibility to backpropagate SGM was previously missing and may be useful.
\begin{algorithm}[t]
\KwIn{CRF scores $g\in \Real^{\V \times \L}$, $f\in \Real^{\E \times \L^2}$\;}
\KwOut{Beliefs $b \in \Real^{\V \times \L}$\;}
\ForP{each direction $k$ in $\{\rm L, R, U, D\}$}{
 \ForP{each chain $(\V', \E')$ in direction to $k$}{
		$m^k_{\V'} := DP(g_{\V'}, f_{\E'})$\;
	}
}
\Return $b = g + \sum_{k} m^k$\;
\caption{Semi-Global Matching\label{alg:SGM}}
\end{algorithm}

\textbf{Tree-structured DP}~\citet{Bleyer08simplebut} proposed an improvement to SGM by extending the local tree as shown in~\cref{fig:tree-DP} (left), later used \eg in a very accurate stereo matching method~\cite{YANG2014}. It seems it has not been noticed before that sweep BP~\cite{tappen03} is exactly equivalent to the tree-structured DP of~\cite{Bleyer08simplebut}, as clearly seen from our presentation.

\textbf{TRW and TBCA }%
With minor modifications of the already defined DP subroutines, it is possible to implement and back-propagate several inference algorithms addressing the dual of the LP relaxation of the CRF: the Tree-Reweighted (TRW) algorithm by~\citet{wainwright2003tree} and Tree Block Coordinate Ascent (TBCA) by~\citet{sontag2009tree}, which we show in~\cref{sec:A}.
These algorithms are parallel, incorporate long-range interactions and avoid the evidence over-counting problems associated with loopy BP~\cite{wainwright2003tree}. In addition, the TBCA algorithm is monotone and has convergence guarantees. These methods are therefore good candidates for end-to-end learning, however they may require more iterations due to cautious monotone updates, which is undesirable in the applications we consider. %

\section{Models}
\begin{figure}[t]
    \centering

    \includegraphics[width=\columnwidth, clip, trim={175, 0, 173, 0}]{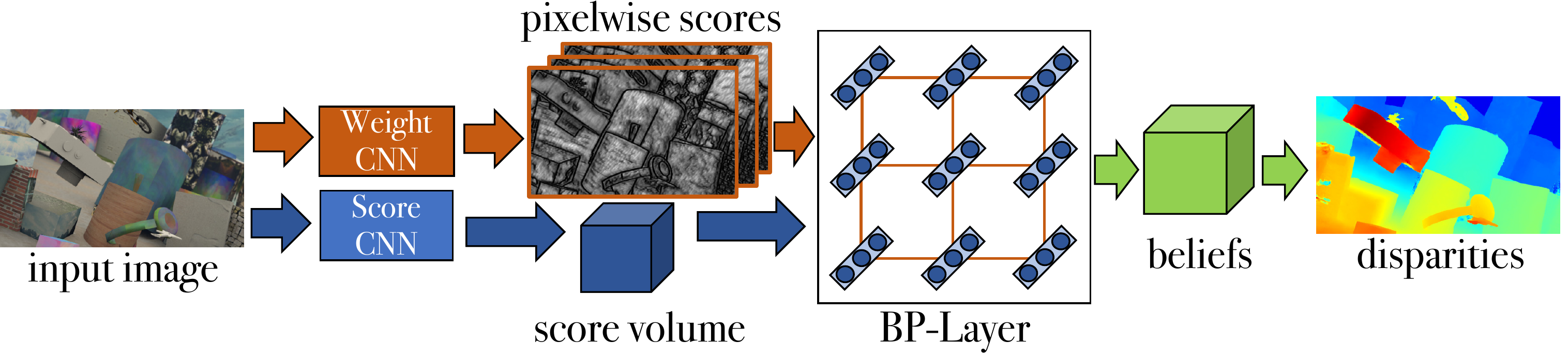}
    \caption{\bpl overview. The weight and score CNNs compute pixelwise weights and a score volume from the input image. This is used as an input for the \bpl which returns beliefs as an output.}
    \label{fig:modelOverview}
\end{figure}

We demonstrate the effectiveness of the \bpl on the three labeling problems: Stereo, Optical Flow and Semantic Segmentation.
We have two CNNs (\cref{tab:featurenet}) which are used to compute i) score-volumes and ii) pairwise jump-scores, at three resolution levels used hierarchically. \cref{fig:modelOverview} shows processing of one resolution level with the BP-Layer.
The label probabilities from these predictions are considered as weak classifiers and the inference block combines them to output a stronger finer-resolution classification.
Accordingly, the unary scores $g_i(s)$, called the {\em score volume}, are set from the CNN prediction probabilities $q_i(s)$ as
\begin{align}\label{unary-model}
g_i(s) = T q_i(s),
\end{align}
where $T$ is a learnable parameter. Note that $g_i$ is itself a linear parameter of the exponential model~\eqref{eq:mrfEnergy}. The preceding work more commonly used the model $g_i(s) = \log q_i(x)$, which, in the absence of interactions, recovers back the input probabilities. In contrast, the model~\eqref{unary-model} has the following interpretation and properties:
i) it can be viewed as just another non-linearity in the network, increasing flexibility;
ii) in case of stereo and flow it corresponds to a robust metric in the feature space (see below), in particular it is robust to CNN predictive probabilities being poorly calibrated.

To combine the up-sampled beliefs $B^{\rm up}$ from the coarser-resolution \bpl with a finer-resolution evidence $q$, 
we trilinearly upsample the beliefs from the lower level and add it to the score-volume of the current level, \ie
\begin{align}\label{eq:unary-model}
g_i(s) = T \big( q_i(s) + B^{\rm up}_i(s) \big).
\end{align}
On the output 
we have an optional refinement block, which is useful for predicting continuous values for  stereo and flow. 
The simplest refinement %
takes the average in a window around the maximum:
\begin{align}\label{basic-refine}
y = \sum_{d:\lvert d - \hat d_i \rvert \leq \tau} d \, B_i(d)
\Big(\sum_{d:\lvert d - \hat d_i \rvert \leq \tau} B_i(d)\Big)^{-1}
, %
\end{align}
where $\hat d_i = \argmax B_i(d)$ and we use the threshold $\tau = 3$. 
Such averaging is not affected by a multi-modal distribution, unlike the full average used in~\cite{kendall2017}. %
As a more advanced refinement block we use a variant of the refinement~\cite{khamis2018stereonet} with one up-sampling step using also the confidence of our prediction as an additional input.
\subsection{Stereo}\label{sec:modelStereo}
For the rectified stereo problem we use two instances of a variant of the UNet detailed in~\cref{sec:implementation}.
This network is relatively shallow and contains significantly fewer parameters than SoTA.
It is applied to the two input images $I^0$, $I^1$ and produces two dense feature maps $f^0$, $f^1$.
The initial prediction of disparity $k$ at pixel $i$ is formed by the distribution 
\begin{align}
q_i(k) = \softmax_{k \in\{0,1,\dots, D\} } \big( -\| f^0(i) - f^1(i-k) \|_1 \big),
\end{align}
where $i-k$ denotes the pixel location in image $I^1$ corresponding to location $i$ in the reference image $I^0$ and disparity $k$ and $D$ is the maximum disparity.
This model is related to robust costs~\cite{KZ-stereo}.
The pairwise terms $f_{ij}$ are parametric like in the SGM model~\cite{Hirschmuller:2008} but with context-dependent parameters.
Specifically, $f_{ij}$ scores difference of disparity labels in the neighbouring pixels. 
Disparity differences of up to 3 pixels have individual scores, all larger disparity jumps have the same score. All these scores are made context dependent by regressing them with our second UNet from the reference image $I^0$. 
\subsection{Optical Flow}\label{sec:modelFlow}
The optical flow problem is very similar to stereo. 
Instead of two rectified images, we consider now two consecutive frames in a video, $I^0$ and $I^1$. 
We use the same UNets to compute the per-pixel features and the jump scores as in the stereo setting.
The difference lies in the computation of the initial prediction of flow $u=(u_1, u_2)$.
The flow for a pixel $i$ is formed by the two distributions
\begin{align}
q^1_i(u_1) = \softmax_{u_1} \max_{u_2} \big(\!{-}\| f^0(i) - f^1(i{+}u) \|_1 \big),\\
q^2_i(u_2) = \softmax_{u_2} \max_{u_1} \big(\!{-}\| f^0(i) - f^1(i{+}u) \|_1 \big),
\label{eq:minProjection}
\end{align}
which follows the scalable model of~\citet{munda2017scalable}, avoiding the storage of all matching scores that for an $M{\times}N$ image have the size $M{\times}N{\times}D^2$. The inner maximization steps correspond to the first iteration of an approximate MAP inference~\cite{munda2017scalable}. They form an ``optimistic'' estimate of the score volume for each component of the optical flow, which we process then independently. This scheme may be sub-optimal in that $u^1$ and $u^2$ components are inferred independently until the refinement layer, but it scales well to high resolutions (the search window size $D$ needs to grow with the resolution as well) and allows us to readily apply the same \bpl model as for the stereo to $q^1$ and $q^2$ input probabilities.
\subsection{Semantic Segmentation}
\label{ssec:model:semanticsegmentation}
The task in semantic segmentation is to assign a semantic class label from a number of classes to each pixel. In our model, the initial prediction probabilities are obtained with the ESPNet~\cite{Mehta18}, a lightweight solution for pixel-wise semantic segmentation. This initial prediction is followed up directly with the BP-Layer, which can work with two different types of pairwise scores $f_{ij}$. The inhomogeneous anisotropic pairwise terms depend on each pixel and on the edge direction, while the homogeneous anisotropic scores depend only on the edge direction. We implement the homogeneous pairwise terms as parameters within the model and constrain them to be non-negative. The pixel-wise pairwise-terms are computed from the input image using the same UNet as in stereo. We follow the training scheme of \cite{Mehta18}.

\section{Learning}
\label{sec:learning}

We use the same training procedure for all three tasks.
Only the loss function is adapted for the respective task. 
The loss function is applied to the output of each \bpl in the coarse-to-fine scheme and also to the final output after the refinement layer. Such a training scheme is known as deep supervision~\cite{lee2015deeply}. %
For BP output beliefs $B^l$ at level $l$ of the coarse-to-fine scheme, we apply at each pixel $i$ the negative log-likelihood loss %
$\ell_{\rm NLL}(B^l_i, d^{*l}_i) = - \log B^l_i(d^{* l}_i)$,
where $d^{*l}_i$ is the ground truth disparity at scale $l$.

For the stereo and flow models that have a refinement block targeting real-valued predictions, we add a loss penalizing at each pixel the distance from the target value according to the Huber function:
\begin{equation}
    \ell_H(y_i, y_i^*) = 
    \begin{cases}
        \frac{r^2}{2\delta} & \text{if } \lvert r \rvert \leq \delta,\\
        \lvert r \rvert - \frac{\delta}{2} & \text{otherwise},
    \end{cases}
\end{equation}
where $y_i$ is the continuous prediction of the model, $y_i^*$ is the ground-truth and $r = y_i - y_i^*$. 

Losses at all levels and the losses on the continuous-valued outputs are combined with equal weights\footnote{the relative weights could be considered as hyper-parameters, but we did not tune them.}.

\section{Experiments}
\label{sec:experiments}
We implemented the \bpl and hierarchical model in PyTorch and used CUDA extensions for time and memory-critical functions (forward and backward for DP, score volume min-projections).\footnote{\url{https://github.com/VLOGroup/bp-layers}} \cref{sec:implementation,sec:app:moreexamples} contain the implementation details and additional qualitative results. %

\begin{table}[t]
  \centering
  \small
  \setlength\tabcolsep{2pt}
  \begin{tabular}{@{}p{2.1cm}ccccccccc@{}}
    \toprule
    \textbf{Model} & \textbf{\#P} & \textbf{time} & \multicolumn{2}{c}{\textbf{bad1}} & \multicolumn{2}{c}{\textbf{bad3}} &  \multicolumn{2}{c}{\textbf{MAE}} \\
    \midrule
    \textbf{WTA}~(NLL)       & 0.13 & 0.07 & 10.3 & (18.0) & 5.27 & (13.2) & 3.82 & (15.1)   \\
    \textbf{BP}~(NLL)  & 0.27 &  0.10 & 12.6 & (17.9) & 4.97 & (8.12) & 1.23 & (3.36) \\
    \textbf{BP+MS}~(NLL)     & 0.33 & 0.11 & 10.0 & (16.5) & 3.66 & (7.86) & 1.13 & (2.84) \\
    \textbf{BP+MS}~(H)        & 0.33 & 0.11 & \underline{8.15} & (\underline{15.1}) & \underline{3.07} & (8.00) & 0.96 & (3.42) \\
    \textbf{BP+MS+Ref}~(H) & 0.56 & 0.15 & \textbf{7.73} &  (\textbf{13.8}) & \textbf{2.67} & (\underline{6.46}) & 
    \textbf{0.74} & (\underline{1.67}) \\
    \hline
    GC-Net \cite{kendall2017} & 3.5 & 0.95 &- & (16.9) & - & (9.34) & - & (2.51)  \\
    GA-Net-1 \cite{zhang2019ga} & 0.5 & 0.17 & - &(16.5) & - & (-) & - & (1.82) \\
    PDS-Net \cite{Tulyakov2018_NIPS} & 2.2 & - & - & (-) & - & (\textbf{3.38}) & - & (\textbf{1.12}) \\
    \bottomrule
  \end{tabular}
  \caption{Ablation Study on the Scene flow validation set. We report for all metrics the result on non-occluded and (all pixels). \#P in millions. bold = best, underline = second best.} %
  \label{tab:stereoAblation}
\end{table}

\subsection{Improvements brought by the BP-Layer}
\label{ssec:stereo}
We investigate the importance of different architectural choices in our general model on the stereo task with the synthetic stereo data from the Scene Flow dataset~\cite{sceneflowdataset16}.
The standard error metric in stereo is the ${\rm bad}X$ error measuring the percentage of disparities having a distance larger than $X$ to the ground-truth. 
This metric is used to assess the robustness of a stereo algorithm. 
The second metric is the mean-absolute-error (MAE) which is more sensitive to the (sub-pixel) precision of a stereo algorithm.

\cref{tab:stereoAblation} shows an overview of all variants of our model. 
We start from the winner-takes-all {\bf (WTA)} model, add the proposed \bpl or the multi-scale model {\bf (MS)}, then add the basic refinement~\eqref{basic-refine} trained with Huber loss {\bf (H)}, then add the refinement~\cite{khamis2018stereonet} ({\bf Ref~(H)}).
The column \#P in~\cref{tab:stereoAblation} shows the number of parameters of our model, which is significantly smaller than SoTA methods applicable to this dataset. 
Each of the parts of our model increase the final performance. 
Our algorithm performs outstandingly well in the robustness metric ${\rm bad}X$. 
The ablation study shows also the impact of the used loss function. 
It turns out that Huber loss function is beneficial to all the metrics but the MAE in occluded pixels. %
The optional refinement yielded an additional improvement, especially in occluded pixels on this data, but we could not obtain a similar improvement when training and validating on Middlebury or Kitti  datasets. We therefore selected BP+MS~(H) model, as the more robust variant, for evaluation in these real-data benchmarks.

\begin{table}[t]
  \centering
\resizebox{\linewidth}{!}{%
\small%
\setlength\tabcolsep{2pt}%
\begin{tabular}{lc|cc|cccc}%
    \toprule
    \multirow{2}{*}{\textbf{Method}} & \multirow{2}{*}{\textbf{\#P[M]}} & \multicolumn{2}{c|}{\textbf{Middlebury 2014}} & \multicolumn{2}{c}{\textbf{Kitti 2015}} \\
    & & \textbf{bad2} & \textbf{time[s]} & \textbf{bad3} & \textbf{time[s]}\\
    \midrule
    PSMNet \cite{Chang_2018_CVPR} & 5.2 & 42.1 (47.2) & 2.62 & 2.14 (2.32) & 0.41 \\
    PDS \cite{Tulyakov2018_NIPS} & \textbf{2.2} & 14.2 (21.0) & 12.5 & 2.36 (2.58) & 0.50 \\
    HSM \cite{yang2019hierarchical} & 3.2 & \textbf{10.2} (\textbf{16.5}) & \textbf{0.51} & \textbf{1.92} (\textbf{2.14}) & \textbf{0.14}  \\
    \midrule
    MC-CNN \cite{Zbontar2016} & \textbf{0.2} & \textbf{9.47} (20.6) & 1.26 & 3.33 (3.89) & 67.0\\
    CNN-CRF \cite{Knobelreiter_2017_CVPR} & 0.3 & 12.5 (21.9) & 3.53 & 4.84 (5.50) & 1.30 \\
    ContentCNN \cite{luo2016efficient} & 0.7 & - & - &  4.00 (4.54) & 1.00\\
    LBPS {\bf (ours)} & 0.3 & 9.68 (\textbf{17.5}) & \textbf{1.05} &  \textbf{3.13} (\textbf{3.44}) & \textbf{0.39} \\
    \bottomrule
\end{tabular}%
}
  \caption{Evaluation on the Test set of the Middlebury and Kitti Stereo Benchmark using the default metrics of the respective benchmarks. {\em Top group}: Large models with $> 1$M parameters. {\em Bottom group}: Light-weight models. Bold indicates the best result in the group.}
  \label{tab:mbAndKittiTest}
\end{table}

\subsection{Stereo Benchmark Performance}
\label{ssec:stereoBenchmark}
\begin{figure}[t]
    \centering
    \includegraphics[width=0.49\linewidth]{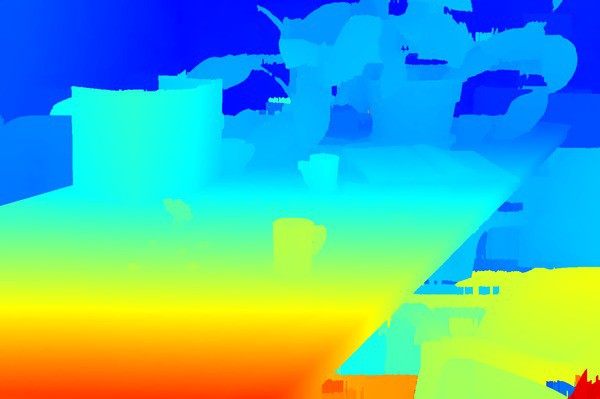}
    \includegraphics[width=0.49\columnwidth]{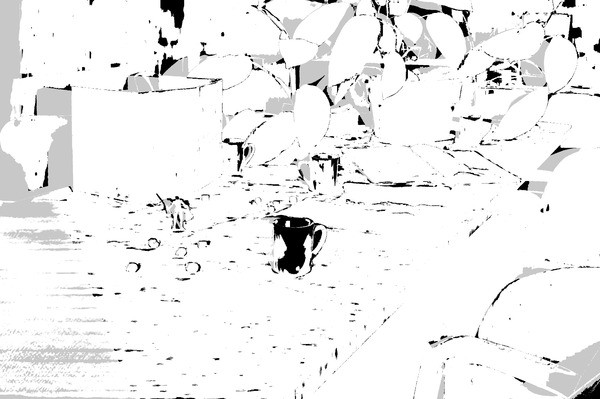}
    \includegraphics[width=0.49\columnwidth]{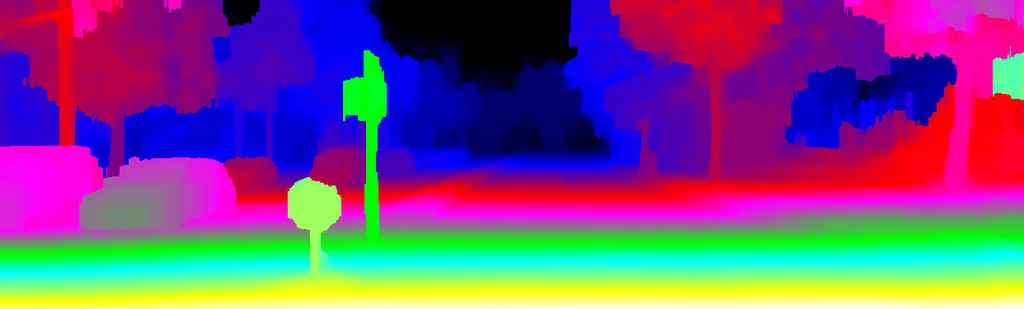}
    \includegraphics[width=0.49\columnwidth]{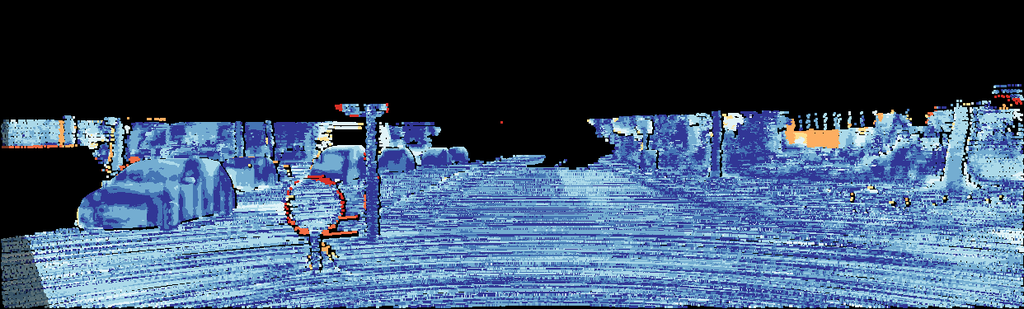}
    \caption{Qualitative results on the test sets of Middlebury 2014 (top)
    and Kitti 2015 (bottom) datasets. Left: Color coded disparity map, right error map, where white/blue = correct, gray = occluded, black/orange = incorrect. Note how our method produces sharp edges in all results.}
    \label{fig:stereoTestset}
\end{figure}

We use the model BP+MS~(H) to participate on the public benchmarks of Middlebury 2014 and Kitti 2015. 
Both benchmarks have real-world scences, Middlebury focusing on high-resolution indoor scenes and Kitti focusing on low-resolution autonomous driving outdoor scenes.
Qualitative test-set results are shown in \cref{fig:stereoTestset}.

The Middlebury benchmark is very challenging due to huge images, large maximum disparities, large untextured regions and difficult illumination. 
These properties make it hard or even impossible for most of the best-performing methods from Kitti to be used on the Middlebury benchmark. 
Due to our light-weight architecture we can easily apply our model on the challenging Middlebury images. The test-set evaluation (\cref{tab:mbAndKittiTest}) shows that we are among the best performing methods with a runtime of up to 10 seconds, and thus convincingly shows the effectiveness of our light-weight model.
The challenges on the Kitti dataset are regions with over- and under-saturation, reflections and complex geometry.
We significantly outperform competitors with a similar number of parameters such as MC-CNN, CNN-CRF and Content CNN, which demonstrates the effectiveness of the learnable \bpl. 
Methods achieving a better performance on Kitti come with the high price of having many more parameters. %

\subsection{Optical Flow}
\label{ssec:opticalFlow}
\begin{figure}[t]
    \centering
    \includegraphics[width=0.45\columnwidth]{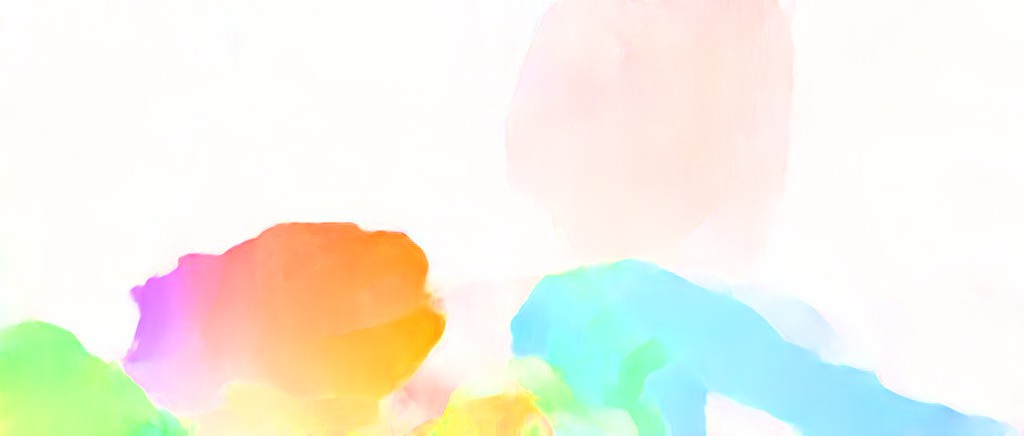}
    \includegraphics[width=0.45\columnwidth]{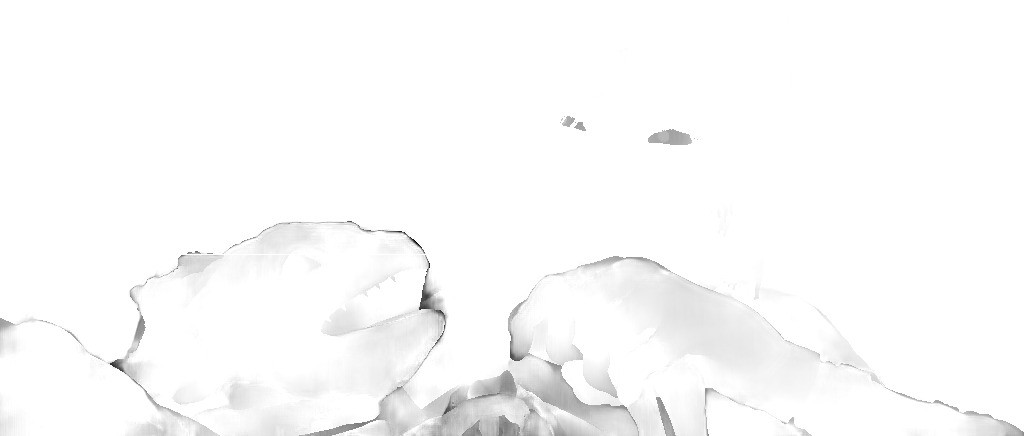}
    \caption{Left: Qualitative optical flow results on the Sintel validation set. Right: Visualization of the endpoint error, where white=correct and darker pixels are erroneous.}
    \label{fig:flowRes}
\end{figure}

Here we show the applicability of our \bpl to the optical flow problem.
We use the FlyingChairs2 dataset \cite{dosovitskiy_flownet,ilg_flow} for pre-training our model and fine-tune then with the Sintel dataset \cite{butler_sintel}. 
In the optical flow setting we set the search-window-size to $109 \times 109$ in the finest resolution.
We compute the $109^2$ similarities per pixel without storing them and compute the two cost-volumes $q^1$ and $q^2$ using \cref{eq:minProjection} on the fly. %
\cref{fig:flowRes} shows qualitative results and \cref{tab:flowAblation} shows the ablation study on the validation set of the Sintel dataset. 
We use only scenes where the flow is not larger than our search-window in this study.
We compare the endpoint-error (EPE) and the ${\rm bad}2$ error on the EPE. 
The results show that our \bpl can be directly used for optical flow computation and that the \bpl is an important building block to boost performance.

\begin{table}[t]
  \centering
  \small
  \setlength\tabcolsep{4pt}
  \begin{tabular}{lcccc}
    \toprule
    \textbf{Model} & \textbf{\#P[M]} & \textbf{time} & \textbf{bad2} &  \textbf{EPE} \\
    \midrule
    \textbf{WTA}         & \textbf{0.13} & \textbf{0.27} & 4.46 (5.67) & 1.25 (1.65) \\
    \textbf{BP+MS}~(CE)   & 0.34 & 0.44 & 2.56 (3.46) &  0.83 (0.94)\\
    \textbf{BP+MS}~(H)    & 0.34 & 0.44 & 2.24 (3.19) & 0.66 (0.79) \\
    \textbf{BP+MS+Ref}~(H) & 0.56 & 0.49 & \textbf{2.06 (2.64)} & \textbf{0.63 (0.72)}\\
    \bottomrule
  \end{tabular}
  \caption{Ablation Study on the Sintel Validation set.}
  \label{tab:flowAblation}
\end{table}

\subsection{Semantic Segmentation}
\label{ssec:sematicSegmentation}
\begin{figure}[t]
    \centering
    \begin{subfigure}[b]{0.53\columnwidth}
    \includegraphics[width=\textwidth, clip, trim={0, 180, 0, 0}]{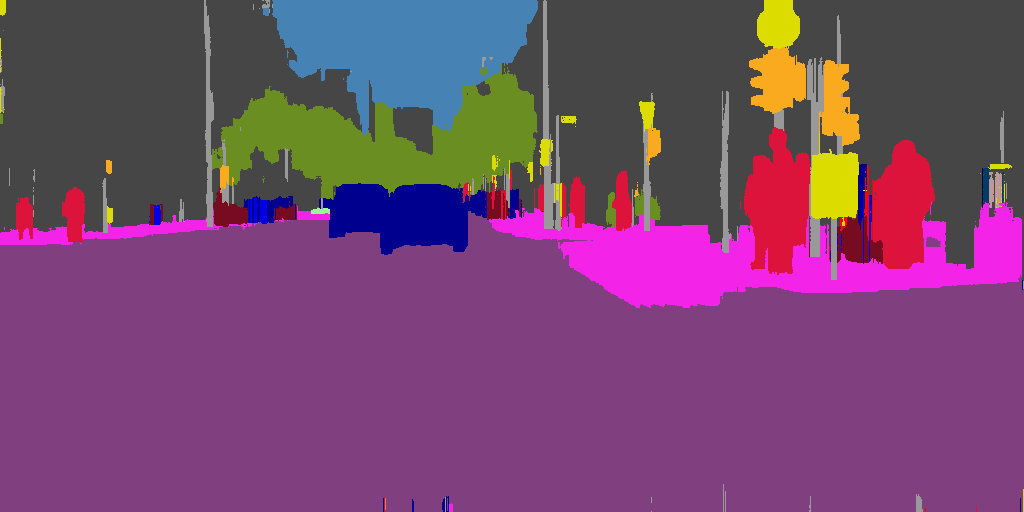} \\
    \includegraphics[width=\textwidth, clip, trim={0, 180, 0, 0}]{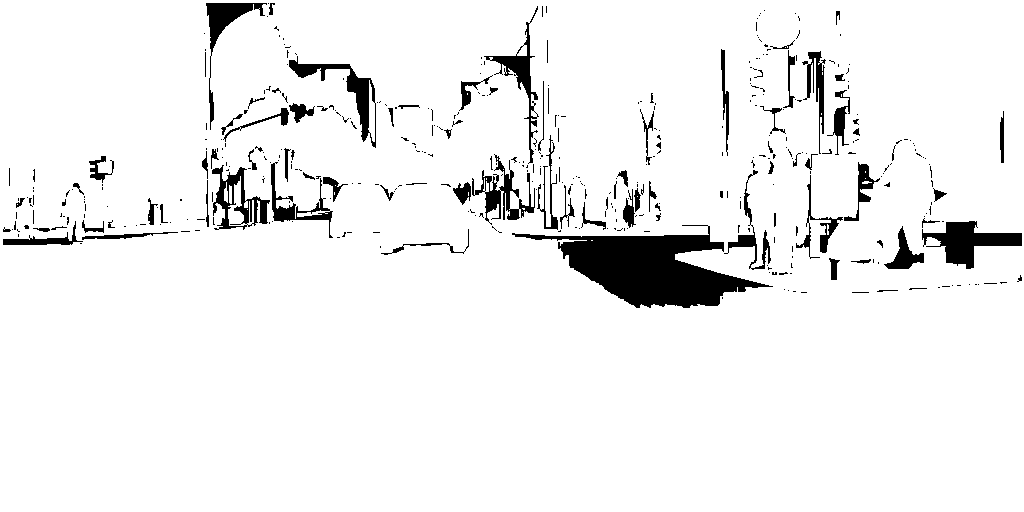}
    \end{subfigure}
    \hfill
   \begin{subfigure}[b]{0.45\columnwidth}
    \includegraphics[width=\textwidth]{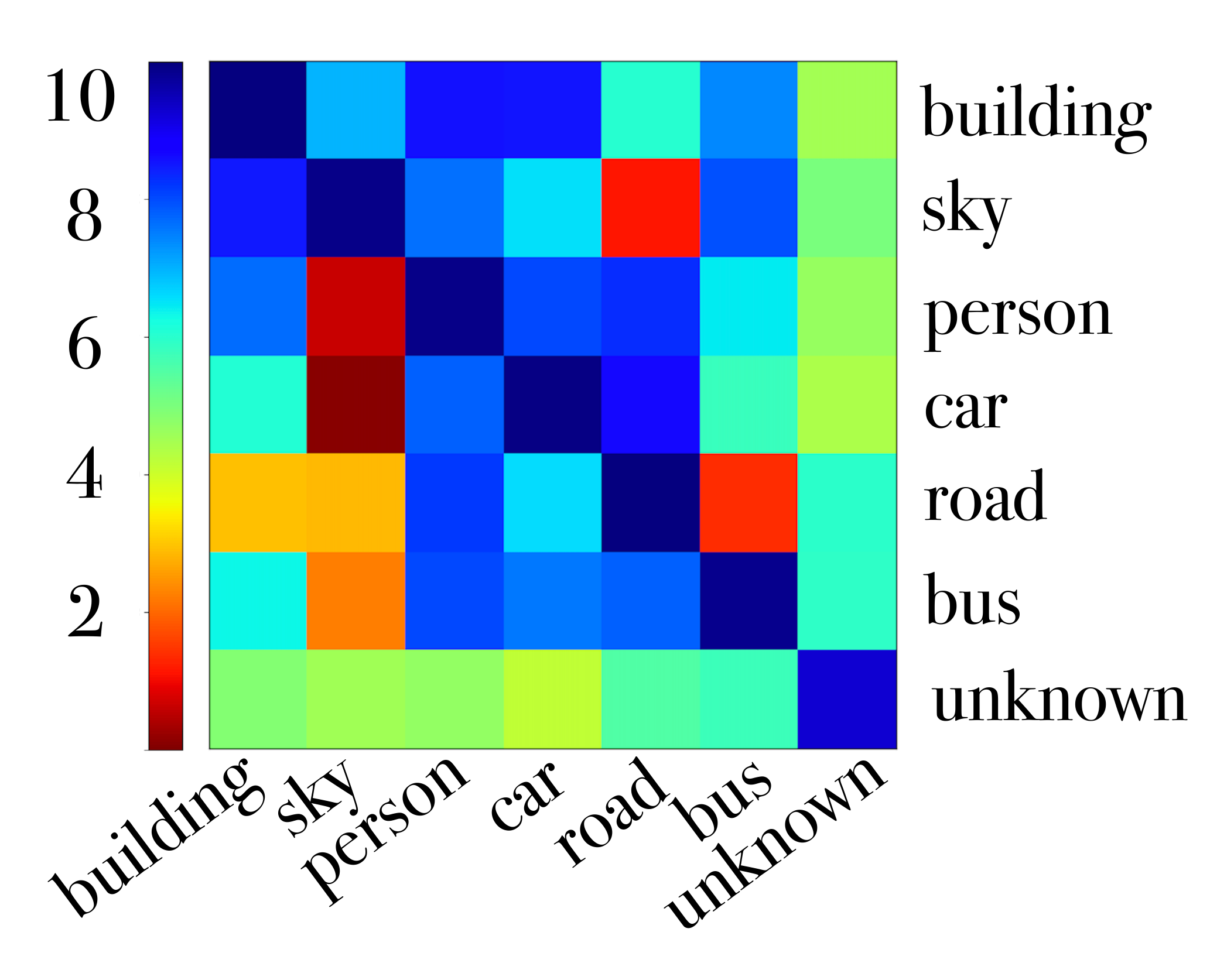} 
    \end{subfigure}
    \begin{tikzpicture}[overlay]
    \draw[red, thick, scale=0.8, xshift=-6.7cm, yshift=2.6cm] (0,0) -- (0,1) -- (1,1) -- (1,0) --(0,0);
    \draw[red, thick, scale=0.8, xshift=-6.7cm, yshift=0.8cm] (0,0) -- (0,1) -- (1,1) -- (1,0) --(0,0);
    \end{tikzpicture}
    
    \vspace{-0.5em}
    \caption{{\em Top Left}: Semantic segmentation result with the \bpl. {\em Bottom Left}: Corresponding error where black = incorrect, white = correct. The red square highlights the region where fine details were accurately reconstructed.
    {\em Right}: Visualization of learned 
    vertical 
    pairwise scores. }
    \label{fig:sem:pw}
\end{figure}

We apply the \bpl also to semantic segmentation to demonstrate its general applicability.
In~\cref{tab:ssAblation:20} we show results with our model variants described in \cref{ssec:model:semanticsegmentation} using the same CNN block as ESPNet~\cite{Mehta18}, evaluated on the Cityscapes~\cite{Cordts16} dataset. %
All model variants using the \bpl improve on ESPNet~\cite{Mehta18} in both the class mean intersection over union (mIOU) and the category mIOU. The best model is, as expected, the jointly trained pixel-wise model referred to as \textit{LBPSS joint}.
We have submitted this model to the Cityscapes benchmark. 
\Cref{tab:ssTest:avg} shows the results on the test set and we can see that we outperform the baseline.
\Cref{fig:sem:pw} shows that
the \bpl refines the prediction by aligning the semantic boundaries to actual object boundaries in the image. 
Due to the long range interaction, the \bpl is also able to correct large incorrect regions such as on \eg the road.
One of the advantages of our model is that the learned parameters can be interpreted. 
\cref{fig:sem:pw} shows the learned non-symmetric score matrix, which allows to learn different scores for \eg person $\rightarrow$ car and car $\rightarrow$ person. 
The upper and lower triangular matrix represent pairwise scores when jumping upwards and downwards in the image, respectively. %
We can read from the matrix that, \eg, an upward jump from sky to road is not allowed.
This confirms the intuition, since the road never occurs above the sky. 
Our model has thus automatically learned appropriate semantic relations which have been hand-crafted in prior work such as \eg \cite{felzenszwalb2010tiered}.

\begin{table}[t]
  \centering
  \small
  \setlength\tabcolsep{6pt}
  \begin{tabular}{lccccc}
    \toprule
    \textbf{Method} & \textbf{pw} & \textbf{mIOU} & \textbf{CatmIOU} &  \textbf{\#P} & \textbf{time} \\
     \hline
     ESPNet~\cite{Mehta18} & - & 61.4 & 82.2 & \textbf{0.36} & \textbf{0.01}  \\ %
     \shortstack{LBPSS} & - & 62.8 & 83.0 & 0.37 & 0.11  \\ %
     \shortstack{LBPSS} & \checkmark &  63.6 & 83.7 & 0.73 & 0.90  \\ %
      \shortstack{LBPSS joint} & \checkmark & \textbf{65.2} & \textbf{84.7} & 0.73 & 0.90  \\ %
     \bottomrule
  \end{tabular}
  \caption{Ablation study on the Cityscapes validation set. \textit{``pw''} = pixel-wise, inhomogeneous scores.}
  \label{tab:ssAblation:20}
\end{table}

\begin{table}[t]
  \centering
  \small
  \setlength\tabcolsep{6pt}
  \begin{tabular}{lccccc}
    \toprule
    \textbf{Method} & \textbf{pw} & \textbf{mIOU} & \textbf{CatmIOU} & \textbf{\#P} & \textbf{time} \\
     \hline
     ESPNet~\cite{Mehta18} & - & 60.34 & 82.18 & \textbf{0.36} & \textbf{0.01}  \\
     \shortstack{LBPSS joint} & \checkmark & \textbf{61.00} & \textbf{84.31} & 0.73 & 0.90 \\ %
      \bottomrule
  \end{tabular}
  \caption{Benchmark results on the Cityscapes~\cite{Cordts16} test set.} %
  \label{tab:ssTest:avg}
\end{table}

\section{Conclusion}
\label{sec:conclusion}
We have proposed a novel combination of CNN and CRF techniques, aiming to resolve practical challenges. We took one of the simplest inference schemes, showed how to compute its backprop and connected it with the marginal losses. 
The following design choices were important for achieving a high practical utility: using max-product for fast computation and backprop of approximate marginals, propagating the information over a long range with sequential subproblems; training end-to-end without approximations; coarse-to-fine processing at several resolution levels; context-dependent learnable unary and pairwise costs.
We demonstrated the model can be applied to three dense prediction problems and gives robust solutions with more efficient parameter complexity and time budget than comparable CNNs. In particular in stereo and flow, the model performs strong regularization in occluded regions and this regularization mechanism is interpretable in terms of robust fitting with jump scores.

\vspace{0.5em}
\noindent
\textbf{Acknowledgements} This work was supported by the ERC starting grant HOMOVIS (No. 640156), Pro$^2$Future (FFG No. 854184) and the project ``International Mobility of Researchers MSCA-IF II at CTU in Prague''
(CZ.02.2.69/0.0/0.0/18\_070/0010457). %

\clearpage
{\small
\bibliographystyle{splncsnat}
\bibliography{newbib}

\begin{thebibliography}{54}
\providecommand{\natexlab}[1]{#1}
\providecommand{\url}[1]{\texttt{#1}}
\providecommand{\urlprefix}{}

\bibitem[{Alahari et~al.(2010)Alahari, Russell, and Torr}]{Alahari10a}
Alahari, K., Russell, C., Torr, P.H.S.: Efficient piecewise learning for
  conditional random fields.
\newblock In: Conference on Computer Vision and Pattern Recognition (2010)

\bibitem[{Bleyer and Gelautz(2008)}]{Bleyer08simplebut}
Bleyer, M., Gelautz, M.: Simple but effective tree structures for dynamic
  programming-based stereo matching.
\newblock In: In VISAPP. pp. 415--422 (2008)

\bibitem[{Butler et~al.(2012)Butler, Wulff, Stanley, and Black}]{butler_sintel}
Butler, D.J., Wulff, J., Stanley, G.B., Black, M.J.: A naturalistic open source
  movie for optical flow evaluation.
\newblock In: {A. Fitzgibbon et al. (Eds.)} (ed.) European Conference on
  Computer Vision (ECCV). pp. 611--625 (2012)

\bibitem[{Chang and Chen(2018)}]{Chang_2018_CVPR}
Chang, J.R., Chen, Y.S.: Pyramid stereo matching network.
\newblock In: IEEE Conference on Computer Vision and Pattern Recognition
  (CVPR). pp. 5410--5418 (2018)

\bibitem[{Chen et~al.(2015)Chen, Papandreou, Kokkinos, Murphy, and
  Yuille}]{chen14semantic}
Chen, L.C., Papandreou, G., Kokkinos, I., Murphy, K., Yuille, A.L.: Semantic
  image segmentation with deep convolutional nets and fully connected crfs.
\newblock In: {ICLR} (2015)

\bibitem[{Cheng et~al.(2018)Cheng, Wang, and Yang}]{Xinjing-18-CSPN}
Cheng, X., Wang, P., Yang, R.: Learning depth with convolutional spatial
  propagation network.
\newblock {CoRR} abs/1810.02695 (2018)

\bibitem[{Cordts et~al.(2016)Cordts, Omran, Ramos, Rehfeld, Enzweiler,
  Benenson, Franke, Roth, and Schiele}]{Cordts16}
Cordts, M., Omran, M., Ramos, S., Rehfeld, T., Enzweiler, M., Benenson, R.,
  Franke, U., Roth, S., Schiele, B.: The cityscapes dataset for semantic urban
  scene understanding.
\newblock In: IEEE Conference on Computer Vision and Pattern Recognition (CVPR)
  (2016)

\bibitem[{Domke(2011)}]{Domke2011ParameterLW}
Domke, J.: Parameter learning with truncated message-passing.
\newblock IEEE Conference on Computer Vision and Pattern Recognition (CVPR) pp.
  2937--2943 (2011)

\bibitem[{Domke(2013)}]{Domke-learning}
Domke, J.: Learning graphical model parameters with approximate marginal
  inference.
\newblock IEEE Transactions on Pattern Analysis and Machine Intelligence
  35(10), 2454--2467 (October 2013)

\bibitem[{Dosovitskiy et~al.(2015)Dosovitskiy, Fischer, Ilg, H{\"a}usser,
  Haz{\i}rba{\c{s}}, Golkov, v.d. Smagt, Cremers, and
  Brox}]{dosovitskiy_flownet}
Dosovitskiy, A., Fischer, P., Ilg, E., H{\"a}usser, P., Haz{\i}rba{\c{s}}, C.,
  Golkov, V., v.d. Smagt, P., Cremers, D., Brox, T.: Flownet: Learning optical
  flow with convolutional networks.
\newblock In: IEEE International Conference on Computer Vision (ICCV) (2015)

\bibitem[{Eaton and Ghahramani(2009)}]{eaton2009choosing}
Eaton, F., Ghahramani, Z.: Choosing a variable to clamp: {A}pproximate
  inference using conditioned belief propagation.
\newblock In: International Conference on Artificial Intelligence and
  Statistics. vol.~5, pp. 145--152 (April 2009)

\bibitem[{Felzenszwalb and Veksler(2010)}]{felzenszwalb2010tiered}
Felzenszwalb, P.F., Veksler, O.: Tiered scene labeling with dynamic
  programming.
\newblock In: 2010 IEEE Computer Society Conference on Computer Vision and
  Pattern Recognition. pp. 3097--3104 (2010)

\bibitem[{Hirschmüller(2008)}]{Hirschmuller:2008}
Hirschmüller, H.: Stereo processing by semiglobal matching and mutual
  information.
\newblock IEEE Transactions on Pattern Analysis and Machine Intelligence 30(2),
  328--341 (February 2008)

\bibitem[{Ilg et~al.(2018)Ilg, Saikia, Keuper, and Brox}]{ilg_flow}
Ilg, E., Saikia, T., Keuper, M., Brox, T.: Occlusions, motion and depth
  boundaries with a generic network for disparity, optical flow or scene flow
  estimation.
\newblock In: European Conference on Computer Vision (ECCV) (2018)

\bibitem[{Kakade et~al.(2002)Kakade, Teh, and Roweis}]{Kakade02analternate}
Kakade, S., Teh, Y.W., Roweis, S.T.: An alternate objective function for
  markovian fields (2002)

\bibitem[{Kendall et~al.(2017)Kendall, Martirosyan, Dasgupta, Henry, Kennedy,
  Bachrach, and Bry}]{kendall2017}
Kendall, A., Martirosyan, H., Dasgupta, S., Henry, P., Kennedy, R., Bachrach,
  A., Bry, A.: End-to-end learning of geometry and context for deep stereo
  regression.
\newblock In: IEEE International Conference on Computer Vision (ICCV) (2017)

\bibitem[{Keshet(2014)}]{Keshet-14}
Keshet, J.: Optimizing the Measure of Performance (2014)

\bibitem[{Khamis et~al.(2018)Khamis, Fanello, Rhemann, Kowdle, Valentin, and
  Izadi}]{khamis2018stereonet}
Khamis, S., Fanello, S., Rhemann, C., Kowdle, A., Valentin, J., Izadi, S.:
  Stereonet: Guided hierarchical refinement for real-time edge-aware depth
  prediction.
\newblock In: European Conference on Computer Vision (ECCV). pp. 573--590
  (2018)

\bibitem[{Kingma and Ba(2014)}]{kingma2014adam}
Kingma, D.P., Ba, J.: Adam: A method for stochastic optimization.
\newblock arXiv preprint arXiv:1412.6980  (2014)

\bibitem[{Kirillov et~al.(2015)Kirillov, Schlesinger, Forkel, Zelenin, Zheng,
  Torr, and Rother}]{KirillovSFZ0TR15}
Kirillov, A., Schlesinger, D., Forkel, W., Zelenin, A., Zheng, S., Torr,
  P.H.S., Rother, C.: Efficient likelihood learning of a generic {CNN-CRF}
  model for semantic segmentation.
\newblock {CoRR} abs/1511.05067 (2015)

\bibitem[{Kn\"obelreiter and Pock(2019)}]{Knobelreiter_2019_GCPR}
Kn\"obelreiter, P., Pock, T.: Learned collaborative stereo refinement.
\newblock In: German Conference on Pattern Recognition (GCPR) (2019)

\bibitem[{Kn\"obelreiter et~al.(2017)Kn\"obelreiter, Reinbacher, Shekhovtsov,
  and Pock}]{Knobelreiter_2017_CVPR}
Kn\"obelreiter, P., Reinbacher, C., Shekhovtsov, A., Pock, T.: End-to-end
  training of hybrid cnn-crf models for stereo.
\newblock In: IEEE Conference on Computer Vision and Pattern Recognition
  (CVPR). pp. 2339--2348 (2017)

\bibitem[{Kohli and Torr(2006)}]{Kohli:2006}
Kohli, P., Torr, P.H.S.: Measuring uncertainty in graph cut solutions --
  efficiently computing min-marginal energies using dynamic graph cuts.
\newblock In: European Conference on Computer Vision (ECCV). pp. 30--43.
  Springer-Verlag (2006)

\bibitem[{Kolmogorov et~al.(2014)Kolmogorov, Monasse, and Tan}]{KZ-stereo}
Kolmogorov, V., Monasse, P., Tan, P.: {Kolmogorov and Zabih’s Graph Cuts
  Stereo Matching Algorithm}.
\newblock {Image Processing On Line} 4, 220--251 (2014)

\bibitem[{Lauritzen(1998)}]{Lauritzen96}
Lauritzen, S.L.: Graphical Models.
\newblock No.~17 in Oxford Statistical Science Series, Oxford Science
  Publications (1998)

\bibitem[{Lee et~al.(2015)Lee, Xie, Gallagher, Zhang, and Tu}]{lee2015deeply}
Lee, C.Y., Xie, S., Gallagher, P., Zhang, Z., Tu, Z.: Deeply-supervised nets.
\newblock In: Artificial intelligence and statistics. pp. 562--570 (2015)

\bibitem[{Lin et~al.(2015)Lin, Shen, Reid, and van~den Hengel}]{LinSRH15}
Lin, G., Shen, C., Reid, I.D., van~den Hengel, A.: Efficient piecewise training
  of deep structured models for semantic segmentation.
\newblock {CoRR} abs/1504.01013 (2015)

\bibitem[{Liu et~al.(2017)Liu, De~Mello, Gu, Zhong, Yang, and
  Kautz}]{Liu-17-SPN}
Liu, S., De~Mello, S., Gu, J., Zhong, G., Yang, M.H., Kautz, J.: Learning
  affinity via spatial propagation networks.
\newblock In: Proceedings of Advances in Neural Information Processing Systems,
  pp. 1520--1530 (2017)

\bibitem[{Luo et~al.(2016)Luo, Schwing, and Urtasun}]{luo2016efficient}
Luo, W., Schwing, A.G., Urtasun, R.: Efficient deep learning for stereo
  matching.
\newblock In: IEEE Conference on Computer Vision and Pattern Recognition
  (CVPR). pp. 5695--5703 (2016)

\bibitem[{Marin et~al.(2019)Marin, Tang, Ayed, and Boykov}]{Marin_2019_CVPR}
Marin, D., Tang, M., Ayed, I.B., Boykov, Y.: Beyond gradient descent for
  regularized segmentation losses.
\newblock In: IEEE Conference on Computer Vision and Pattern Recognition (CVPR)
  (June 2019)

\bibitem[{Mayer et~al.(2016)Mayer, Ilg, H{\"a}usser, Fischer, Cremers,
  Dosovitskiy, and Brox}]{sceneflowdataset16}
Mayer, N., Ilg, E., H{\"a}usser, P., Fischer, P., Cremers, D., Dosovitskiy, A.,
  Brox, T.: A large dataset to train convolutional networks for disparity,
  optical flow, and scene flow estimation.
\newblock In: IEEE Conference on Computer Vision and Pattern Recognition (CVPR)
  (2016)

\bibitem[{Mehta et~al.(2018)Mehta, Rastegari, Caspi, Shapiro, and
  Hajishirzi}]{Mehta18}
Mehta, S., Rastegari, M., Caspi, A., Shapiro, L., Hajishirzi, H.: Espnet:
  Efficient spatial pyramid of dilated convolutions for semantic segmentation.
\newblock In: European Conference on Computer Vision (ECCV) (2018)

\bibitem[{Mensch and Blondel(2018)}]{mensch2018differentiable}
Mensch, A., Blondel, M.: Differentiable dynamic programming for structured
  prediction and attention (2018)

\bibitem[{Munda et~al.(2017)Munda, Shekhovtsov, Kn{\"o}belreiter, and
  Pock}]{munda2017scalable}
Munda, G., Shekhovtsov, A., Kn{\"o}belreiter, P., Pock, T.: Scalable full flow
  with learned binary descriptors.
\newblock In: German Conference on Pattern Recognition (GCPR). pp. 321--332
  (2017)

\bibitem[{Pal et~al.(2012)Pal, Weinman, Tran, and
  Scharstein}]{Christopher-12-learnig}
Pal, C.J., Weinman, J.J., Tran, L.C., Scharstein, D.: On learning conditional
  random fields for stereo - exploring model structures and approximate
  inference.
\newblock International Journal of Computer Vision 99(3), 319--337 (2012)

\bibitem[{Papandreou and Yuille(2014)}]{PaYu14}
Papandreou, G., Yuille, A.: Perturb-and-map random fields: Reducing random
  sampling to optimization, with applications in computer vision.
\newblock In: Nowozin, S., Gehler, P., Jancsary, J., Lampert, C. (eds.)
  Advanced Structured Prediction. MIT-Press (2014)

\bibitem[{Pearl(1982)}]{Pearl:1982}
Pearl, J.: Reverend {B}ayes on inference engines: {A} distributed hierarchical
  approach.
\newblock In: AAAI. pp. 133--136 (1982)

\bibitem[{Pearl(1988)}]{Pearl:1988}
Pearl, J.: Probabilistic Reasoning in Intelligent Systems: Networks of
  Plausible Inference.
\newblock Morgan Kaufmann Publishers Inc., San Francisco, CA, USA (1988)

\bibitem[{Riegler et~al.(2016)Riegler, R{\"u}ther, and
  Bischof}]{riegler2016atgv}
Riegler, G., R{\"u}ther, M., Bischof, H.: Atgv-net: Accurate depth
  super-resolution.
\newblock In: European Conference on Computer Vision (ECCV). pp. 268--284
  (2016)

\bibitem[{Seki and Pollefeys(2017)}]{Seki-2017-CVPR}
Seki, A., Pollefeys, M.: Sgm-nets: Semi-global matching with neural networks.
\newblock In: IEEE Conference on Computer Vision and Pattern Recognition (CVPR)
  (July 2017)

\bibitem[{Sontag and Jaakkola(2009)}]{sontag2009tree}
Sontag, D., Jaakkola, T.: Tree block coordinate descent for {MAP} in graphical
  models.
\newblock In: Artificial Intelligence and Statistics. pp. 544--551 (2009)

\bibitem[{Szeliski et~al.(2008)Szeliski, Zabih, Scharstein, Veksler,
  Kolmogorov, Agarwala, Tappen, and Rother}]{Szeliski:2008}
Szeliski, R., Zabih, R., Scharstein, D., Veksler, O., Kolmogorov, V., Agarwala,
  A., Tappen, M., Rother, C.: A comparative study of energy minimization
  methods for markov random fields with smoothness-based priors.
\newblock IEEE Transactions on Pattern Analysis and Machine Intelligence 30(6),
  1068--1080 (June 2008)

\bibitem[{Tappen and Freeman(2003)}]{tappen03}
Tappen, M., Freeman, W.T.: Comparison of graph cuts with belief propagation for
  stereo, using identical mrf parameters.
\newblock In: IEEE International Conference on Computer Vision (ICCV). pp.
  900--906 (2003)

\bibitem[{Tulyakov et~al.(2018)Tulyakov, Ivanov, and
  Fleuret}]{Tulyakov2018_NIPS}
Tulyakov, S., Ivanov, A., Fleuret, F.: Practical deep stereo (pds): Toward
  applications-friendly deep stereo matching.
\newblock In: Proceedings of Advances in Neural Information Processing Systems.
  pp. 5871--5881 (2018)

\bibitem[{Van Den~Oord et~al.(2016)Van Den~Oord, Kalchbrenner, and
  Kavukcuoglu}]{VanDenOord-16}
Van Den~Oord, A., Kalchbrenner, N., Kavukcuoglu, K.: Pixel recurrent neural
  networks.
\newblock In: {ICML}. pp. 1747--1756 (2016)

\bibitem[{Vemulapalli et~al.(2016)Vemulapalli, Tuzel, Liu, and
  Chellapa}]{vemulapalli2016gaussianCRF}
Vemulapalli, R., Tuzel, O., Liu, M.Y., Chellapa, R.: Gaussian conditional
  random field network for semantic segmentation.
\newblock In: IEEE Conference on Computer Vision and Pattern Recognition
  (CVPR). pp. 3224--3233 (2016)

\bibitem[{Vogel et~al.(2018)Vogel, Kn{\"o}belreiter, and
  Pock}]{vogel2018learning}
Vogel, C., Kn{\"o}belreiter, P., Pock, T.: Learning energy based inpainting for
  optical flow.
\newblock In: Asian Conference on Computer Vision (ACCV) (2018)

\bibitem[{Wainwright et~al.(2003)Wainwright, Jaakkola, and
  Willsky}]{wainwright2003tree}
Wainwright, M.J., Jaakkola, T.S., Willsky, A.S.: Tree-reweighted belief
  propagation algorithms and approximate ml estimation by pseudo-moment
  matching.
\newblock In: International Conference on Artificial Intelligence and
  Statistics (2003)

\bibitem[{Yang et~al.(2019)Yang, Manela, Happold, and
  Ramanan}]{yang2019hierarchical}
Yang, G., Manela, J., Happold, M., Ramanan, D.: Hierarchical deep stereo
  matching on high-resolution images.
\newblock In: IEEE Conference on Computer Vision and Pattern Recognition
  (CVPR). pp. 5515--5524 (2019)

\bibitem[{Yang et~al.(2014)Yang, Ji, Li, Yao, and Zhang}]{YANG2014}
Yang, Q., Ji, P., Li, D., Yao, S., Zhang, M.: Fast stereo matching using
  adaptive guided filtering.
\newblock Image and Vision Computing 32(3), 202 -- 211 (2014)

\bibitem[{Yin et~al.(2019)Yin, Darrell, and Yu}]{hd3_cvpr}
Yin, Z., Darrell, T., Yu, F.: Hierarchical discrete distribution decomposition
  for match density estimation.
\newblock In: IEEE Conference on Computer Vision and Pattern Recognition (CVPR)
  (June 2019)

\bibitem[{{\v{Z}}bontar and LeCun(2016)}]{Zbontar2016}
{\v{Z}}bontar, J., LeCun, Y.: Stereo matching by training a convolutional
  neural network to compare image patches.
\newblock Journal of Machine Learning Research  (2016)

\bibitem[{Zhang et~al.(2019)Zhang, Prisacariu, Yang, and Torr}]{zhang2019ga}
Zhang, F., Prisacariu, V., Yang, R., Torr, P.H.: Ga-net: Guided aggregation net
  for end-to-end stereo matching.
\newblock In: IEEE Conference on Computer Vision and Pattern Recognition
  (CVPR). pp. 185--194 (2019)

\bibitem[{Zheng et~al.(2015)Zheng, Jayasumana, Romera-Paredes, Vineet, Su, Du,
  Huang, and Torr}]{zheng2015conditional}
Zheng, S., Jayasumana, S., Romera-Paredes, B., Vineet, V., Su, Z., Du, D.,
  Huang, C., Torr, P.H.: Conditional random fields as recurrent neural
  networks.
\newblock In: IEEE Conference on Computer Vision and Pattern Recognition
  (CVPR). pp. 1529--1537 (2015)

\end{thebibliography}
}

\clearpage
\onecolumn
\twocolumn[
\begin{center}
{\Large
\vskip0.5cm
\textbf{Belief Propagation Reloaded: Learning BP-Layers for Labeling Problems} \\ Supplementary Material
\vskip1cm}
\end{center}
]
\appendix
\addtocontents{toc}{\protect\setcounter{tocdepth}{2}}
\tableofcontents
\numberwithin{figure}{section}
\setcounter{figure}{0}
\setcounter{table}{0}
\counterwithin{figure}{section}
\counterwithin{table}{section}

\section{Differentiable TRW and TBCA algorithms}\label{sec:A}
Here we consider two other inference methods that have similar properties of long-range spatial propagation and parallelization and can be implemented with same or similar subroutines. As they improve on the issues of BP in loopy graphs, this makes them potential candidates for drop-in replacement of our sweep BP-layer.
\paragraph{Tree Re-weighted BP}
\citet{wainwright2003tree} proposed a correction to BP, which turns it into a variational inference algorithm optimizing the dual of the LP relaxation.
Suppose that we are given an edge-disjoint decomposition of the graph into trees. For our models it is convenient to take horizontal and vertical chain subproblems. The TRW-T algorithm~\cite{wainwright2003tree} can be implemented as proposed in~\cref{alg:TRW}. In this representation we keep the decomposition into subproblems explicitly and messages are encapsulated in the computation of max-marginals. This is in order to reuse the same subroutines we already have for \bpl. An explicit form of updates in terms of messages only which reveals the similarity to loopy belief propagation with weighting coefficients can be also given~\cite{wainwright2003tree}.
This algorithm is not guaranteed to be monotonous because it does block-coordinate ascent steps in multiple blocks in parallel. However thanks to parallelization it is fast to compute (in particular on a GPU), incorporates long-range interactions and avoids the over-counting problems associated with loopy BP~\cite{wainwright2003tree}.
\begin{algorithm}[t]
\KwIn{CRF scores $g\in \Real^{\V \times \L}$, $f\in \Real^{\E \times \L^2}$\;}
\KwOut{Beliefs $B \in \Real^{\V\times \L}$\;}
$g^{\rm h} := g^{\rm v} := \tfrac{1}{2} g$\;
\For{iteration $t = 1 \dots T$}{
\tcc{Compute max-marginals:}
\ForP{horizontal chain subgraphs $(\V', \E')$}{
	$b^{\rm h}_{\V'} := \maxmarginals(g^{\rm h}_{\V'}, f_{\E'})$\;
}
\ForP{vertical chain subgraphs $(\V', \E')$}{
	$b^{\rm v}_{\V'} := \maxmarginals(g^{\rm v}_{\V'}, f_{\E'})$\;
}
\tcc{Enforce consistency:}
	$b := (b^{\rm h} + b^{\rm v})$\;
	$g^{\rm h} \:{+}{=}\: (\tfrac{1}{2}b - b^h)$\;
	$g^{\rm v} \:{+}{=}\: (\tfrac{1}{2}b - b^v)$\;
}
\Return Log-beliefs $b$\;
\caption{Tree Reweighted BP (TRW-T)~\label{alg:TRW}}
\end{algorithm}
\paragraph{Tree Block Coordinate Ascent}
The TBCA method~\cite{sontag2009tree} is an inference algorithm optimizing the dual of the LP relaxation. It does so by a block-coordinate ascent in the variables associated with tree-structured subproblems. The variables are the same as the messages in BP. At each iteration a sub-tree $(\V', \E')$ from the graph is selected. For simplicity and ease of parallelization we will assume $(\V',\E')$ is a horizontal chain and consider it to be ordered from left to right. The following updates are performed on this chain:
\begin{itemize}
\item Compute the current reparametrized costs, excluding the messages from inside the chain:
\begin{align}
a_{i}(s) = g_i(s) + \sum_{(i,j) \in \E \backslash \E'} m_{ji}(s) \forall i\in \V'.
\end{align}
\item Compute the right messages $m^{\rm R}$ by DP in the direction R$\rightarrow$L.
\item Compute the left messages $m^{\rm L}$ by a {\em redistribution DP} (rDP) in the direction R$\rightarrow$L.
\end{itemize}
We can write the rDP update equation~\cite{sontag2009tree} in the form
\begin{align}
m^{\rm L}_{i+1}(t) := \max_{s}\Big(\tilde g_i(s) + r_i m^{\rm L}_{i}(s) + f_{i, i+1}(s,t)\Big),
\end{align}
where $\tilde g_{i}(s) = g_{i}(s) + (1-r_i) m^{\rm R}_{i}(s)$ and $r_i \in [0,1]$ are the redistribution coefficients. For $r=1$, this recovers the regular dynamic programming. Similarly to DP, the update is linear and depend on the current maximizers that we record as $o_{i+1}(t)$. It differs from DP in two ways: i) it depends on the right messages, which we have taken into account by incorporating them to the unary costs in $\tilde g_{i}(s)$ and ii) there are coefficients $r_i$ in the recursion. To handle the latter, we only need to modify \cref{back-DP:line:z} of \cref{alg:back-DP} to
\begin{align}
z := \dv{m}_{i+1}(t) + r_{i+1} \dv{g}_{i+1}(t).
\end{align}
It follows that we have defined the TBCA subproblem update with standard operations on tensors and the two new operations DP and rDP, for which we have shown efficient backprop methods.
The TBCA method~\cite{sontag2009tree}, when specialized to horizontal and vertical chains, would then alternate the above updates in parallel for all horizontal chains and then for all vertical chains. This method also achieves high parallelization efficiency and long-range propagation. Thanks to the redistribution mechanism it is also guaranteed to be monotonous. However, this monotonicity may slow down the information propagation, which can make it less suitable as a truncated inference technique in deep learning. 

\begin{figure}[t]
\centering
 \includegraphics[width=0.9\linewidth]{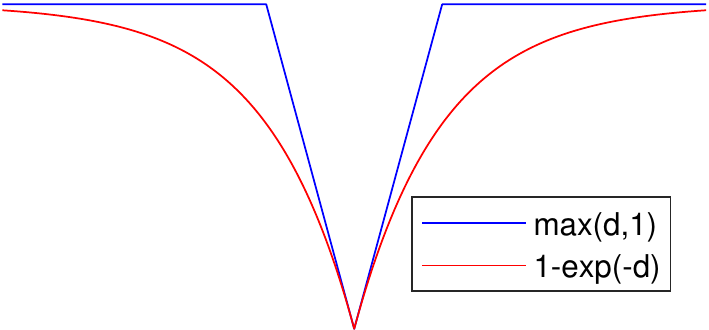}
 \caption{The cost $-g_i(k)$ as a function of $d = \| f^0(i) - f^1(i-k) \|_1$ in our model is similar to robust costs $\max(d, \tau)$ previously used to better handle occlusions~\cite{KZ-stereo}.\label{fig:unary-cost}}
\end{figure} 

\section{Implementation Details}\label{sec:implementation}
\begin{figure}[t]
    \centering
    \includegraphics[width=0.99\columnwidth]{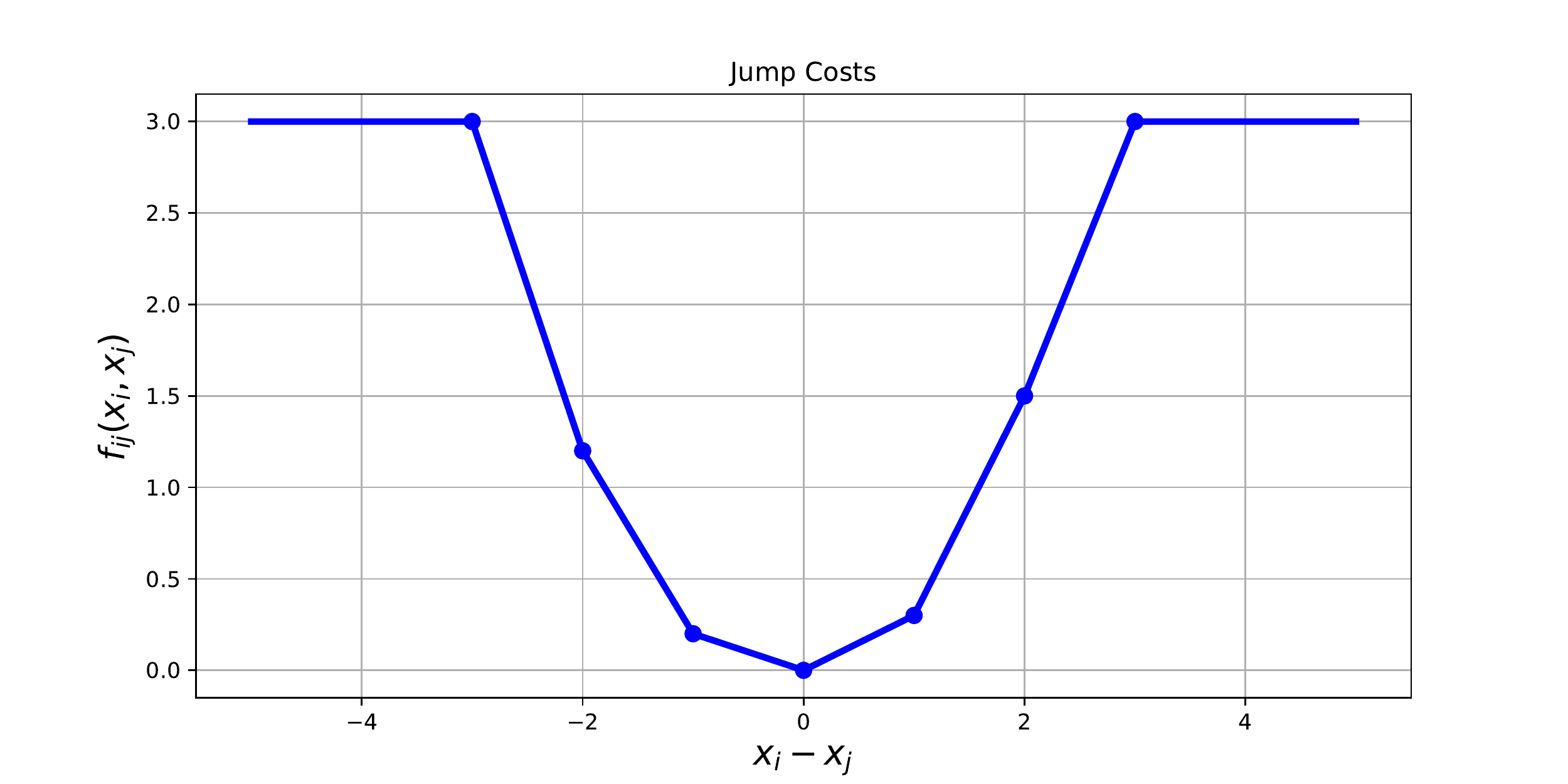}
    \caption{Robust penalty function. Similar as the $P1,P2$ model in SGM, but with one additional learnable step. We allow to learn this function asymmetrically, because positive occlusions appear only on left-sided object boundaries.}
    \label{fig:jumps}
\end{figure}

We implemented our model in PyTorch\footnote{\url{https://pytorch.org}} and
the core of the BP-layer as a highly efficient CUDA kernel. 
For geometrical problems such as stereo and optical flow, we use a truncated compatibility function (see~\cref{fig:jumps}). 
This is allows us to decrease the asymptotic runtime for $K$ labels to $\mathcal{O}(K)$ and makes very efficient inference and training possible.
For semantic segmentation we want to learn the full compatibility matrix.
Nevertheless, since we learn the cost from any source label to any target label, the runtime is $\mathcal{O}(K^2)$ and thus quadratic in the number of labels. 
The practical impact on the runtime can be seen in~\cref{tab:stereoAblation,tab:ssAblation:20}.

In our optimized CUDA implementation we utilize the following parallelization: 
All chains of the same direction as well as the chains in the opposing directions can be processed in parallel. Furthermore, the message-passing also parallelizes over the labels. For an image of size $N{\times}N$, assuming that the number of disparities also grows as $K=O(N)$, our implementation achieves parallelism of $O(N^2)$ while requiring sequential processing $O(N)$, which is an acceptable scaling with the image size. The backprop operation of the DP, has the same level of parallelism, which is important for large-scale learning. These implementations are connected as extensions to PyTorch, which allows them to be used in any computation graphs.
In order to increase numerical accuracy, we also normalize the messages by subtracting the maximum over all labels on each step of DP. This does not affect the output beliefs, as the normalization cancels in the $\softmax$ operation.

We trained the model with the Adam optimizer \cite{kingma2014adam} with a learning rate of $3\cdot10^{-3}$.
We always start with a pre-training for 300k iterations on large scale synthetic data for stereo and optical flow to get a good initialization for our model. 
Finally, we fine-tune the pre-trained models on the target dataset for 1000 epochs using a learn-rate of $10^{-5}$.

\subsection{Runtime Analysis}
We give a brief comparison of the runtime of the proposed BP-Layer and 3D convolutions here.
Compared to other networks such as \cite{zhang2019ga,kendall2017,Chang_2018_CVPR} we completely avoid the usage of the very costly 3D convolution layers. 
3D convolution layers have a runtime of $\mathcal{O}(MNKCP^3)$ while our proposed BP-Layer has a runtime of $\mathcal{O}(MNK)$, where $M$ and $N$ are the width and the height of the image, $K$ is the number of disparities, $C$ is the number of feature channels and $P$ is the size of the 3D kernel.
Although \citet{zhang2019ga} have a similar runtime of their SGA Layer, they still use 15 3D conv layers with 48 feature volumes in every layer in their full model which is very expensive. Note that their LGA Layer also operates on a 4D input, \ie on multiple 3D feature volumes, where in difference we use only one 3D volume in all stereo experiments.
\citet{kendall2017,Chang_2018_CVPR} use 19 and 25 3D conv layers, respectively.
In difference, as our ablation study in the main paper shows, we are on-par with these methods on several metrics. Furthermore, our method is the only method which is also able to achieve high quality results on the difficult Middlebury 2014 benchmark. 

\subsection{Model Architecture}
\cref{tab:featurenet} shows our very lightweight architecture which we use for feature extraction.
We actually maintain two copies of this networks with non-shared parameters. 
The first one is used as the feature network for matching and the second one is the feature network for predicting the pairwise jump-scores. 
\cref{fig:unary-cost,fig:jumps} show the functions used for unary costs and pairwise costs respectively. 
Note, that both functions are robust due to the truncation.

On every hierarchical level we add one convolution layer to map the features to pixel-wise descriptors used for matching and to pixel-wise jump-scores respectively. We denote the convolutions as ``convD\{0,1,2\}'' and ``convS\{0,1,2\}'', where D stands for disparity and S for scores.
The highest resolution is here level 0 and the lowest resolution is level 2 in our setting. 
In the last group in \cref{tab:stereoModel} we show the hierarchical inference block. 
We apply our BP-Layer on the score-volume with the coarsest scale, \ie level 2, upsample the result trilinearly and combine it with SAD matching  from the next level.
We apply this procedure until we get a regularized score-volume on the finest level, \ie level 0.

Note that the resolutions given in \cref{tab:featurenet,tab:stereoModel} are relative to the input image size.
We use with a factor 2 bilinearly downsampled images as the input to our feature networks in all experiments but Kitti. 
In Kitti we do all computations on the full-size images directly.

\label{ssec:unetArchitecture}
\begin{table}[t]
  \centering
  \renewcommand{\arraystretch}{1.3}
  \setlength{\tabcolsep}{2pt}
  \small
  \begin{tabular}{ccccc}
    \toprule
    \textbf{Layer} & \textbf{KS} & \textbf{Resolution} & \textbf{Channels} & \textbf{Input} \\
    \midrule
    conv00 & 3 & $W \times H ~/ ~ W \times H$ & $3 ~/ ~16$ & Image \\
    conv01 & 3 & $W \times H ~/ ~ W \times H$ & $16~/~16$ & conv00 \\
    pool0 & 2 & $W \times H ~/ ~ \frac{W}{2} \times \frac{H}{2}$ & $16~/~16$ & conv01\\
    \midrule
    conv10 & 3 & $\frac{W}{2} \times \frac{H}{2} ~/~ \frac{W}{2} \times \frac{H}{2}$ & $16 ~/~32$ & pool0 \\
    conv11 & 3 & $\frac{W}{2} \times \frac{H}{2} ~/~ \frac{W}{2} \times \frac{H}{2}$ & $32 ~/~32$ & conv10 \\
    pool1 & 2 & $\frac{W}{2} \times \frac{H}{2} ~/~ \frac{W}{4} \times \frac{H}{4}$ & $32~/~32$ & conv10\\
    \midrule
    conv20 & 3 & $\frac{W}{4} \times \frac{H}{4} ~/~ \frac{W}{4} \times \frac{H}{4}$ & $32 ~/~64$ & pool1 \\
    conv21 & 3 & $\frac{W}{4} \times \frac{H}{4} ~/~ \frac{W}{4} \times \frac{H}{4}$ & $64 ~/~64$ & conv20 \\
    \midrule
    bilin1 & - & $\frac{W}{4} \times \frac{H}{4} ~/~ \frac{W}{2} \times \frac{H}{2}$ & $64 ~/~64$ & conv21 \\
    conv12 & 3 & $\frac{W}{2} \times \frac{H}{2} ~/~ \frac{W}{2} \times \frac{H}{2}$ & $96~/~32$ & \{bilin1, conv11\} \\
    conv13 & 3 & $\frac{W}{2} \times \frac{H}{2} ~/~ \frac{W}{2} \times \frac{H}{2}$ & $32 ~/~32$ & conv12 \\
    \midrule
    bilin0 & - & $\frac{W}{2} \times \frac{H}{2} ~/~ W \times H$ & $32 ~/~32$ & conv12 \\
    conv02 & 3 & $W \times H ~/~ W\times H$ & $48 ~/~32$ & \{bilin0, conv01\} \\
    conv03 & 3 & $W \times H ~/~ W \times H$ & $32 ~/~32$ & conv02 \\
    \bottomrule
  \end{tabular}
  \caption{Detailed Architecture of our UNet for feature extraction.}
  \label{tab:featurenet}
\end{table}

\begin{table}[t]
\centering
  \renewcommand{\arraystretch}{1.3}
  \setlength{\tabcolsep}{2pt}
  \small
\begin{tabular}{ccccc}
\toprule
\textbf{Layer} & \textbf{KS} & \textbf{Resolution} & \textbf{Channels} & \textbf{Input} \\
\midrule
convD2 & 3 & $\frac{W}{4} \times \frac{W}{4} ~/~ \frac{H}{4} \times \frac{H}{4}$ & $64 ~/~ 32$ & conv21 \\
convD1 & 3 & $\frac{W}{2} \times \frac{W}{2} ~/~ \frac{H}{2} \times \frac{H}{2}$ & $32 ~/~ 32$ & conv13 \\
convD0 & 3 & $W \times W ~/~ H \times H$ & $32 ~/~ 32$ & conv03 \\
\midrule
convS2 & 3 & $\frac{W}{4} \times \frac{W}{4} ~/~ \frac{H}{4} \times \frac{H}{4}$ & $64 ~/~ 32$ & conv21 \\
convS1 & 3 & $\frac{W}{2} \times \frac{W}{2} ~/~ \frac{H}{2} \times \frac{H}{2}$ & $32 ~/~ 32$ & conv13 \\
convS0 & 3 & $W \times W ~/~ H \times H$ & $32 ~/~ 32$ & conv03 \\
\midrule
sad2 & - & $\frac{W}{4} \times \frac{W}{4} ~/~ \frac{H}{4} \times \frac{H}{4}$ & $32 ~/~ \frac{D}{4}$ & convD2\_0, convD2\_1 \\
sad1 & - & $\frac{W}{2} \times \frac{W}{2} ~/~ \frac{H}{2} \times \frac{H}{2}$ & $32 ~/~ \frac{D}{2}$ & convD1\_0, convD1\_1 \\
sad0 & - & $W \times W ~/~ H\times H$ & $32 ~/~ D$ & convD0\_0, convD0\_1 \\
\midrule
BP2 & - & $\frac{W}{4} \times \frac{W}{4} ~/~ \frac{H}{4} \times \frac{H}{4}$ & $\frac{D}{4} ~/~ \frac{D}{4}$ & sad2, convS2 \\
BP2\_up & - & $\frac{W}{4} \times \frac{W}{4} ~/~ \frac{H}{2} \times \frac{H}{2}$ & $\frac{D}{4} ~/~ \frac{D}{2}$  & BP2 \\ 
BP1 & - & $\frac{W}{2} \times \frac{W}{2} ~/~ \frac{H}{2} \times \frac{H}{2}$ & $\frac{D}{2} ~/~ \frac{D}{2}$ & sad1 + BP2\_up, convS1 \\
BP1\_up & - & $\frac{W}{2} \times \frac{W}{2} ~/~ W \times H$ & $\frac{D}{2} ~/~ D$  & BP1 \\ 
BP0 & - & $W \times W ~/~ H \times H$ & $D ~/~ D$ & sad0 + BP1\_up, convS0 \\
\bottomrule
\end{tabular}
\caption{Hierarchical BP inference block. We add convolutions to map the features from the feature net to appropriate input to our BP-Layer. The plus operation '+' indicates a point-wise addition.}
\label{tab:stereoModel}
\end{table}

\section{More Details on Experiments}
\label{sec:app:moreexamples}
Due to the limited space in the main paper, we add additional qualitative results and interpretations of these results here. In the following sections, we discuss additional experiments which were performed for Stereo, Semantic Segmentation and Optical Flow.

\begin{figure*}[t]
    \begin{tikzpicture}[overlay, anchor=center]
     \small
     \node[rotate=90] at (0.15,1.2) {Input};
     \node[rotate=90] at (0.15,-1.2) {WTA};
     \node[rotate=90] at (0.15,-3.6) {BP~(NLL)};
     \node[rotate=90] at (0.15,-6) {BP+MS~(NLL)};
     \node[rotate=90] at (0.15,-8.4) {BP+MS~(H)};
     \node[rotate=90] at (0.15,-10.8) {BP+MS+ref~(H)};
     \node[rotate=90] at (0.15,-13.2) {GT};
     \node[rotate=90] at (0.15,-15.6) {Edges};
    \end{tikzpicture}
    \hfill
    \includegraphics[width=0.24\textwidth]{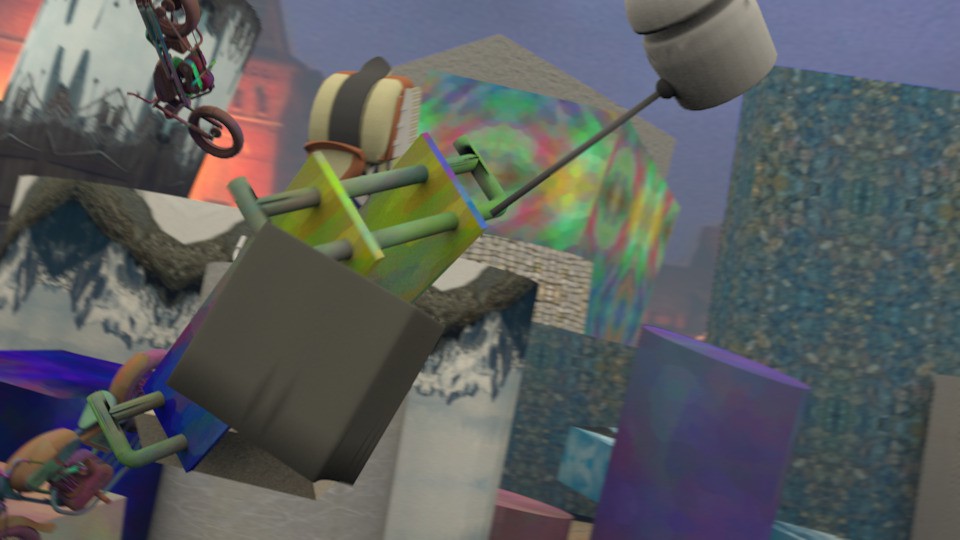}
    \includegraphics[width=0.24\textwidth]{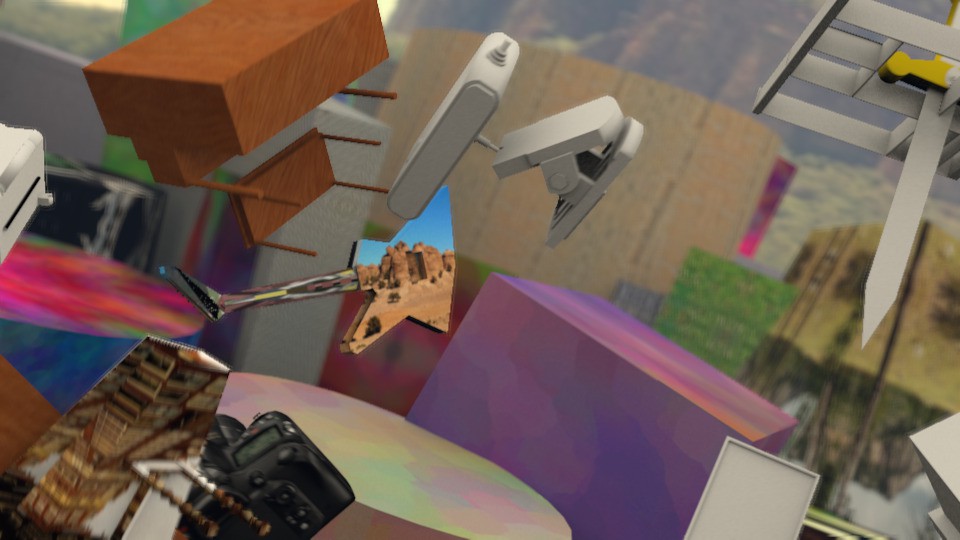}
    \includegraphics[width=0.24\textwidth]{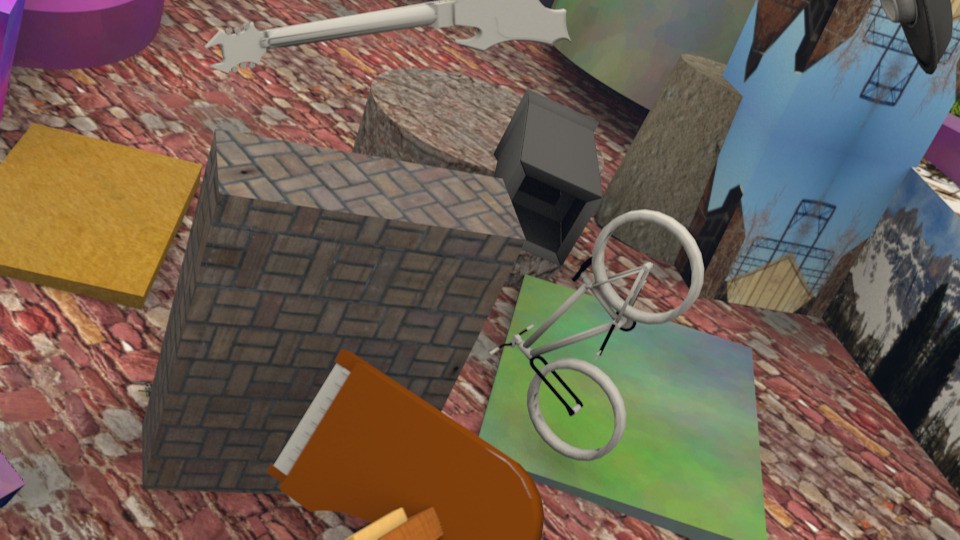}
    \includegraphics[width=0.24\textwidth]{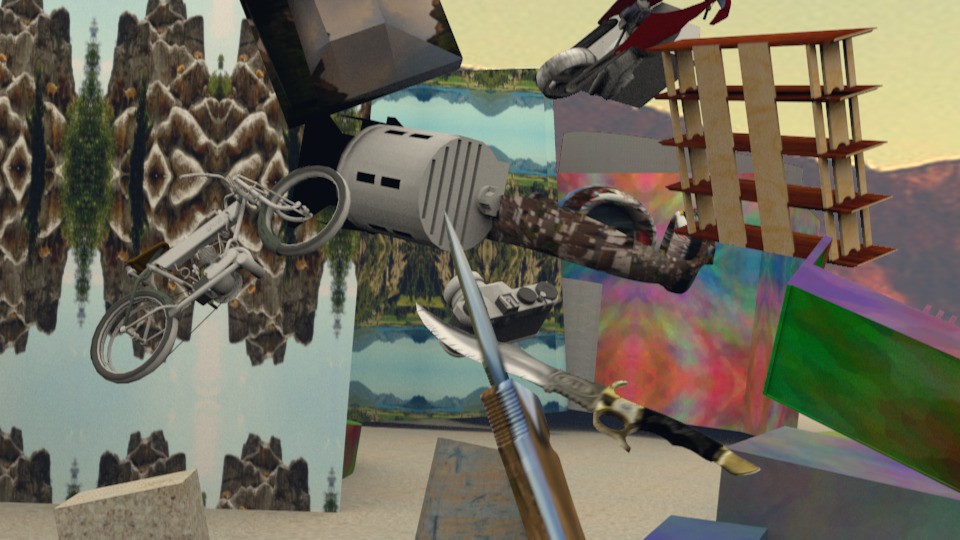}
    
    \hfill
    \includegraphics[width=0.24\textwidth]{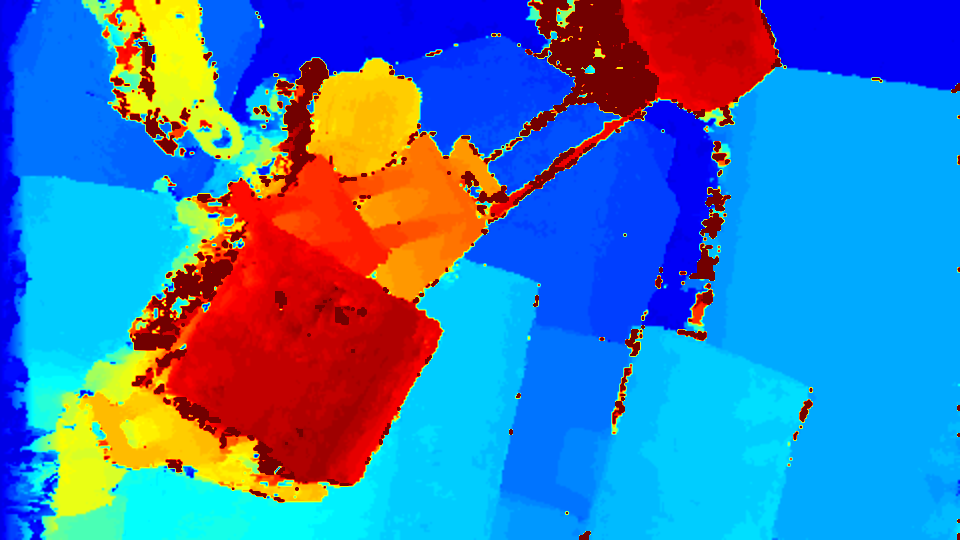}
    \includegraphics[width=0.24\textwidth]{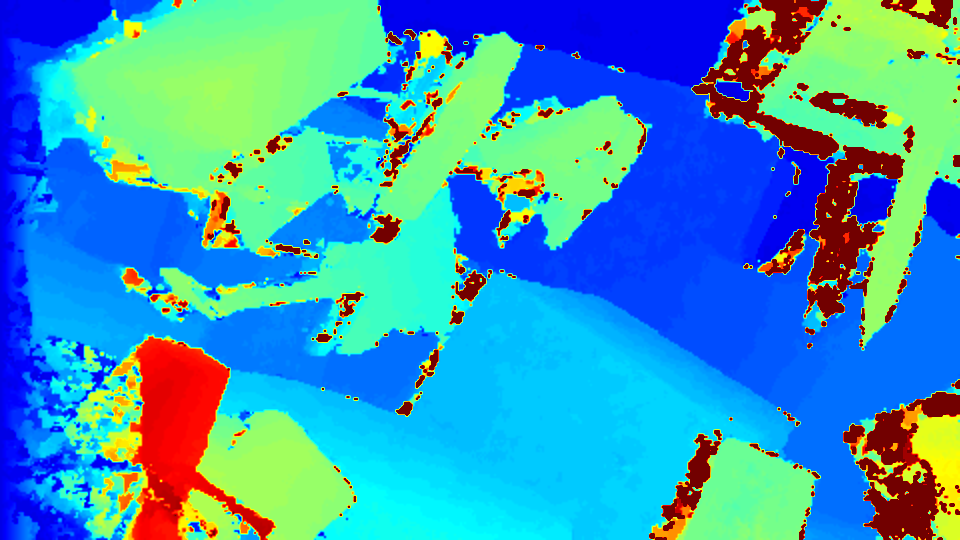}
    \includegraphics[width=0.24\textwidth]{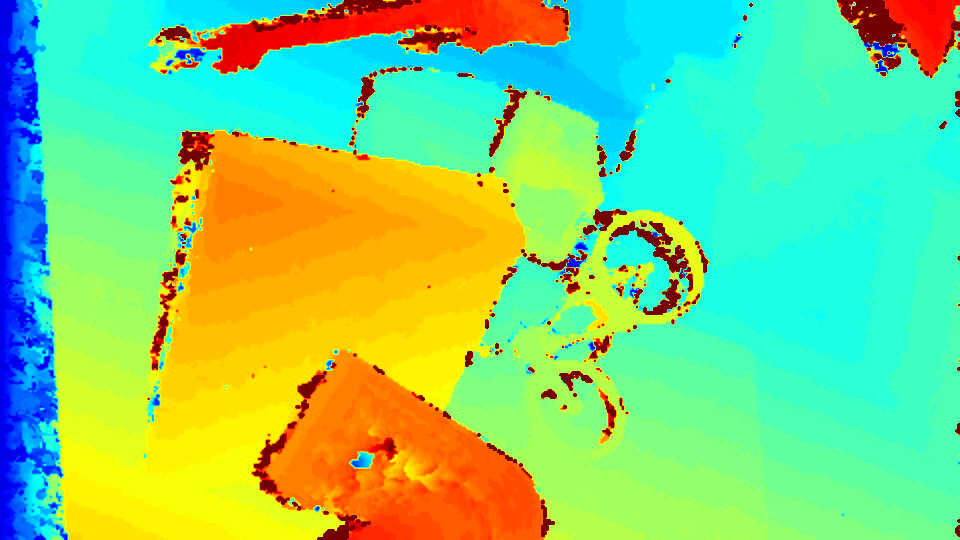}
    \includegraphics[width=0.24\textwidth]{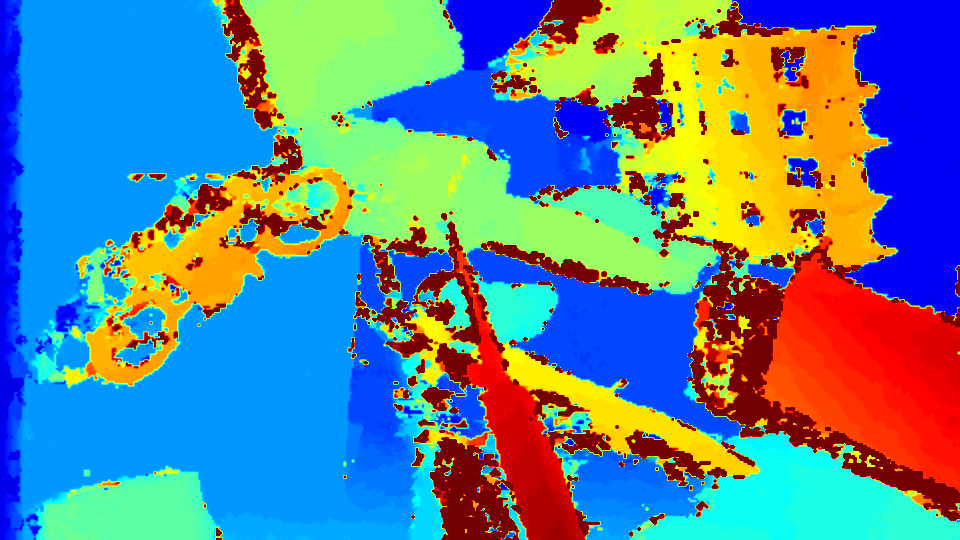}
    
    \hfill
    \includegraphics[width=0.24\textwidth]{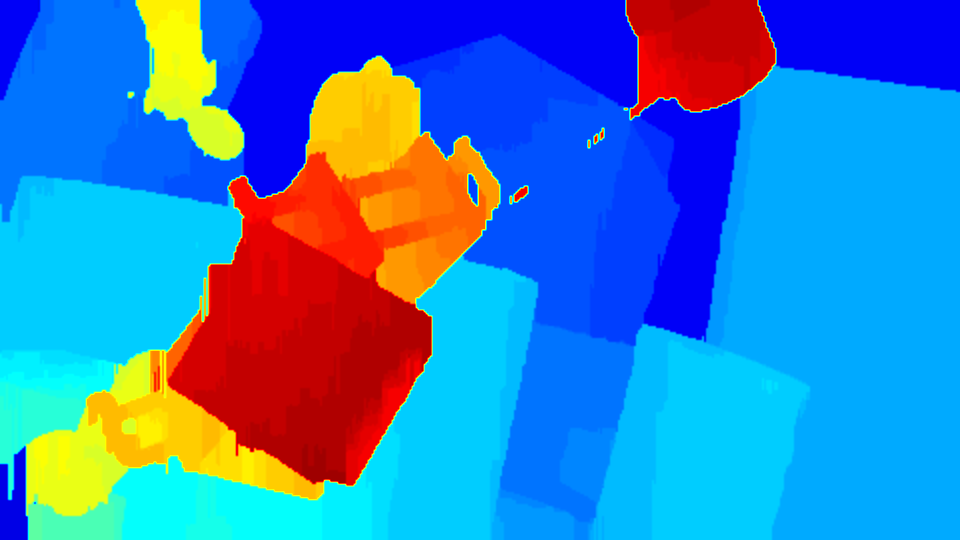}
    \includegraphics[width=0.24\textwidth]{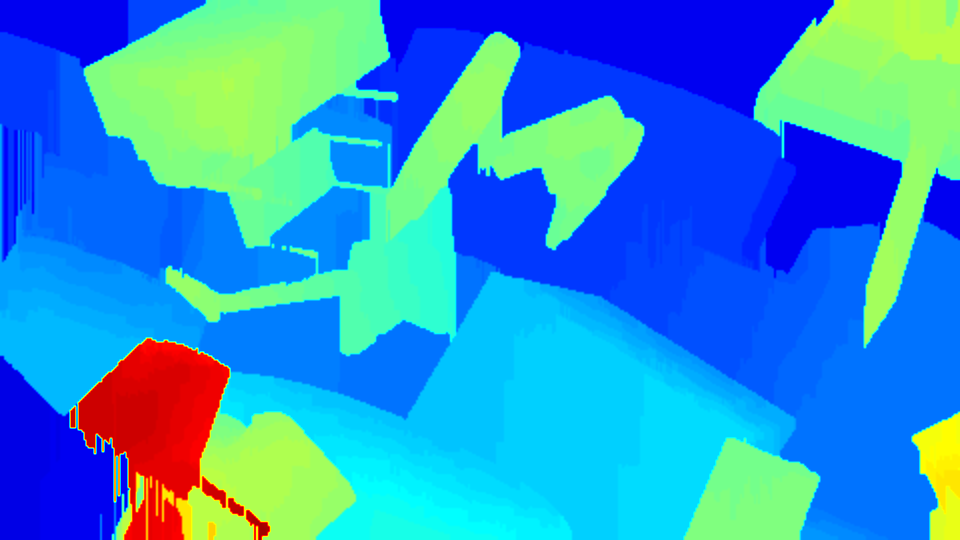}
    \includegraphics[width=0.24\textwidth]{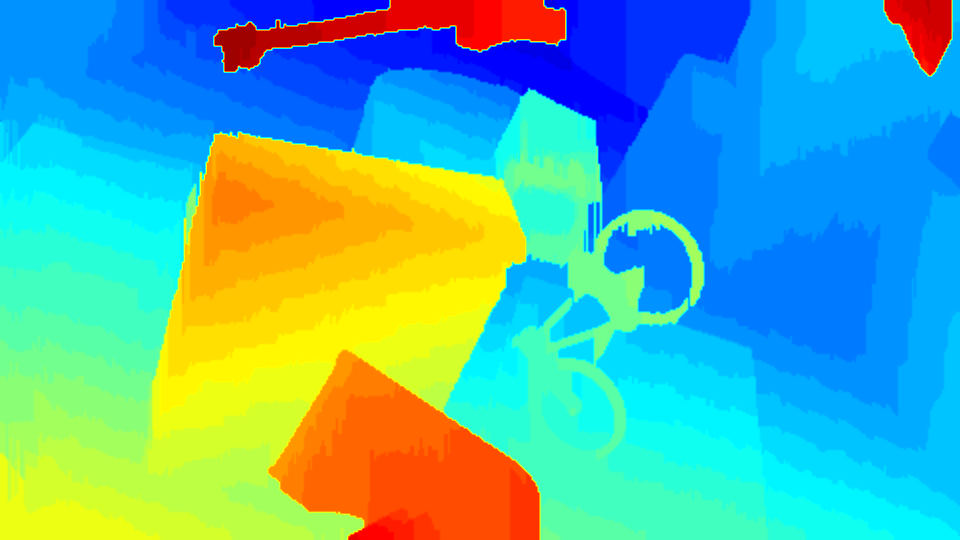}
    \includegraphics[width=0.24\textwidth]{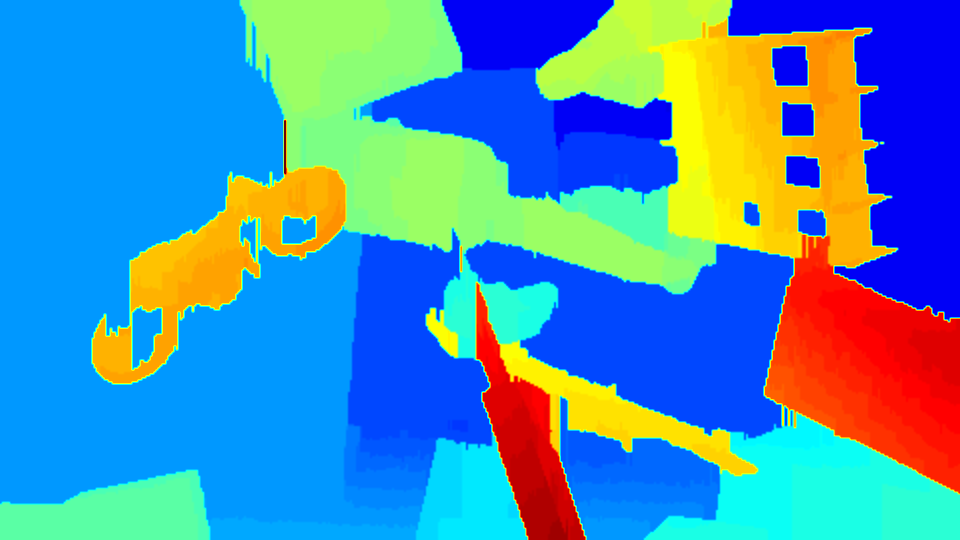}
    
    \hfill
    \includegraphics[width=0.24\textwidth]{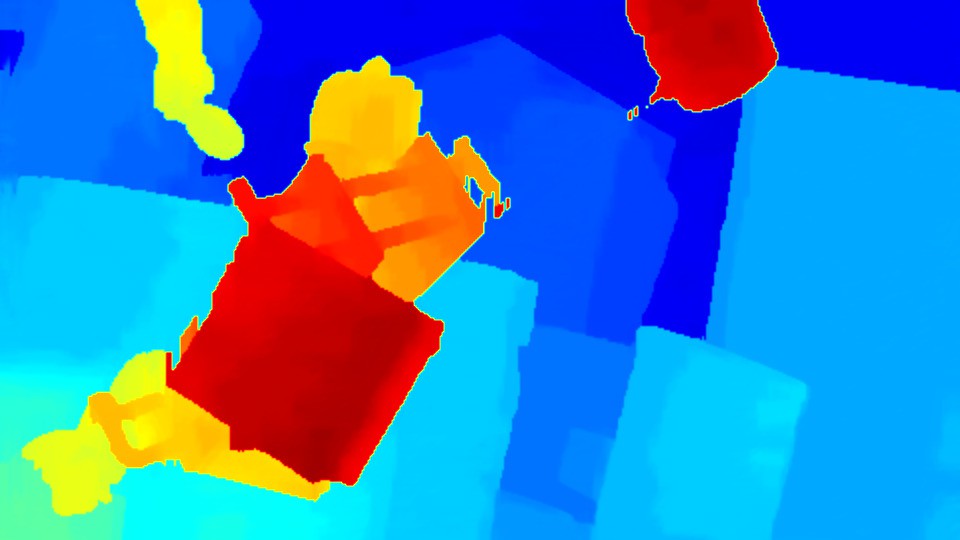}
    \includegraphics[width=0.24\textwidth]{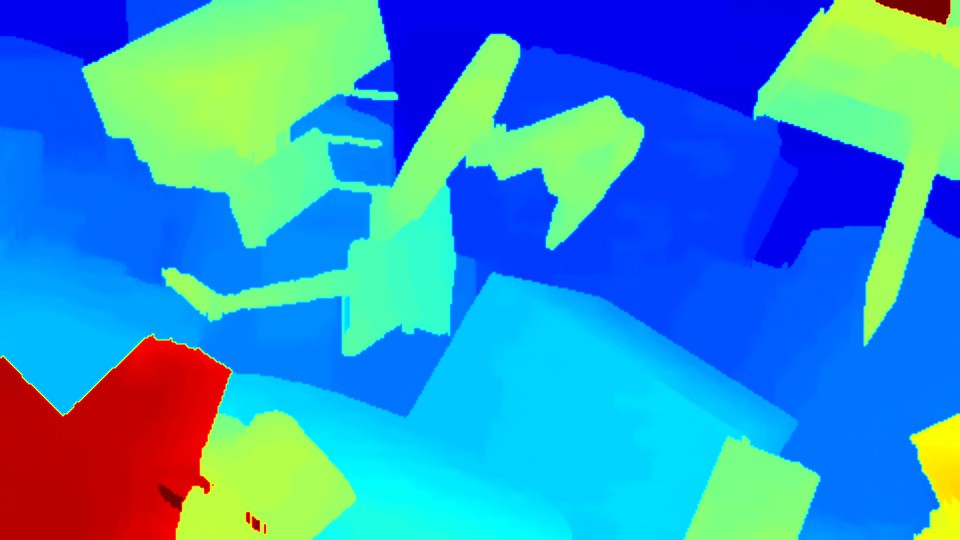}
    \includegraphics[width=0.24\textwidth]{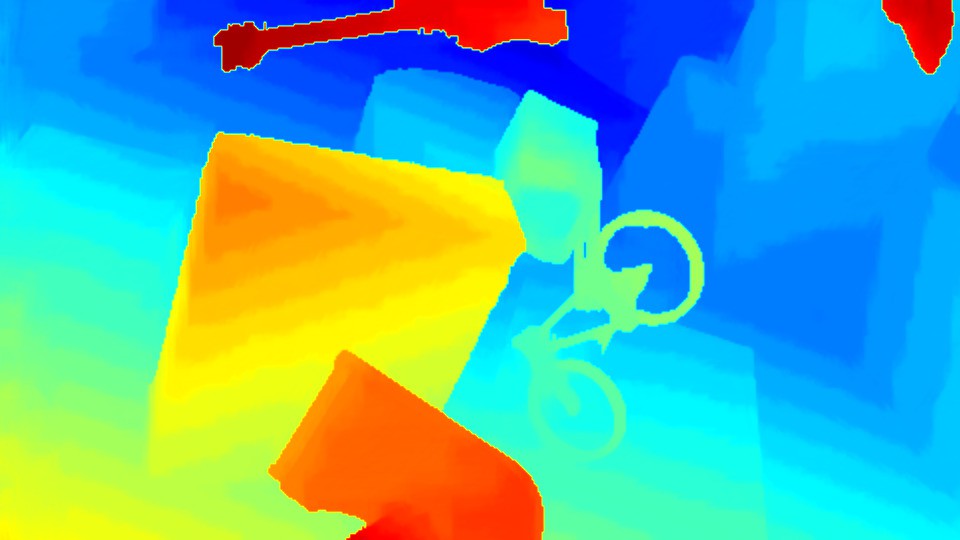}
    \includegraphics[width=0.24\textwidth]{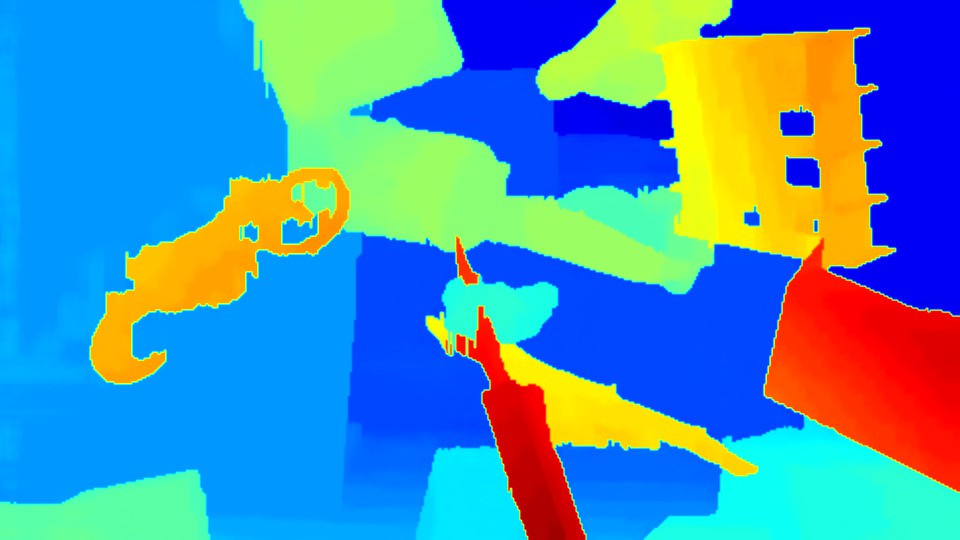}
    
    \hfill
    \includegraphics[width=0.24\textwidth]{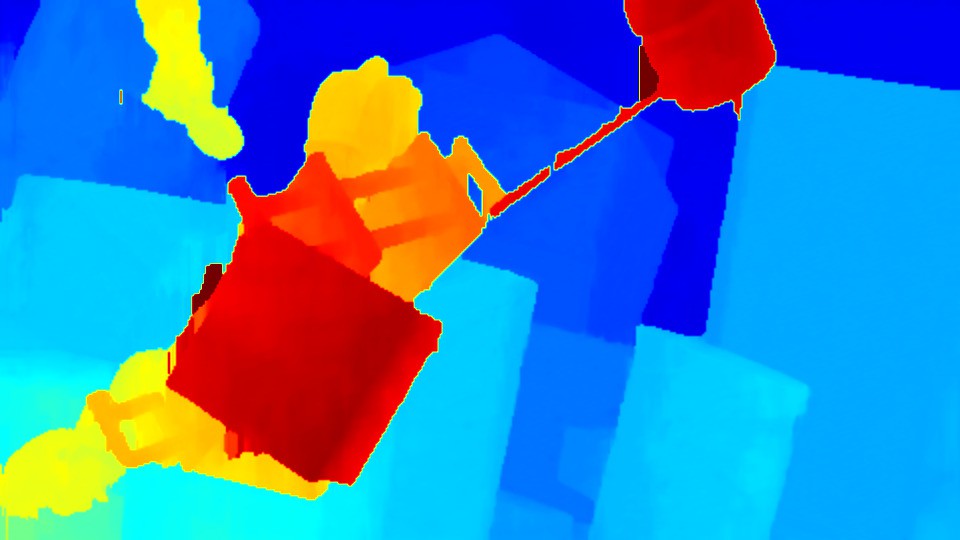}
    \includegraphics[width=0.24\textwidth]{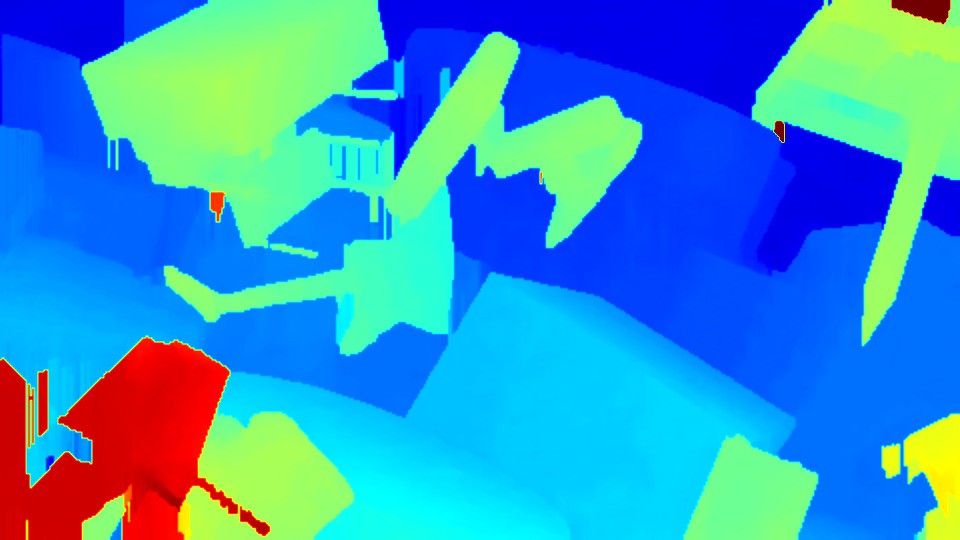}
    \includegraphics[width=0.24\textwidth]{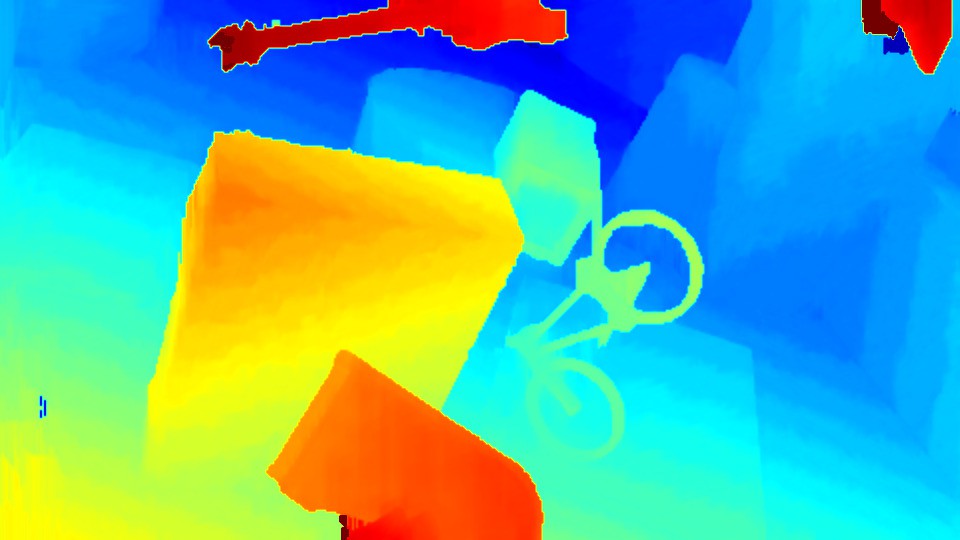}
    \includegraphics[width=0.24\textwidth]{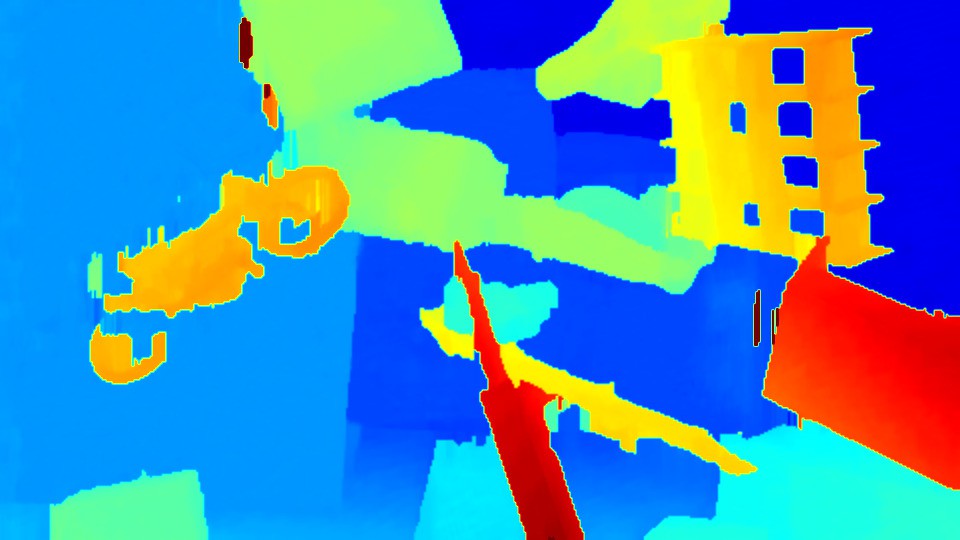}
    
    \hfill
    \includegraphics[width=0.24\textwidth]{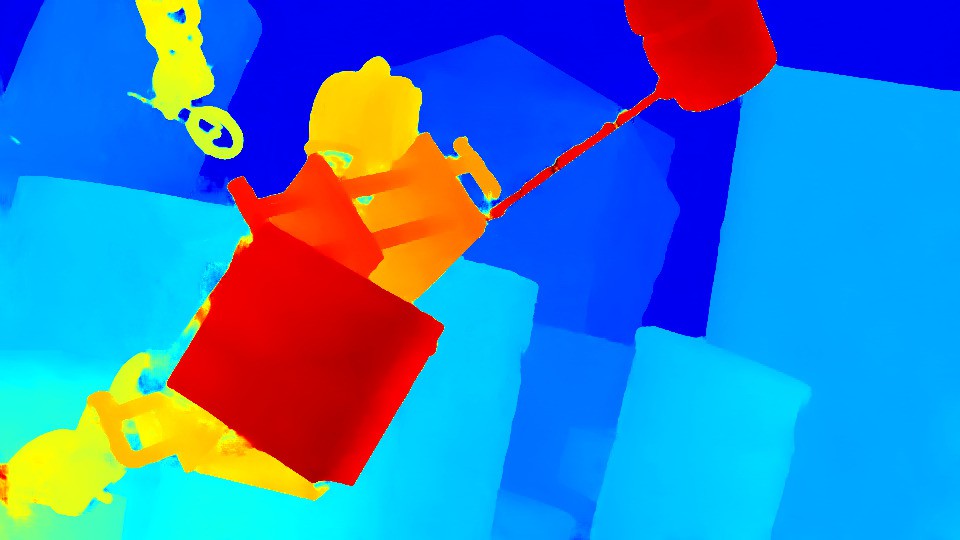}
    \includegraphics[width=0.24\textwidth]{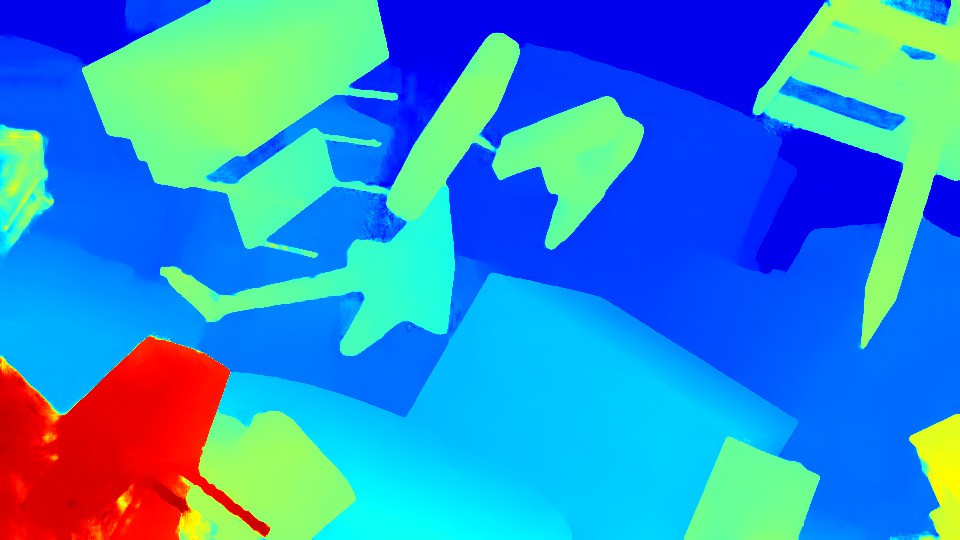}
    \includegraphics[width=0.24\textwidth]{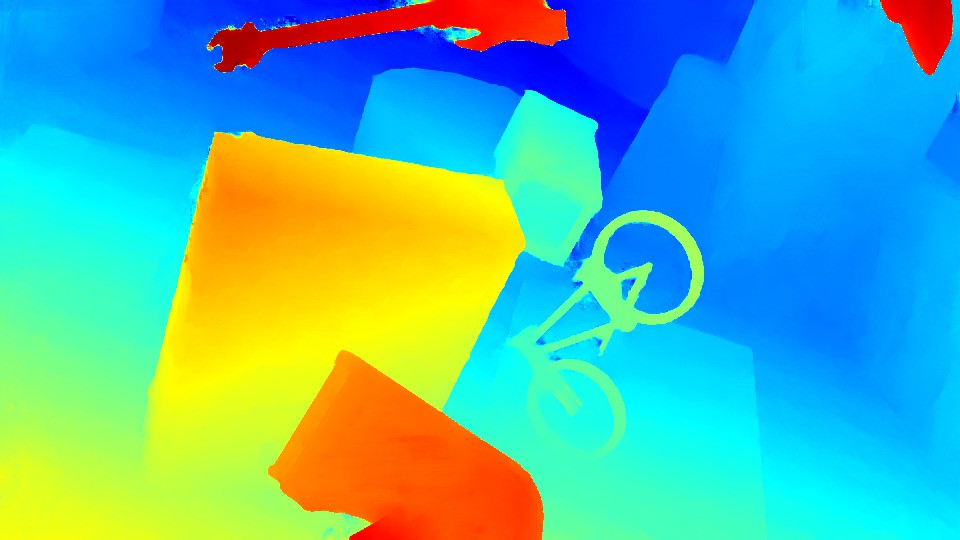}
    \includegraphics[width=0.24\textwidth]{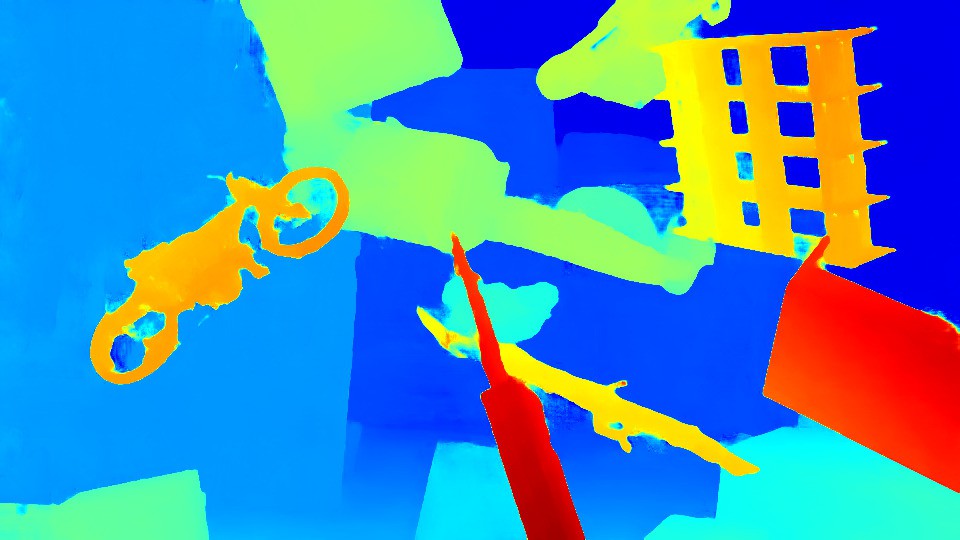}

    \hfill
    \includegraphics[width=0.24\textwidth]{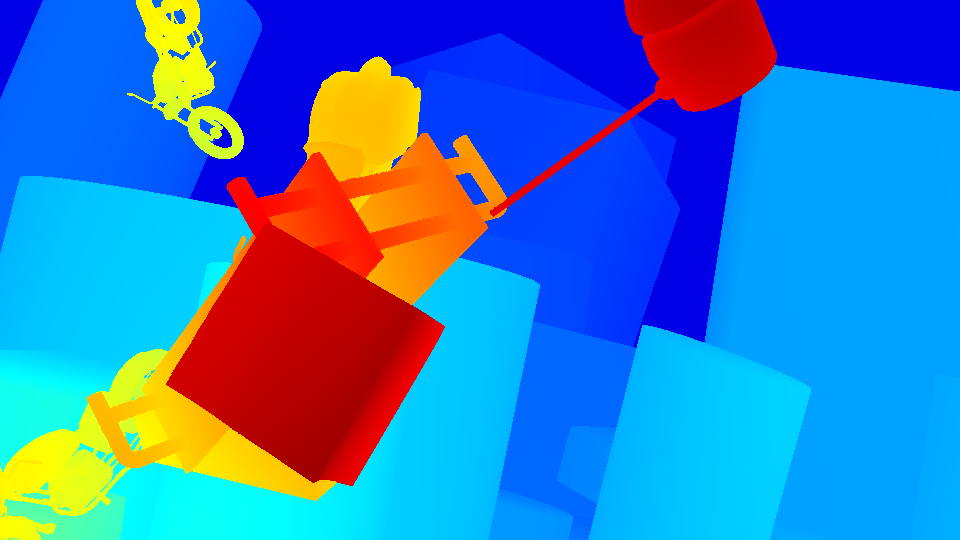}
    \includegraphics[width=0.24\textwidth]{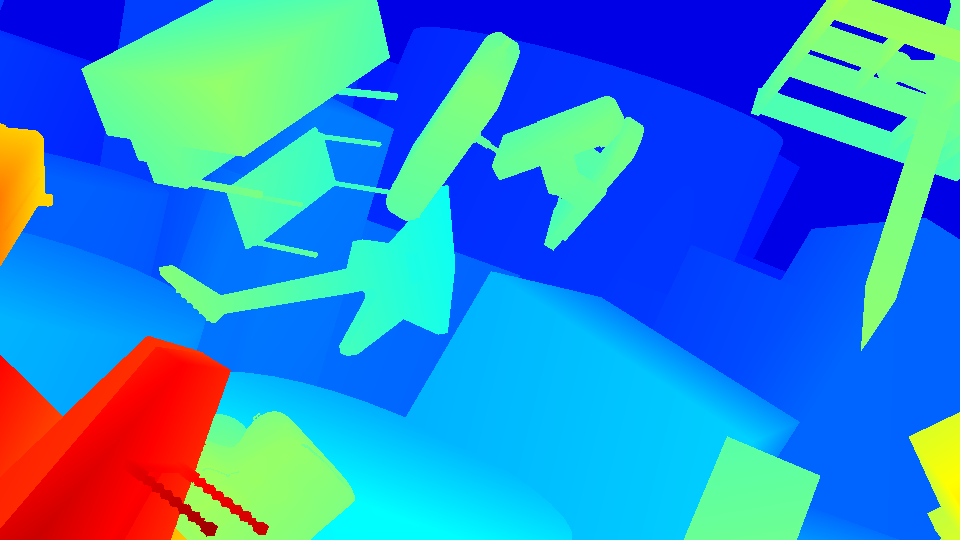}
    \includegraphics[width=0.24\textwidth]{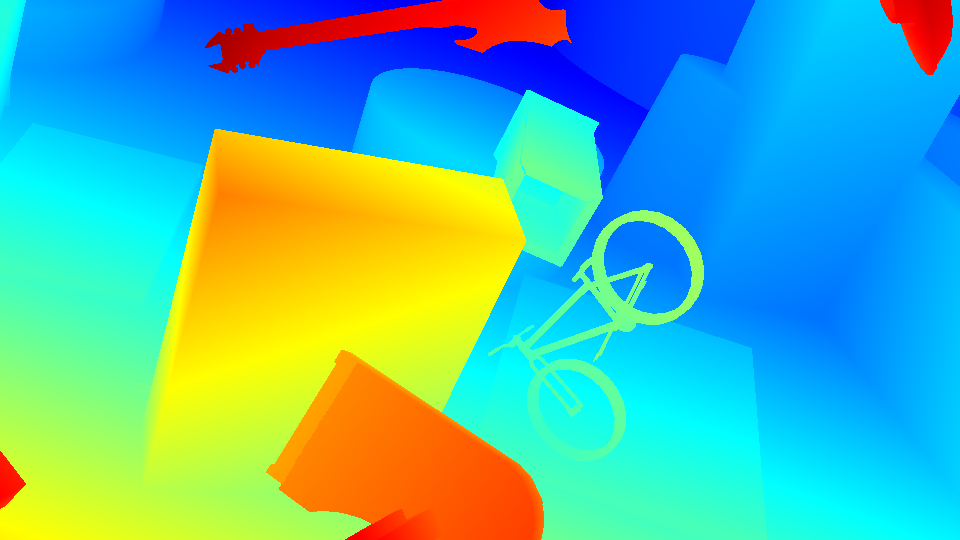}
    \includegraphics[width=0.24\textwidth]{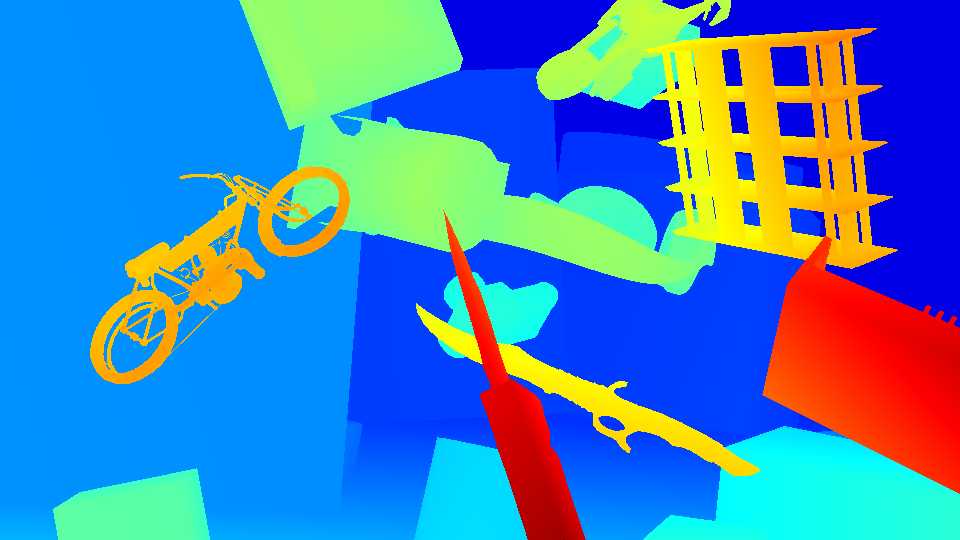}

    \hfill
    \includegraphics[width=0.24\textwidth]{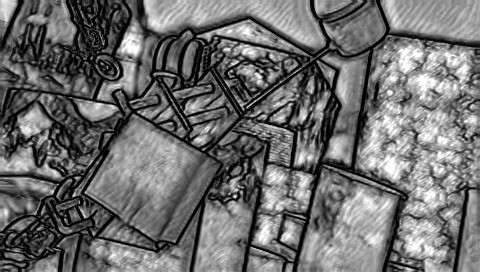}
    \includegraphics[width=0.24\textwidth]{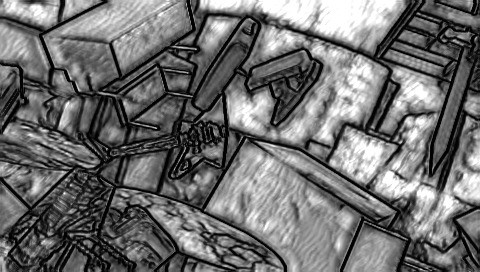}
    \includegraphics[width=0.24\textwidth]{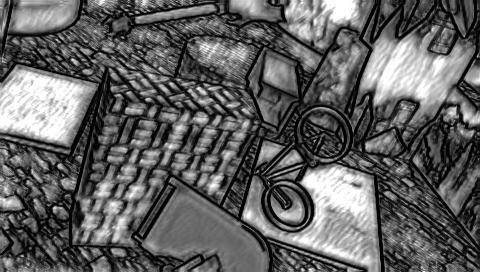}
    \includegraphics[width=0.24\textwidth]{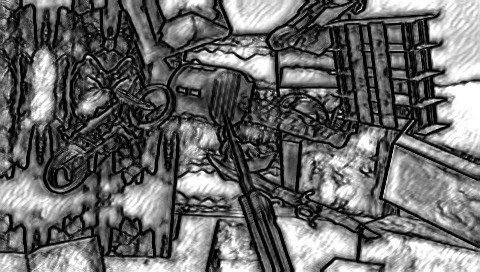}
    
    \caption{Visual ablation study. The methods are the same as used in the quantitative ablation study (\cref{tab:stereoAblation}) and compared from top to bottom. The last row shows the learned jump-costs of BP+MS+Ref~(H) used in our BP-Layer, where black=low cost and white=high cost. The edge images are easily interpretable. We can see that the object edges and depth discontinuities are precisely captured.}
    \vspace{3em}
    \label{fig:abltionExamples}
\end{figure*}
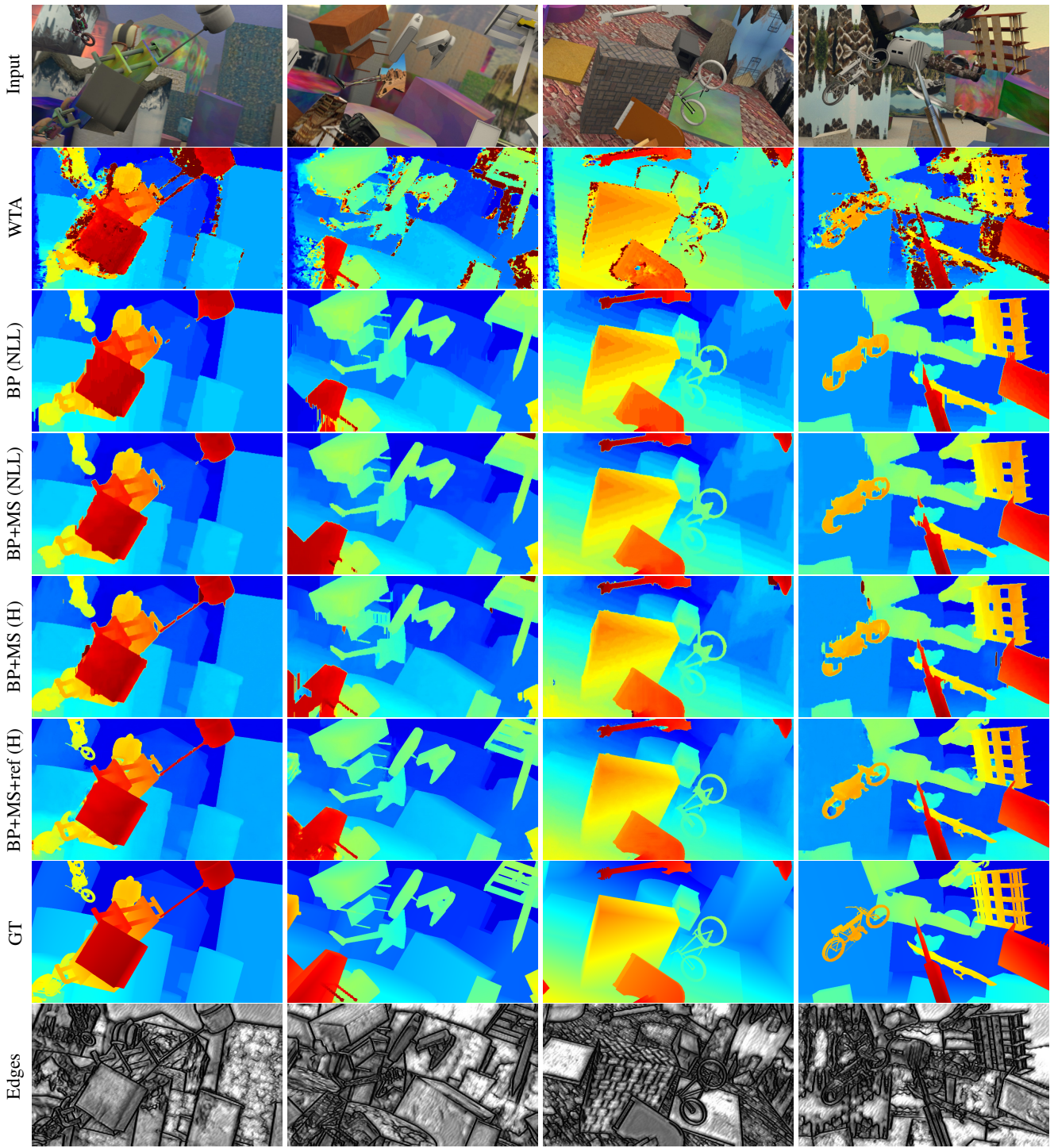

\begin{figure*}
  \centering
  \includegraphics[width=0.3\textwidth]{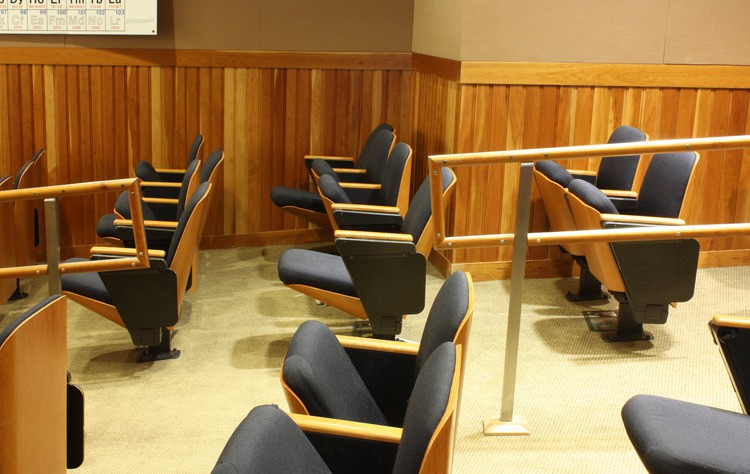}
  \includegraphics[width=0.3\textwidth]{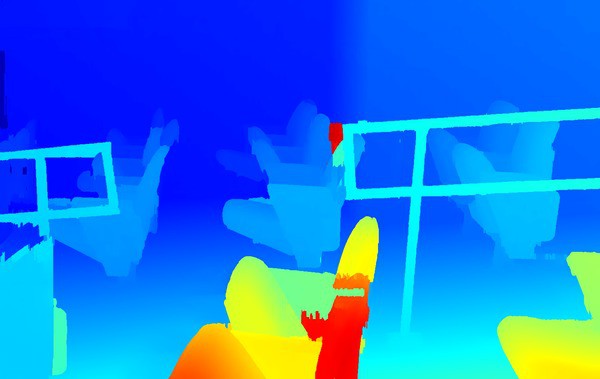}
  \includegraphics[width=0.3\textwidth]{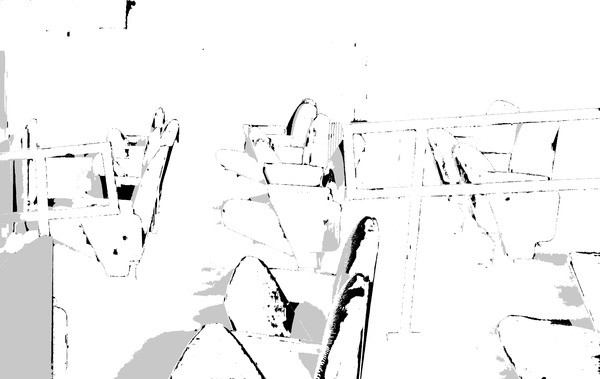}
  \includegraphics[width=0.3\textwidth]{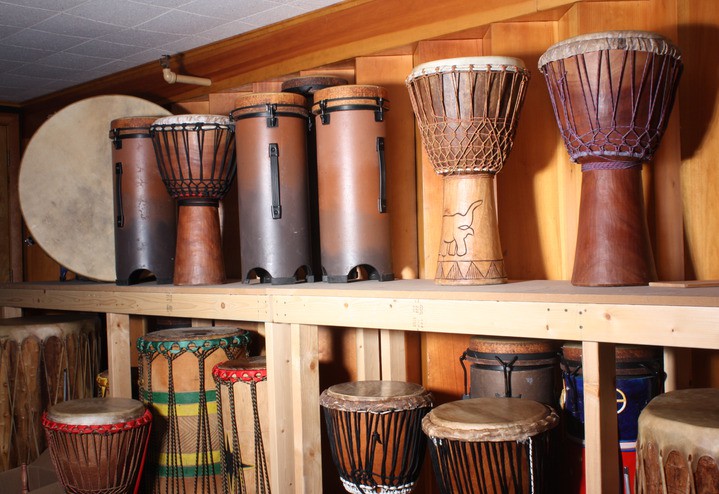}
  \includegraphics[width=0.3\textwidth]{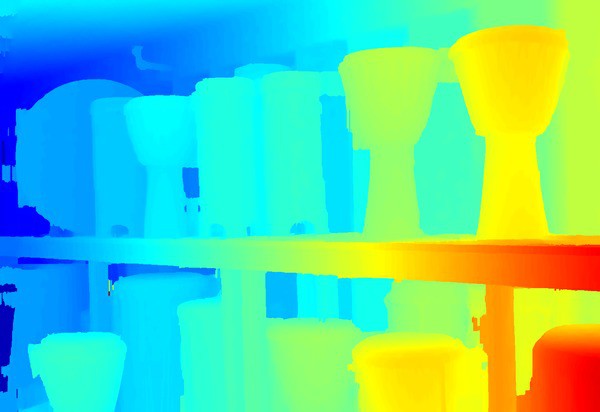}
  \includegraphics[width=0.3\textwidth]{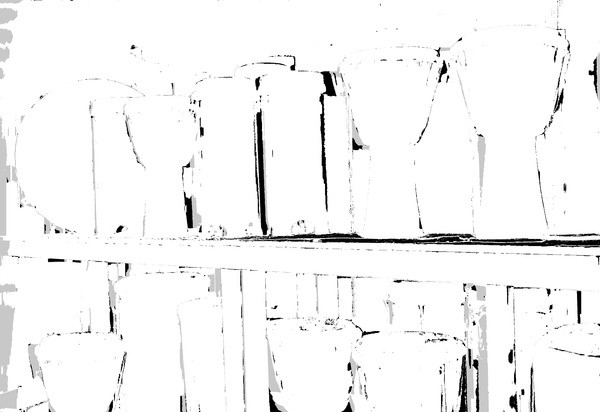}
  \includegraphics[width=0.3\textwidth]{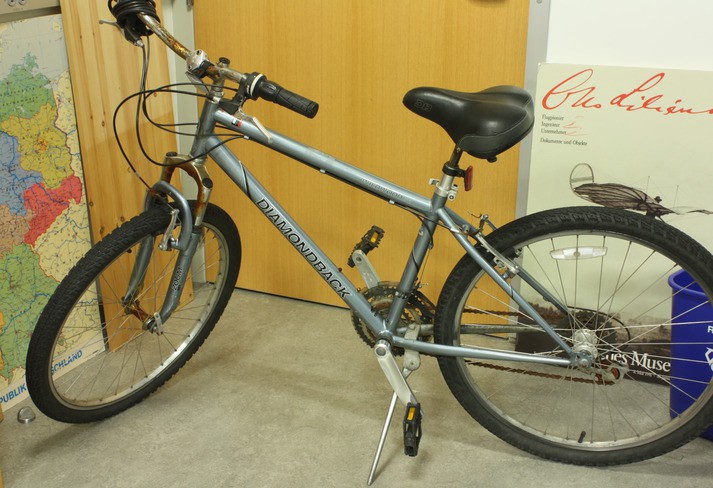}
  \includegraphics[width=0.3\textwidth]{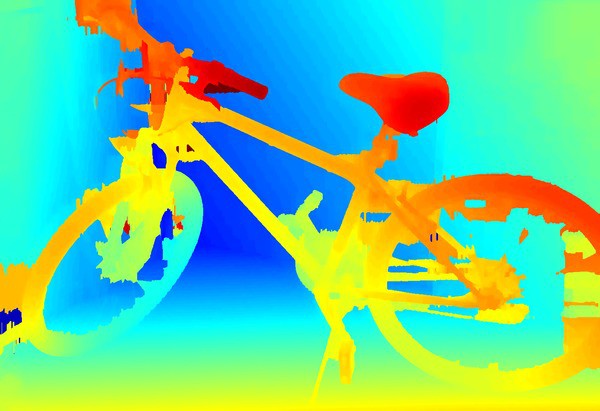}
  \includegraphics[width=0.3\textwidth]{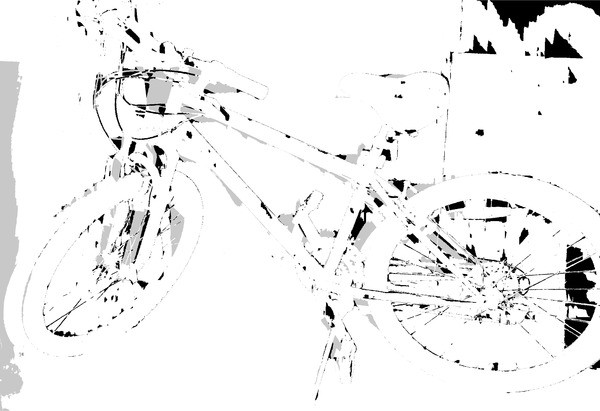}
  \caption{Qualitative results on the Middlebury 2014 test set. Left: color coded disparity map, right error map, where white = correct disparity, black = wrong disparity and gray = occluded area.}
  \label{fig:moreExperiments:mb}
\end{figure*}

\begin{figure*}
  \centering
  \includegraphics[width=0.55\textwidth]{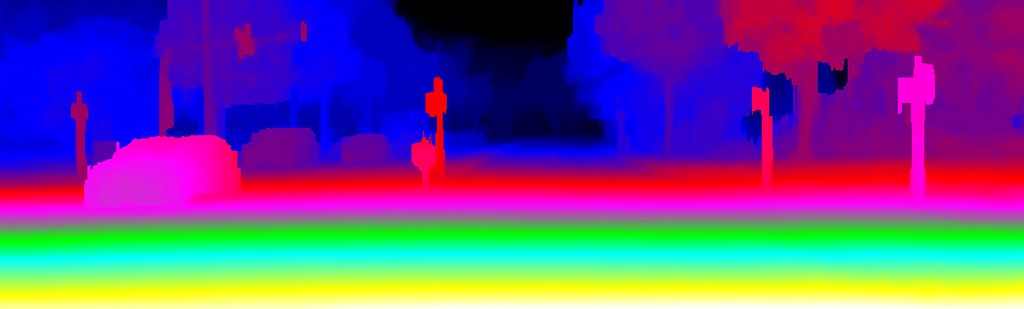}
  \begin{minipage}[b][][c]{0.27\textwidth}
  \includegraphics[width=\textwidth]{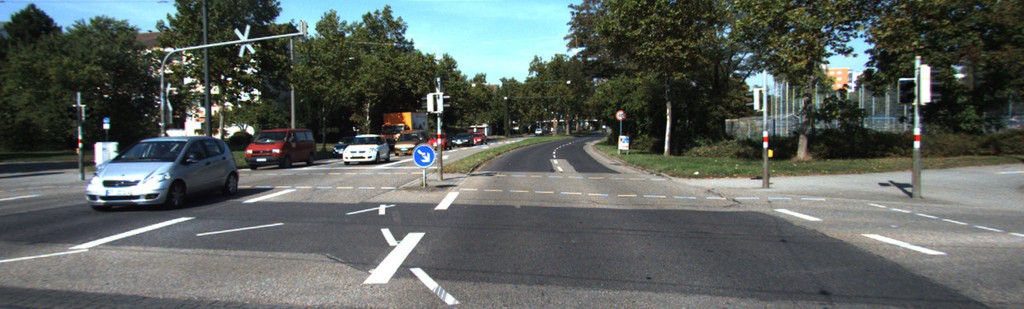}\\
  \includegraphics[width=\textwidth]{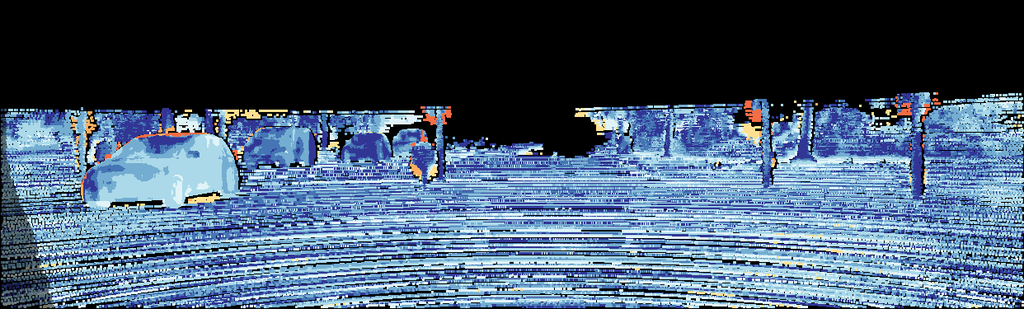}
  \end{minipage}
  
  \includegraphics[width=0.55\textwidth]{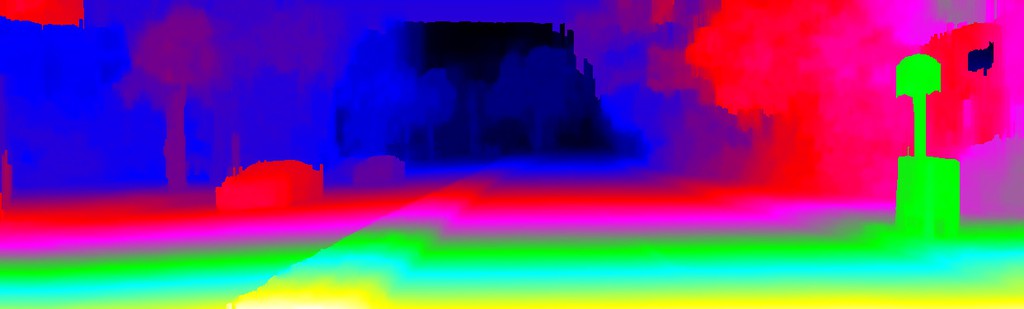}
  \begin{minipage}[b][][c]{0.27\textwidth}
  \includegraphics[width=\textwidth]{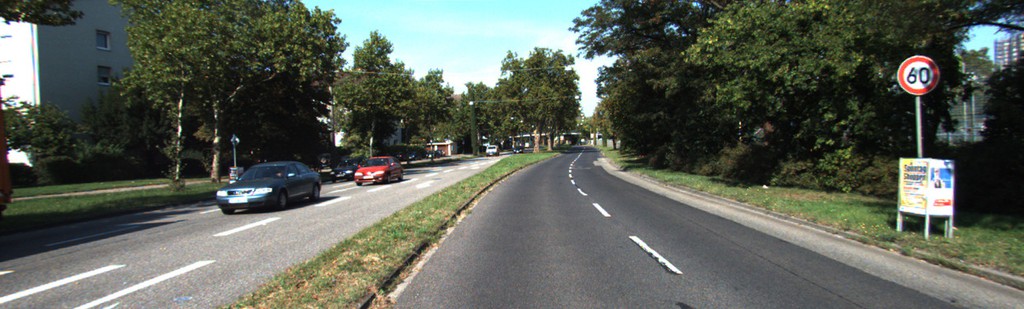}\\
  \includegraphics[width=\textwidth]{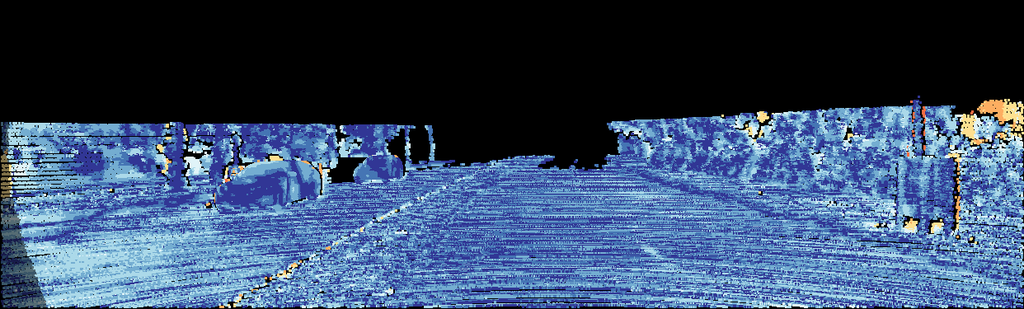}
  \end{minipage}
  
  \includegraphics[width=0.55\textwidth]{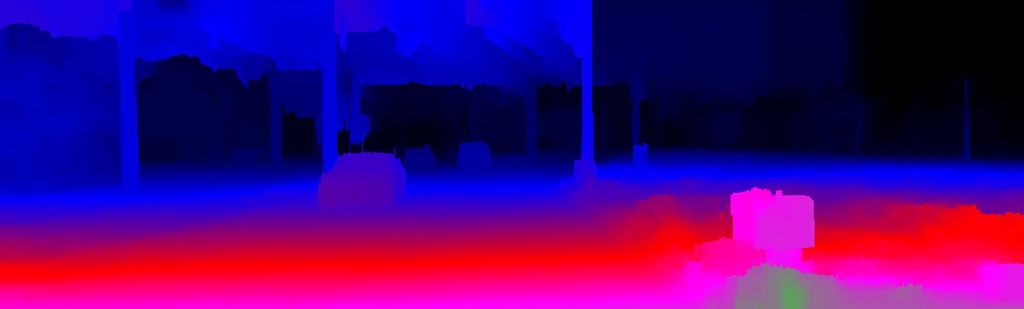}
  \begin{minipage}[b][][c]{0.27\textwidth}
  \includegraphics[width=\textwidth]{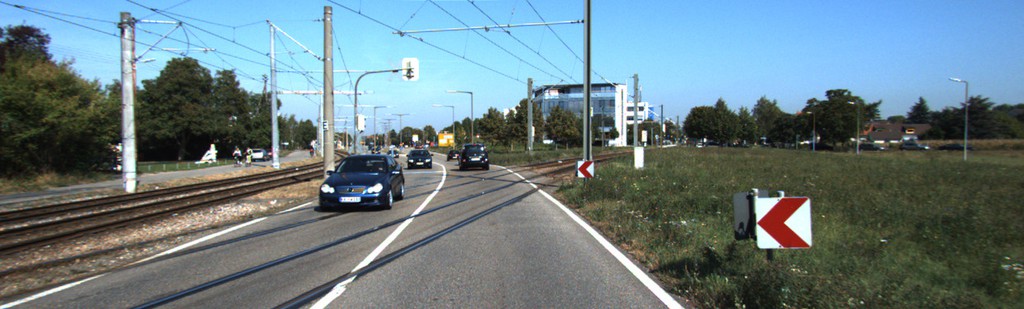}\\
  \includegraphics[width=\textwidth]{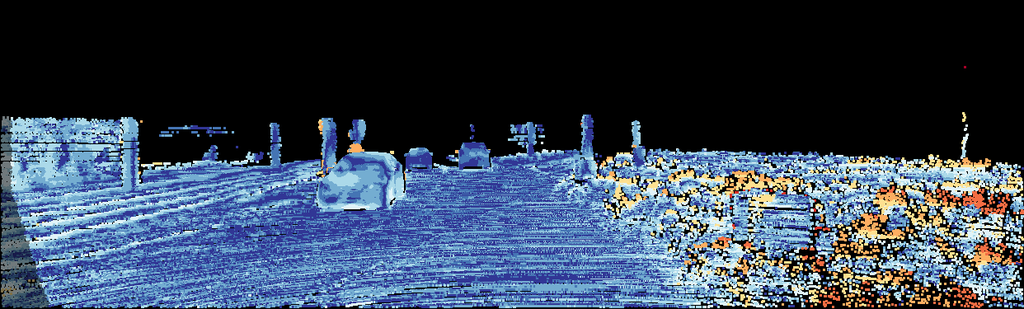}
  \end{minipage}
  \caption{Kitti test set examples. The left column shows the color-coded disparity map, the right column shows on top the input image and on the bottom the official error map on the Kitti benchmark. The blue color in the error map indicates correct predictions, orange indicate wrong predictions and black is unknown. Note how our method produces high quality results also for regions where no ground-truth is available, \ie in the upper third of the images.}
  \label{fig:kitti:more}
\end{figure*}

\begin{figure*}
  \centering
  \includegraphics[width=0.49\textwidth]{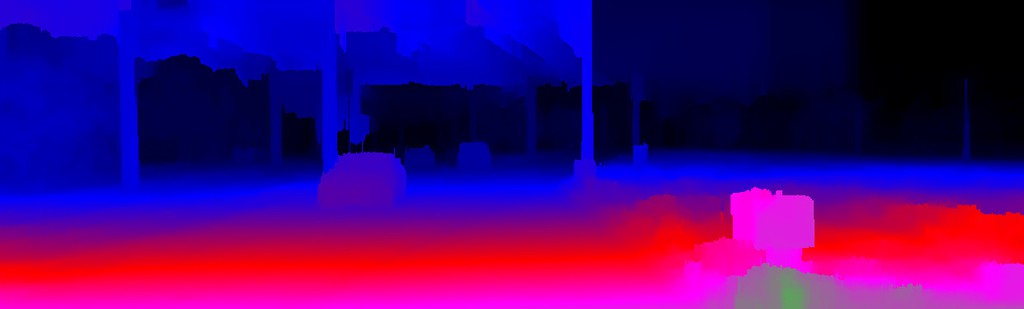}
  \includegraphics[width=0.49\textwidth]{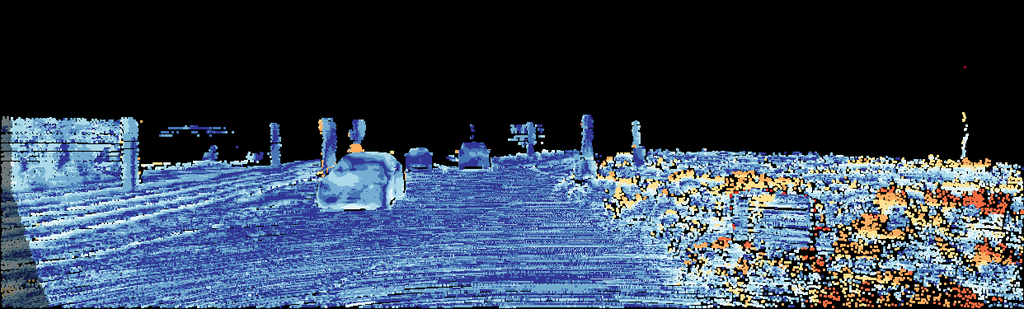}
  \includegraphics[width=0.49\textwidth]{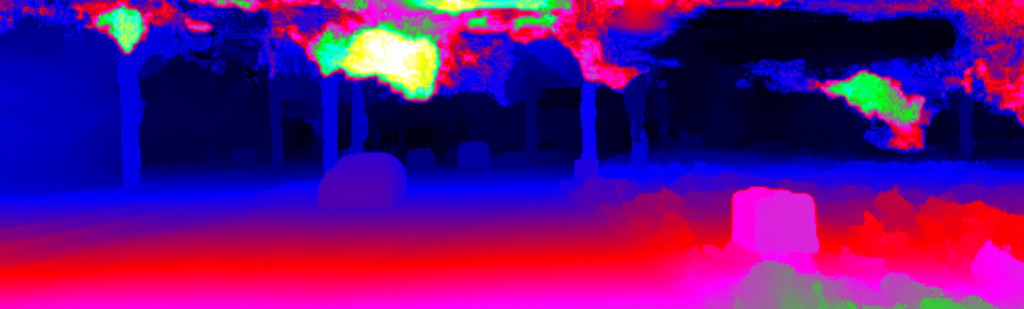}
  \includegraphics[width=0.49\textwidth]{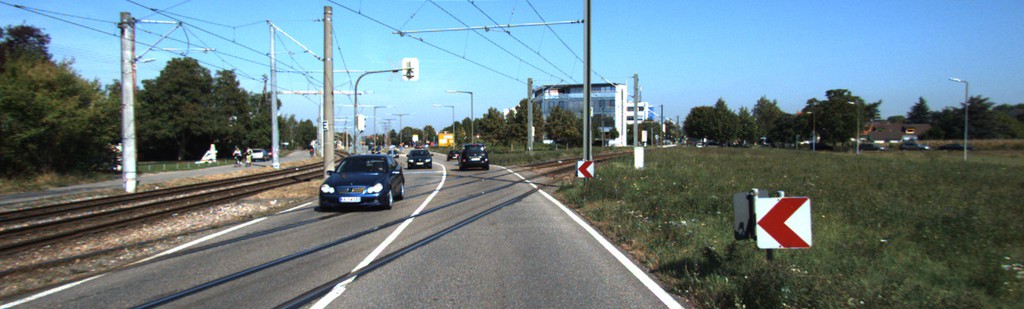}
  \includegraphics[width=0.49\textwidth]{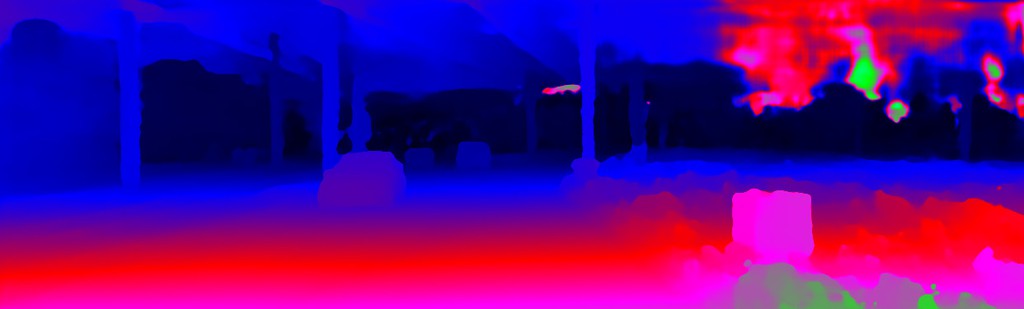}
  \includegraphics[width=0.49\textwidth]{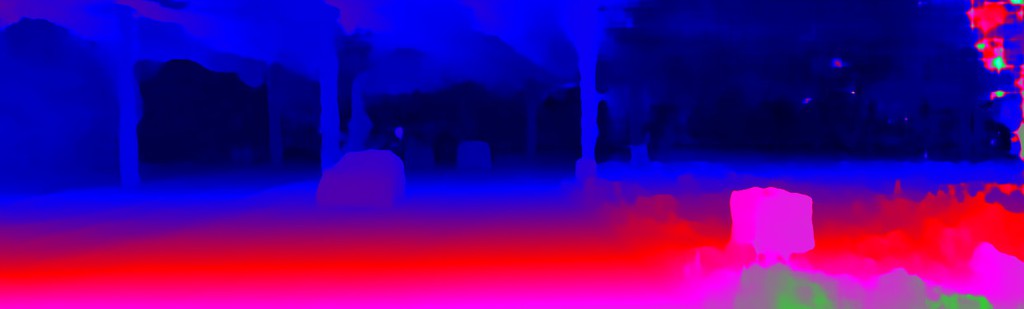}
  \begin{tikzpicture}[overlay, anchor=center]
  
  \begin{scope}[color=orange, very thick]
  \node[draw, minimum width=3cm, minimum height=1.2cm, xshift=-10.3cm, yshift=7.2cm] () {};
  
  \node[draw, minimum width=8.4cm, minimum height=1cm, xshift=-4.3cm, yshift=7.3cm] () {};
  
  \node[draw, xshift=-9.9cm, yshift=4.5cm, minimum width=2.4cm, minimum height=1.5cm] () {};
  \node[draw, xshift=-14.0cm, yshift=4.6cm, minimum width=5.5cm, minimum height=1.2cm] () {};
  
  \node[draw, minimum width=8.4cm, minimum height=1cm, xshift=-4.3cm, yshift=4.7cm] () {};
  
  \node[draw, xshift=-10.0cm, yshift=2cm, minimum width=2.7cm, minimum height=1.2cm] () {};
  
  \node[draw, xshift=-0.4cm, yshift=1.9cm, minimum width=0.6cm, minimum height=1.2cm] () {};
  \end{scope}
  \end{tikzpicture}
  \caption{Comparison with other methods on the Kitti benchmark. Top row: LBPS (ours), LBPS error visualization. Middle row: HD3 Stereo~\cite{hd3_cvpr}, input image. Bottom row: GANet~\cite{zhang2019ga}, PSMNet~\cite{Chang_2018_CVPR}. One can observe that LBPS shows no artifacting in regions where no ground truth is present.}
  \label{fig:kittihalo}
\end{figure*}

\subsection{Stereo}
\cref{fig:abltionExamples} shows a qualitative ablation study comparing our model variants on selected images. Note that we show here exactly the same model variants as in \cref{tab:stereoAblation}.
The visual ablation study shows interesting insights about our models:
First, the WTA result (2nd row in \cref{fig:abltionExamples}) is already a very good initialization on all matchable pixels although we use a very efficient network (\cref{tab:featurenet}) which uses only 130k parameters. 
The \bpl regularizes the WTA solution by removing  most of the artifacts, especially in occluded regions as can be seen in the 3rd row. 
However, due to the NLL loss function the discretization artifacts are visible in \eg the 3rd example from left.
The multi-scale variant adds robustness in large, untextured regions as can be seen in \eg example 1 on the gray box. 
Training with the Huber loss (row 5) enables sub-pixel accurate solutions. Note how this model captures fine details such as the bar better than the previous models.
Our final model can then be used to recover very fine details such as the spokes of the motorcycle in the first example.

\cref{fig:moreExperiments:mb,fig:kitti:more} show additional qualitative results on the Middlebury 2014 test set and the Kitti 2015 test set. 
We include the input image and the error images which are provided by the respective benchmarks. 

In \cref{fig:kittihalo} we compare our prediction with the prediction of current state-of-the-art models. While GA-Net \cite{zhang2019ga}, HD3-Stereo \cite{hd3_cvpr} and PSM-Net \cite{Chang_2018_CVPR} predict precise disparity maps for pixels with available ground-truth, they often hallucinate incorrect disparities on the other pixels. 
In contrast, our method does not seem to be affected at all by this problem and thus this indicates that our model generalizes very well also to previously unseen structures.
For a better comparison we highlighted these regions in \cref{fig:kittihalo}.

\subsection{Optical Flow}
We use the same network architectures for optical flow as for stereo. 
Thus, we have two feature nets \cref{tab:featurenet} and then apply hierarchically our BP-Layer on the cost-volumes.

Here we show here more examples on our validation set and highlight differences until we get our final model BP+MS+Ref~(H).
Therefore, \cref{fig:sintel:more} shows a visual ablation study. 
If we compare the models we see that the quality of the results increase from top to bottom. 
Thus, the components we add are also beneficial for optical flow. 
If we add our BP-Layer and use it to regularize the WTA result we can clearly see that most of the noise, mainly coming from occlusions, is gone. 
The Huber loss function and the refinement successfully predict then contiguous solutions.
Although our approach is very simplistic in comparison with current state-of-the-art models we are still able to compute high quality optical flow.

\begin{figure*}
  \begin{tikzpicture}[overlay, anchor=center]
   \small
   \node[rotate=90] at (0.15,-1) {Input};
   \node[rotate=90] at (0.15,-2.7) {WTA};
   \node[rotate=90] at (0.15,-4.5) {BP+MS(NLL)};
   \node[rotate=90] at (0.15,-6.3) {BP+MS (H)};
   \node[rotate=90] at (0.15,-8.2) {BP+MS+Ref~(H)};
   \node[rotate=90] at (0.15,-10.1) {GT};
  \end{tikzpicture}
  
  \hfill
  \includegraphics[width=0.24\textwidth]{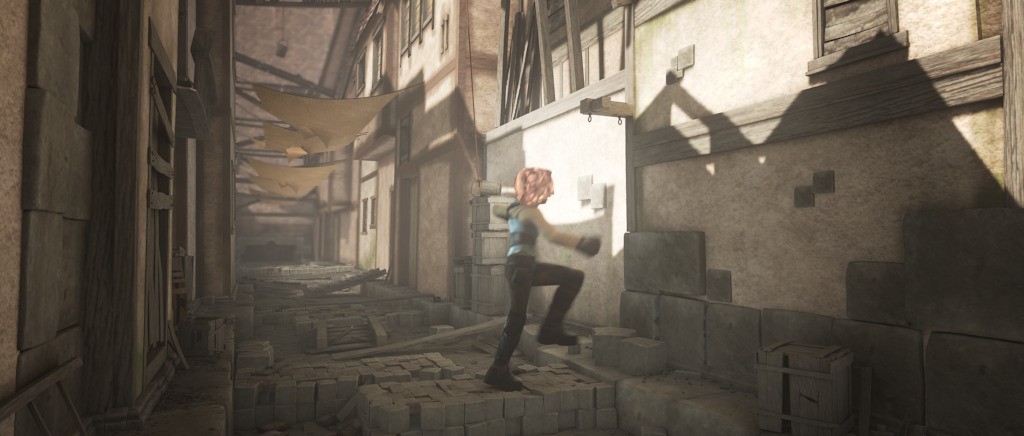}
  \includegraphics[width=0.24\textwidth]{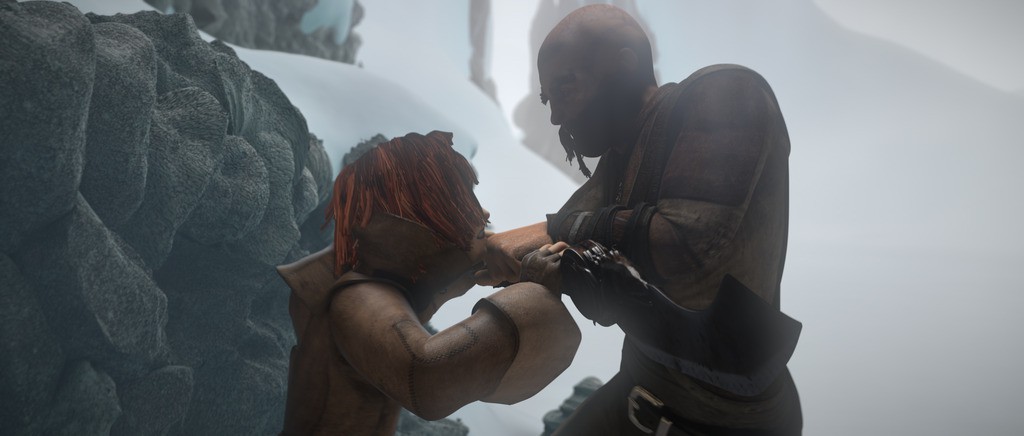}
  \includegraphics[width=0.24\textwidth]{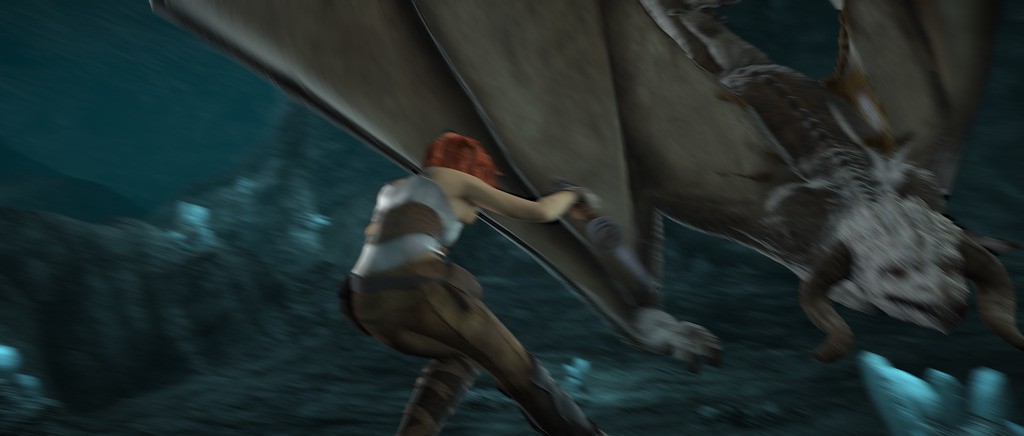}
  \includegraphics[width=0.24\textwidth]{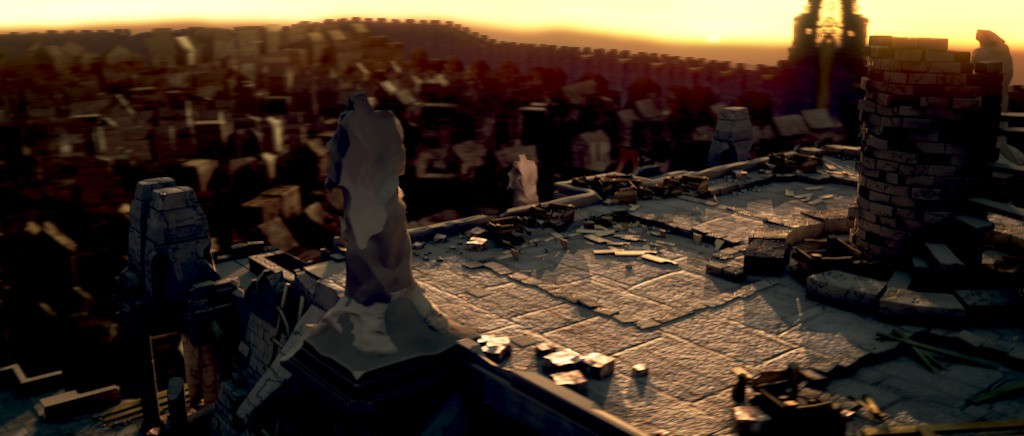}
  
  \hfill
  \includegraphics[width=0.24\textwidth]{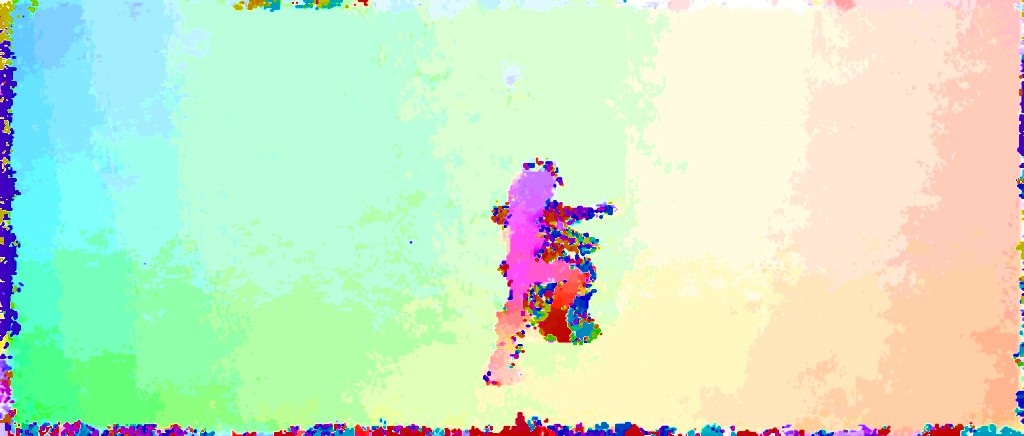}
  \includegraphics[width=0.24\textwidth]{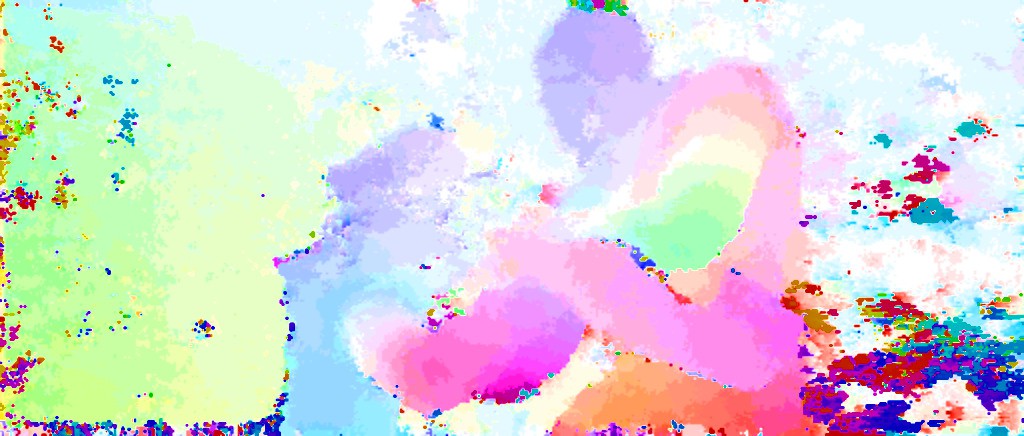}
  \includegraphics[width=0.24\textwidth]{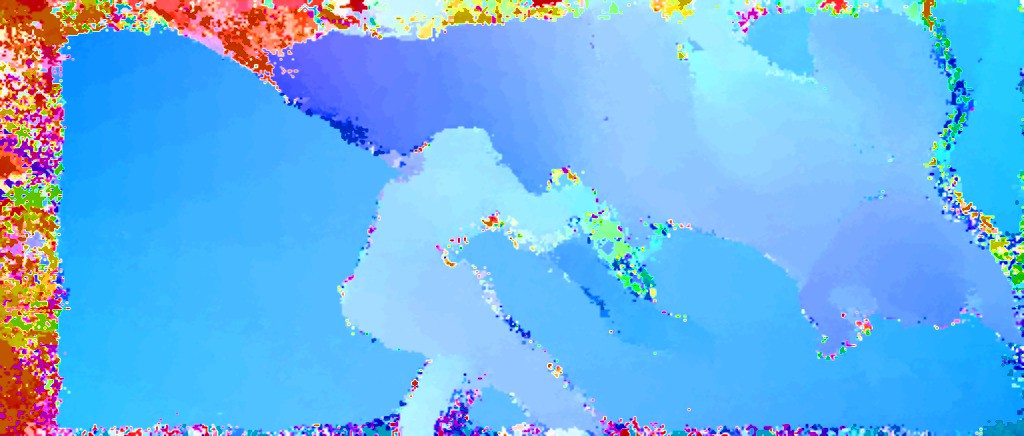}
  \includegraphics[width=0.24\textwidth]{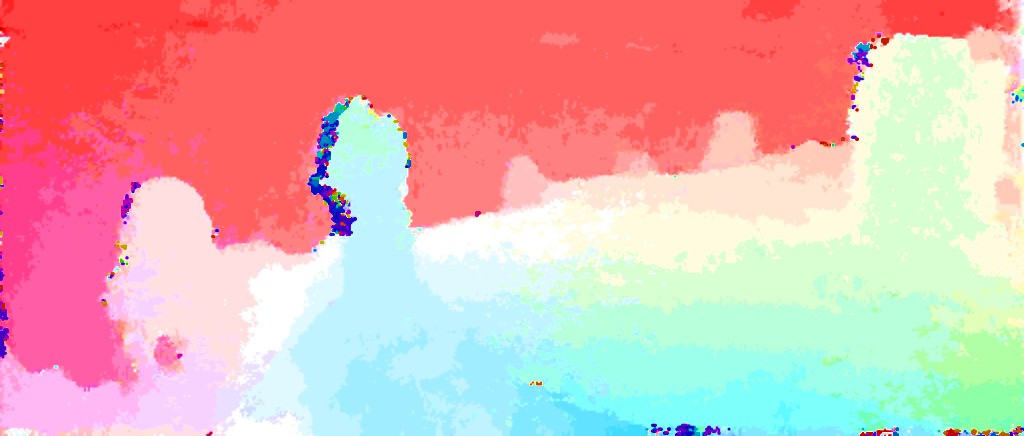}
  
  \hfill
  \includegraphics[width=0.24\textwidth]{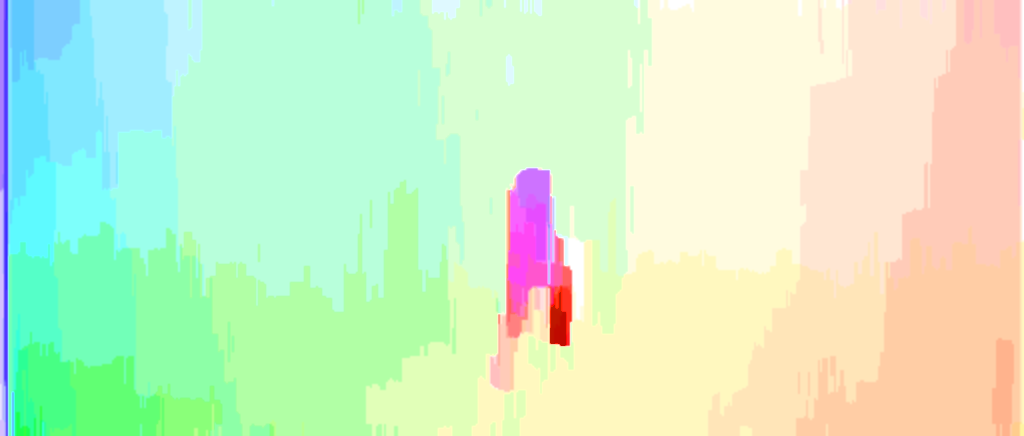}
  \includegraphics[width=0.24\textwidth]{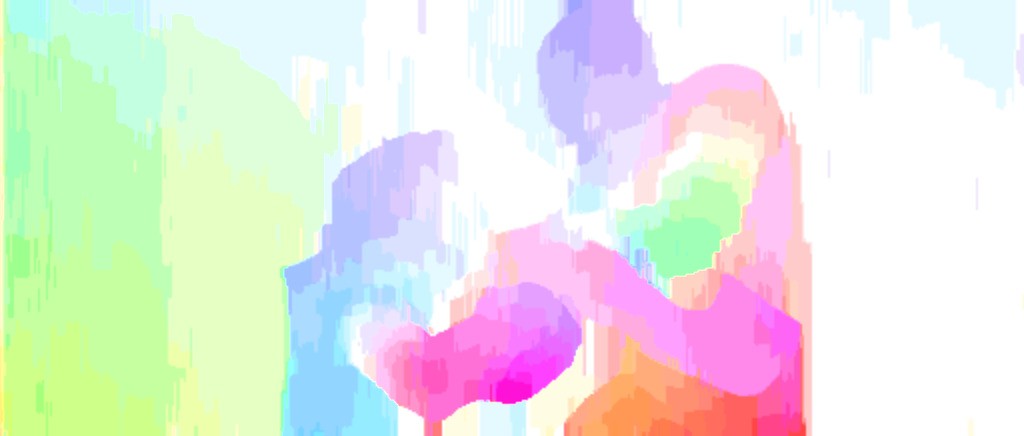}
  \includegraphics[width=0.24\textwidth]{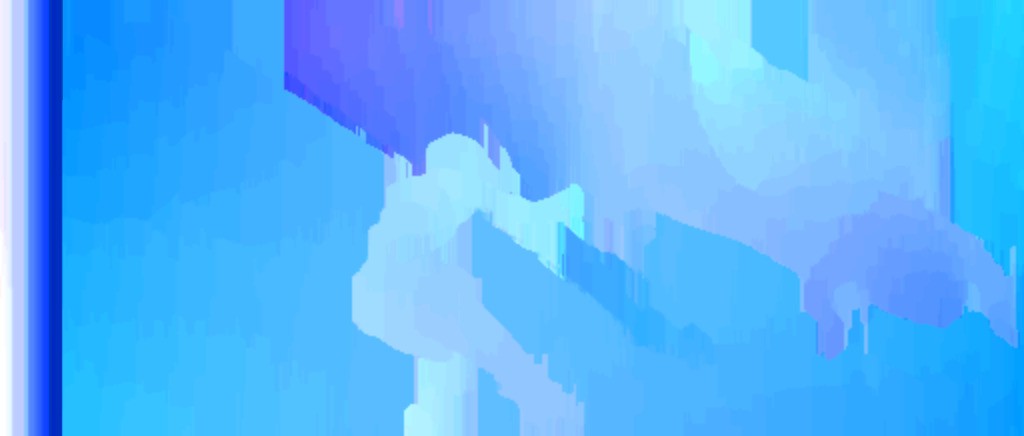}
  \includegraphics[width=0.24\textwidth]{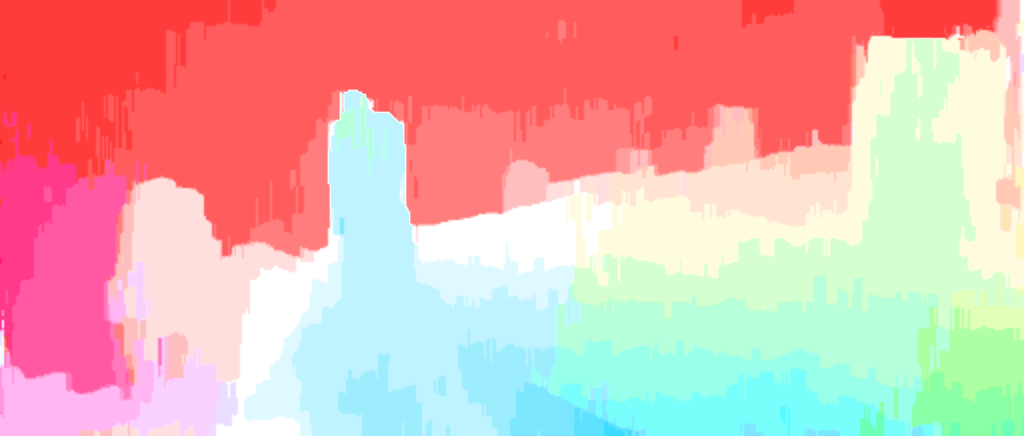}

  \hfill
  \includegraphics[width=0.24\textwidth]{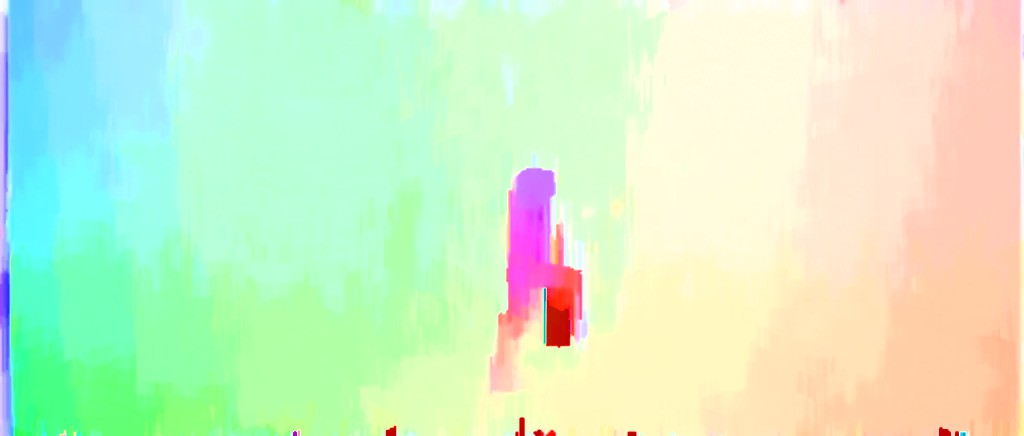}
  \includegraphics[width=0.24\textwidth]{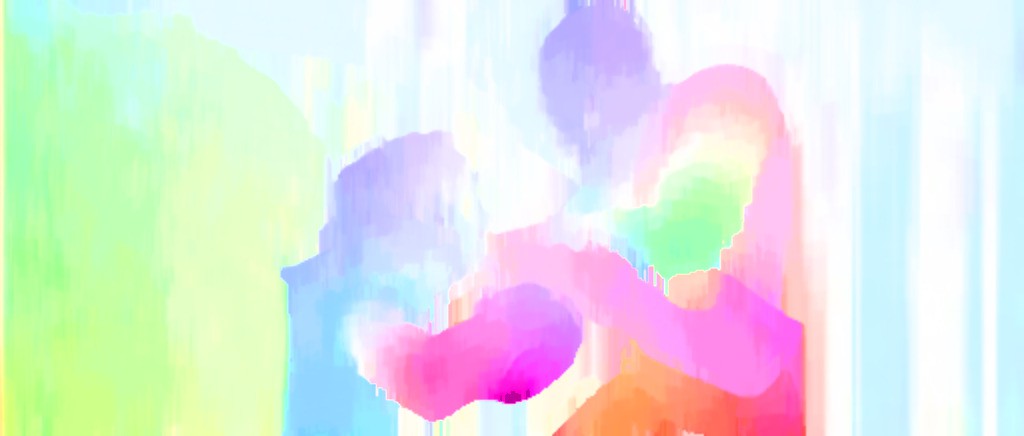}
  \includegraphics[width=0.24\textwidth]{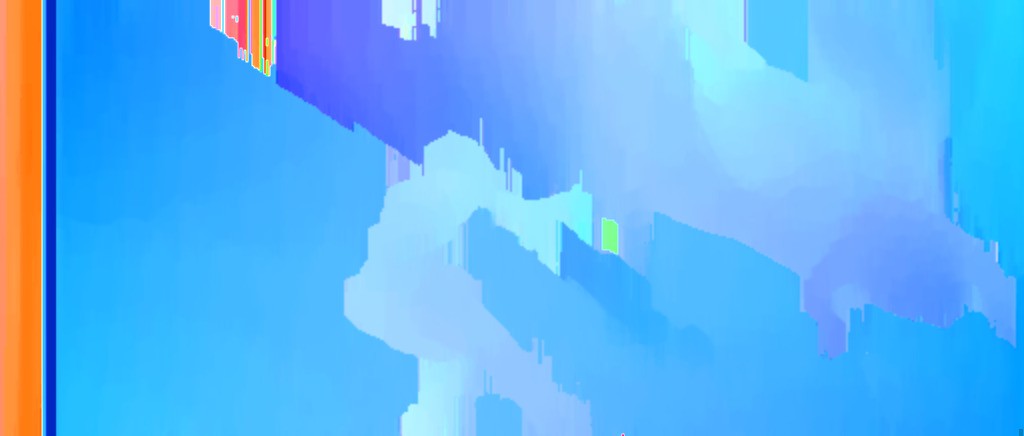}
  \includegraphics[width=0.24\textwidth]{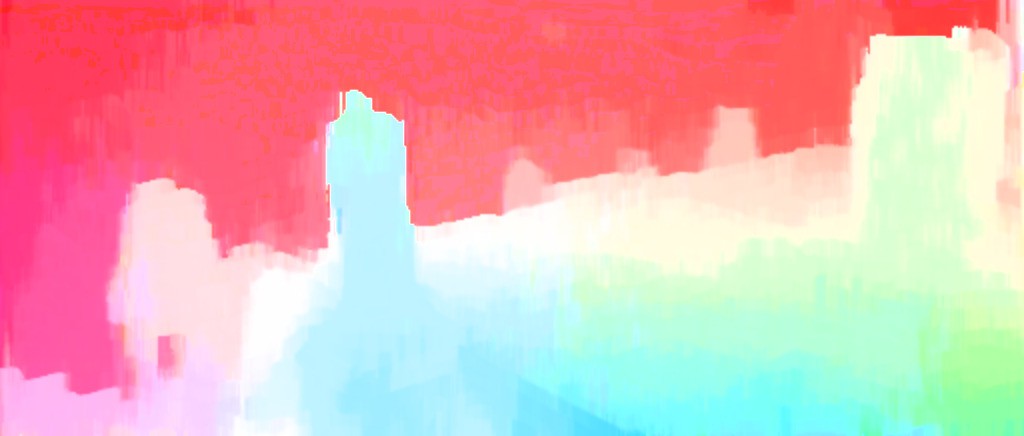}
  
  \hfill
  \includegraphics[width=0.24\textwidth]{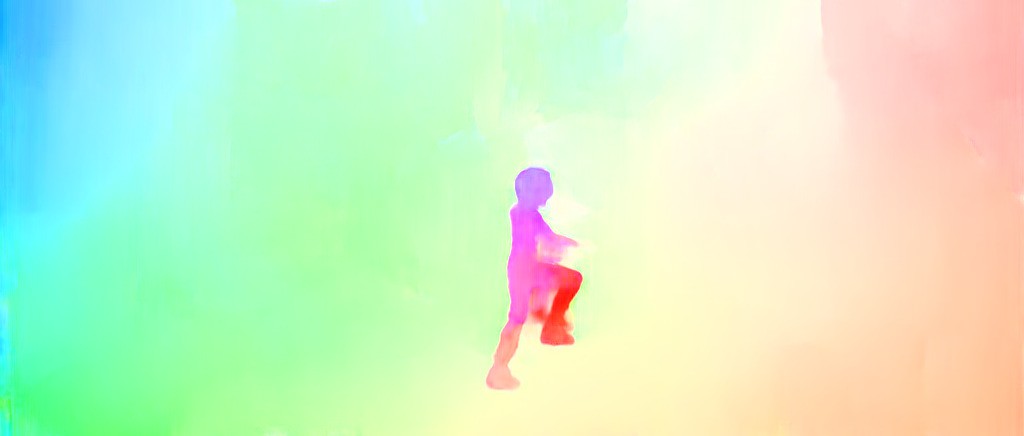}
  \includegraphics[width=0.24\textwidth]{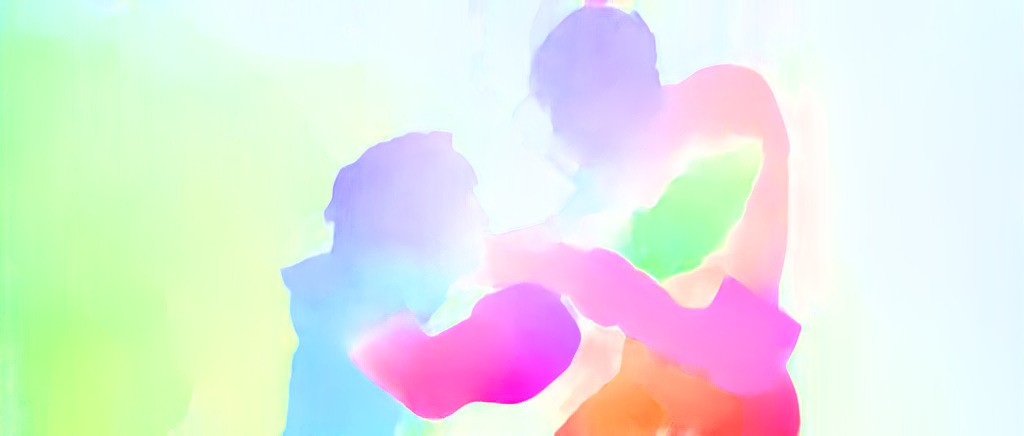}
  \includegraphics[width=0.24\textwidth]{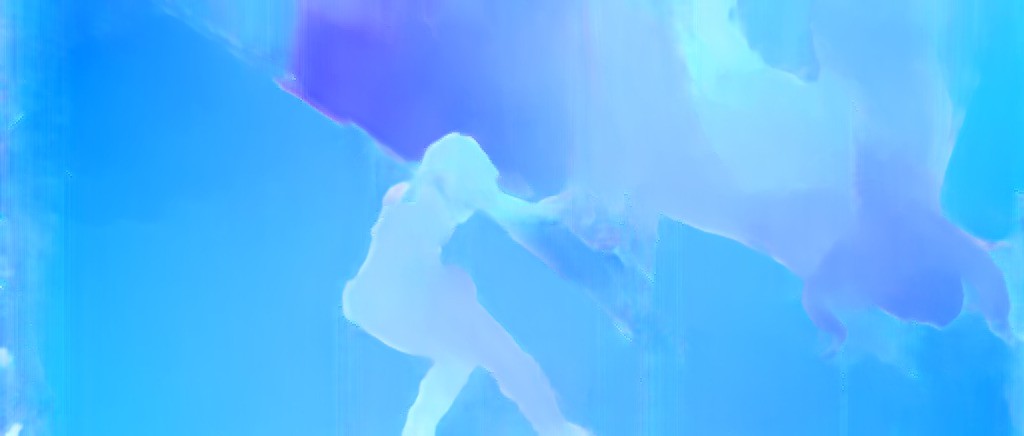}
  \includegraphics[width=0.24\textwidth]{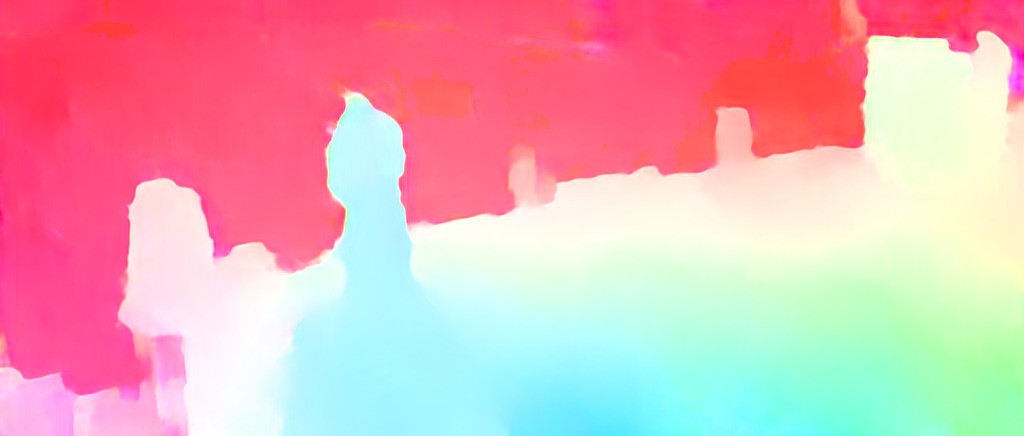}
  
  \hfill
  \includegraphics[width=0.24\textwidth]{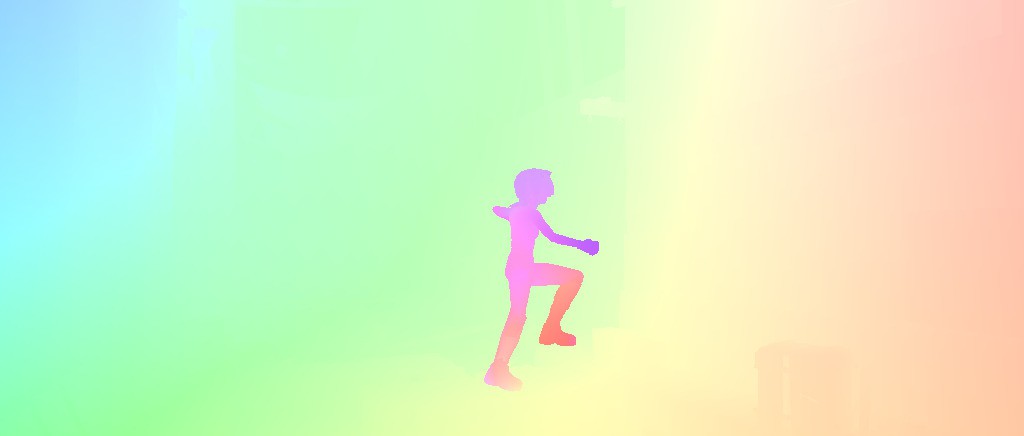}
  \includegraphics[width=0.24\textwidth]{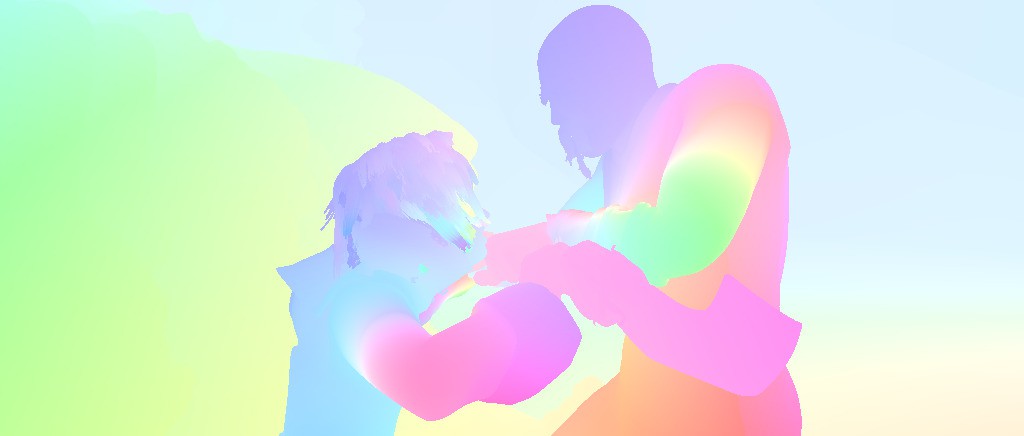}
  \includegraphics[width=0.24\textwidth]{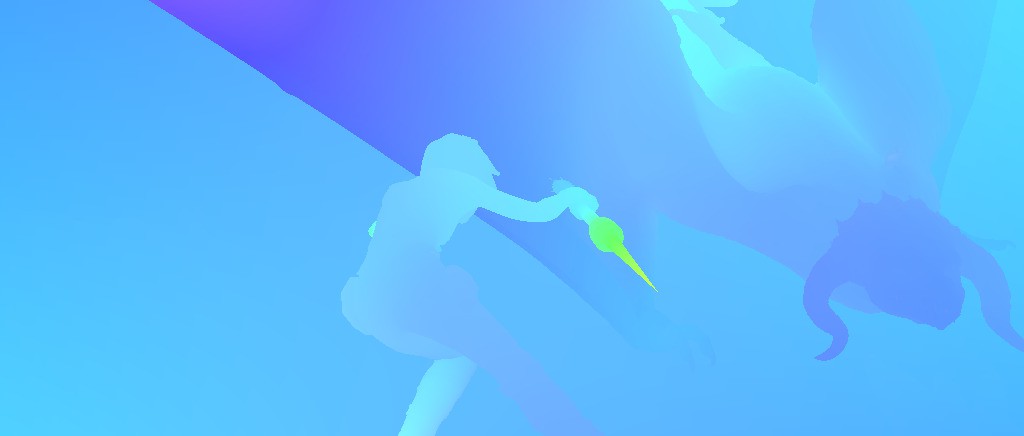}
  \includegraphics[width=0.24\textwidth]{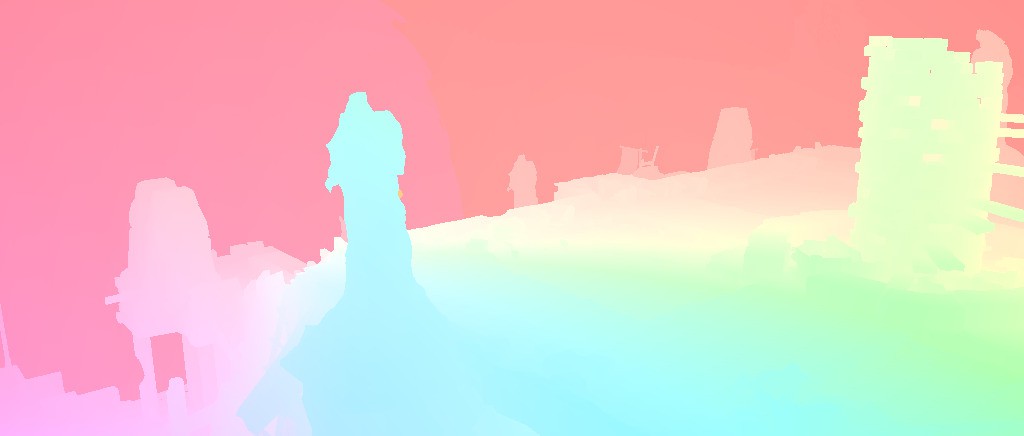}

  \caption{Qualitative ablation study for optical flow. The WTA result clearly shows occluded regions (the noisy regions), while our model is able to successfully inpaint these regions. Note that the details increase from top to bottom.}
  \label{fig:sintel:more}
\end{figure*}
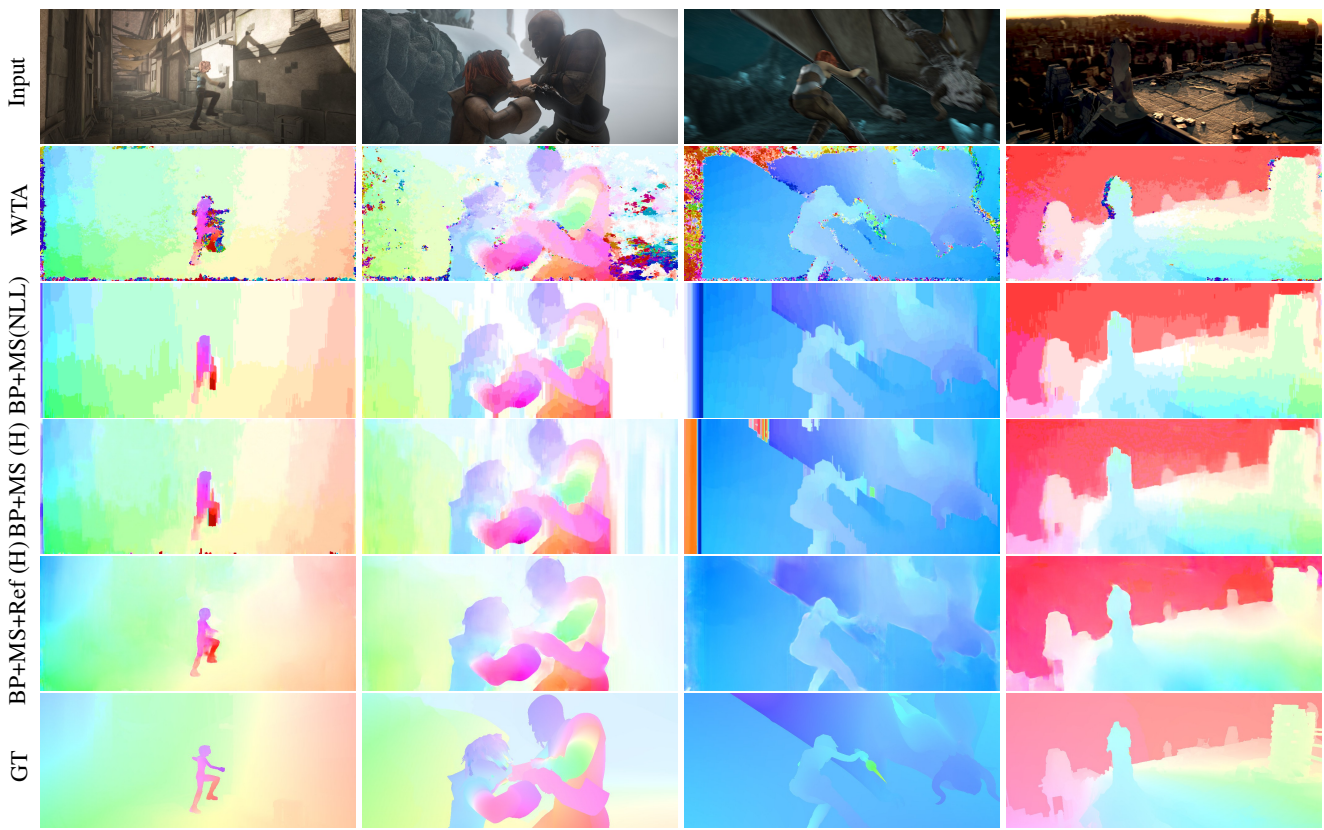
\begin{figure*}
  \centering
  \includegraphics[width=0.49\textwidth]{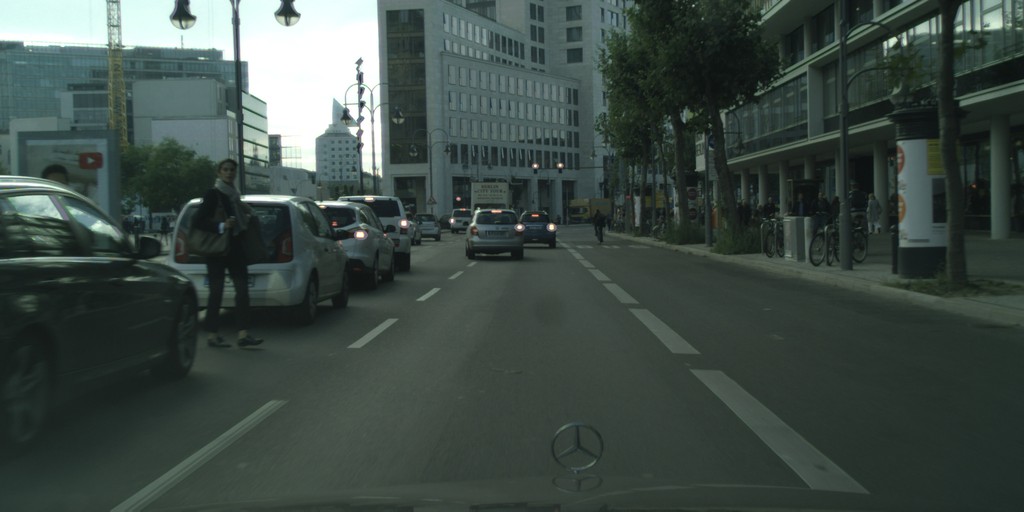}
  \includegraphics[width=0.49\textwidth]{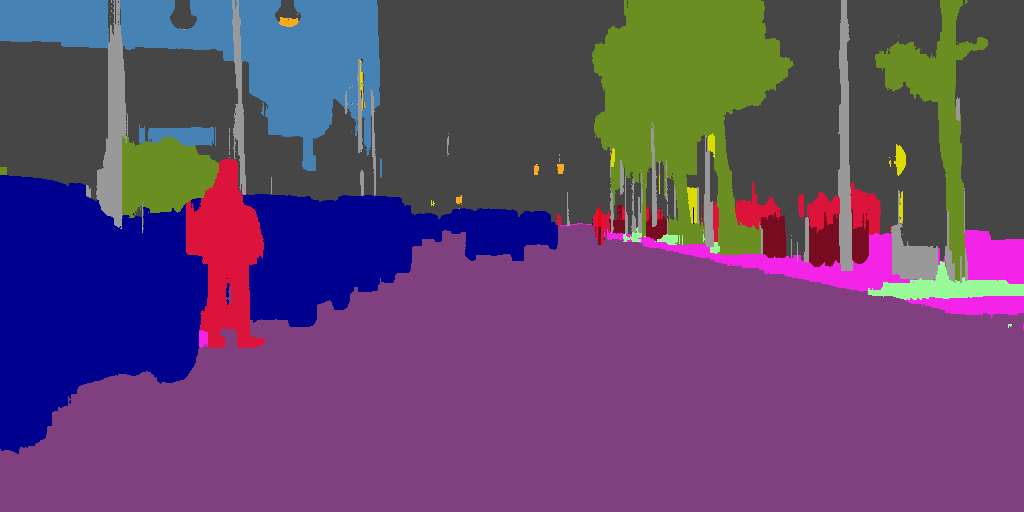} \\
  \includegraphics[width=0.49\textwidth]{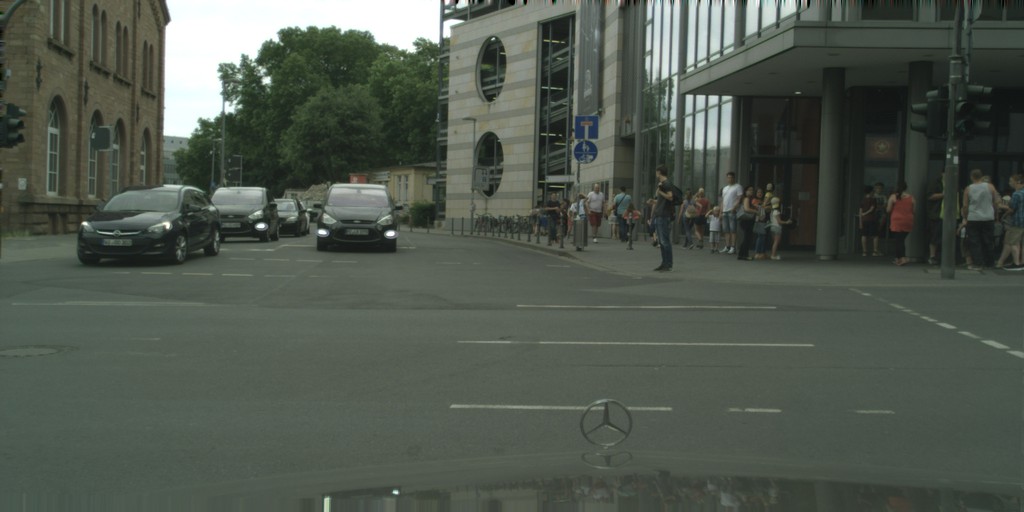}
  \includegraphics[width=0.49\textwidth]{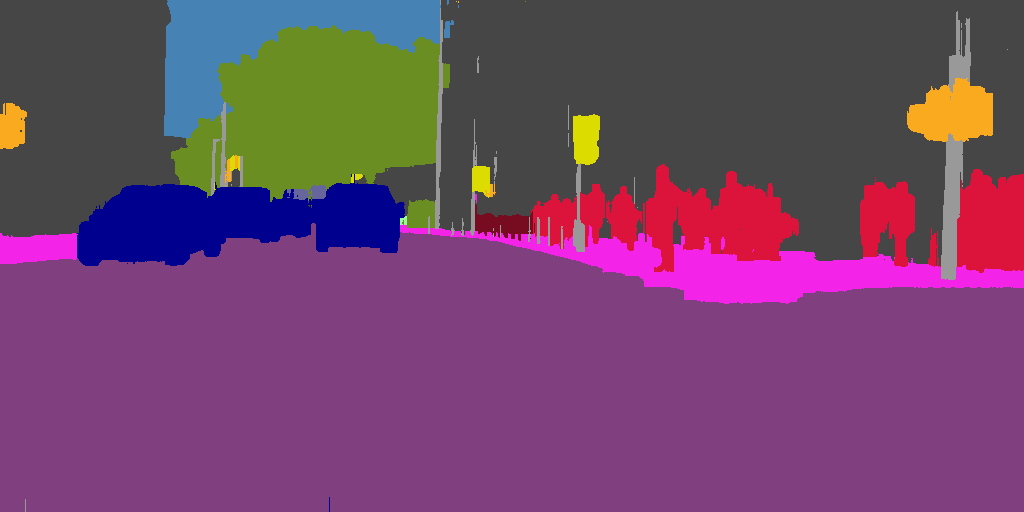}
  \caption{Qualitative results for semantic segmentation on the Cityscapes~\cite{Cordts16} test set. Our model is able to precisely capture object boundaries around \eg pedestrians and cars.}
  \label{fig:suppl:qual:test:semantic}
\end{figure*}
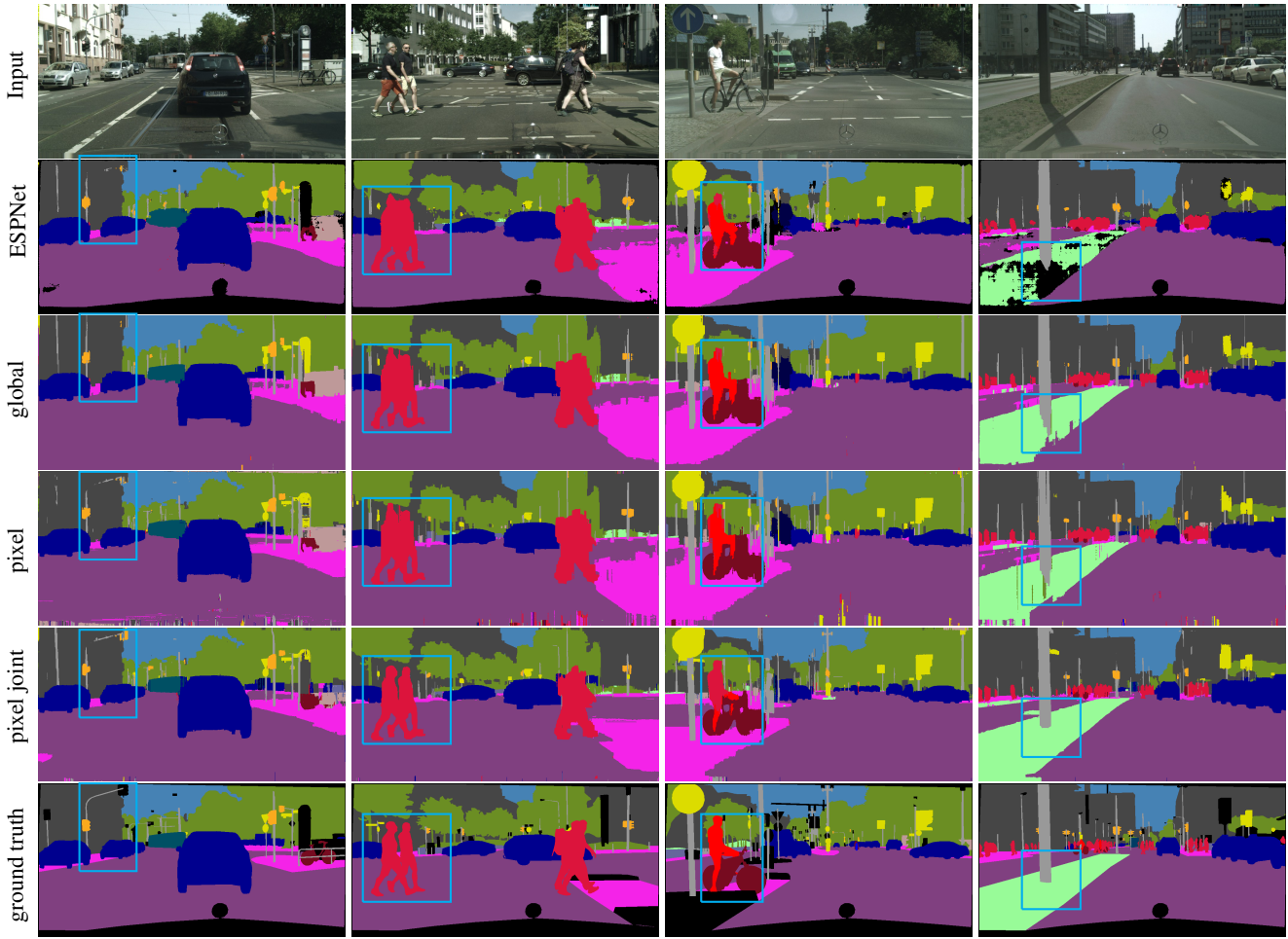
\begin{figure*}
  \begin{tikzpicture}[overlay, anchor=center]
  \small
  \node[rotate=90] at (0.15,-1.0) {Input};
  \node[rotate=90] at (0.15,-3.0) {ESPNet};
  \node[rotate=90] at (0.15,-5.3) {global};
  \node[rotate=90] at (0.15,-7.5) {pixel};
  \node[rotate=90] at (0.15,-9.5) {pixel joint};
  \node[rotate=90] at (0.15,-11.8) {ground truth};
  \end{tikzpicture}
  
  \hfill
  \includegraphics[width=0.24\textwidth]{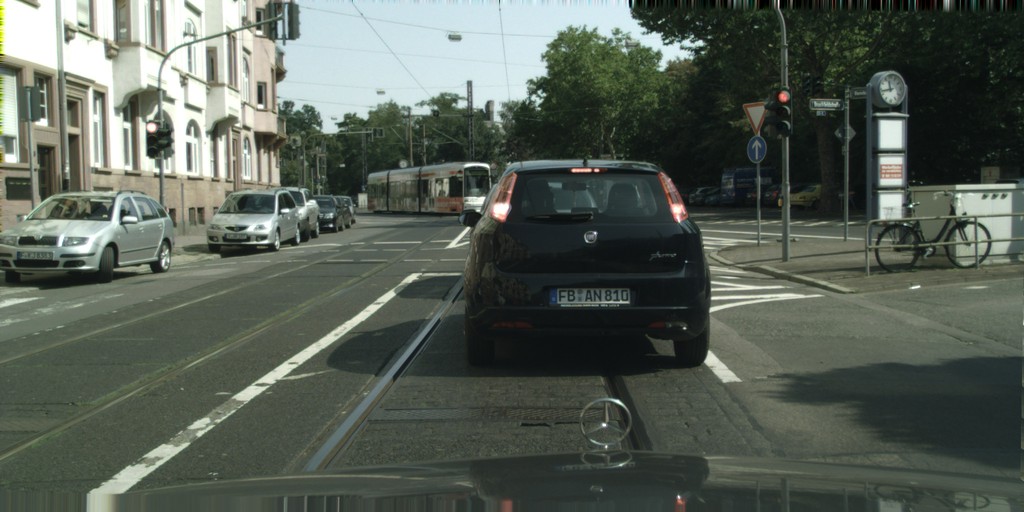}
  \includegraphics[width=0.24\textwidth]{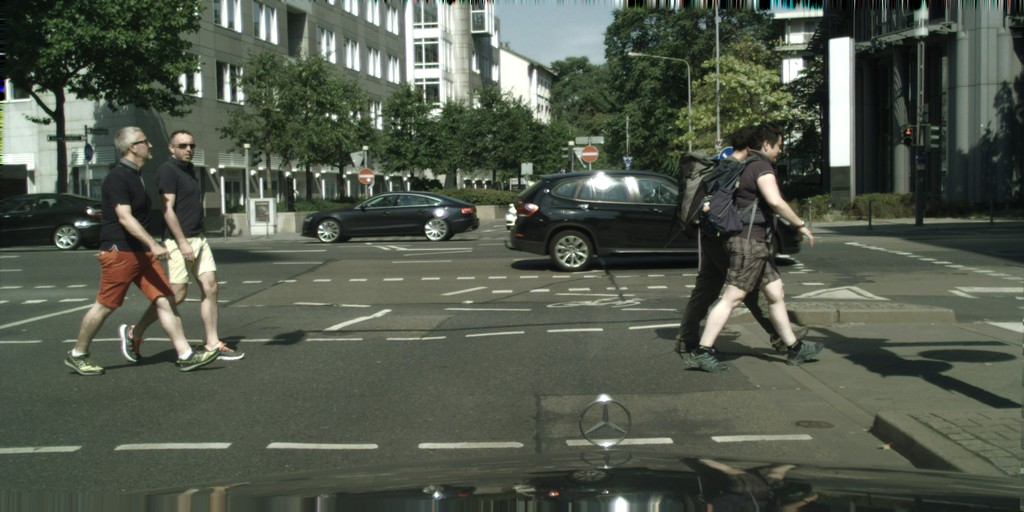}
  \includegraphics[width=0.24\textwidth]{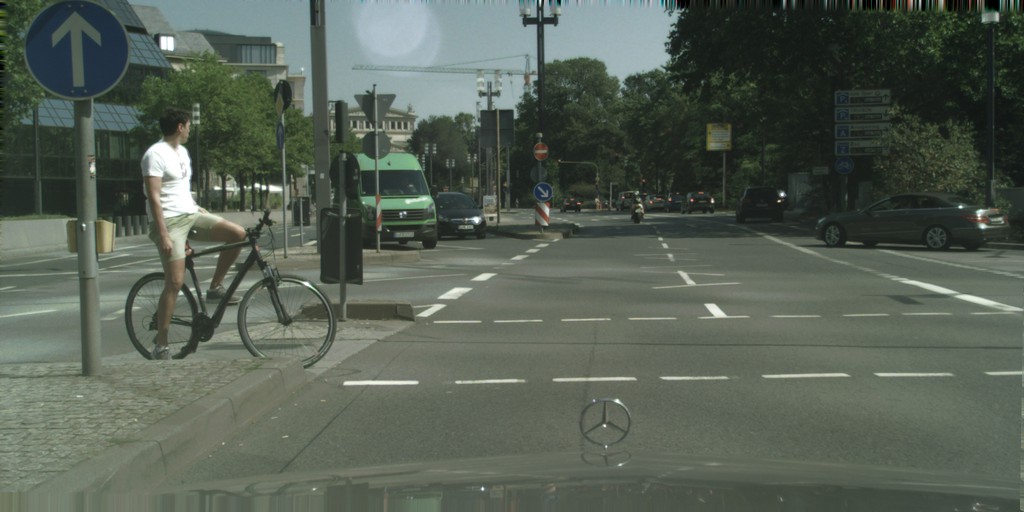}
  \includegraphics[width=0.24\textwidth]{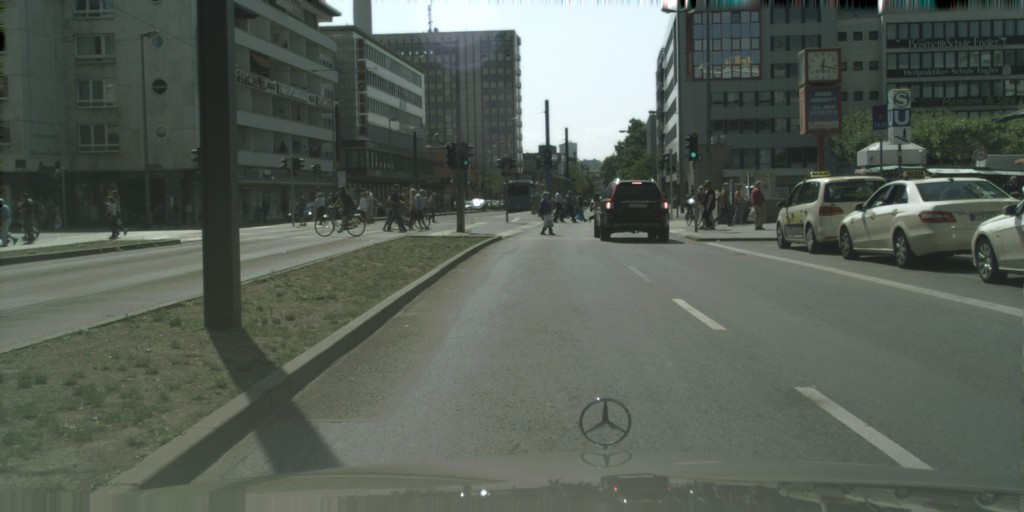}
  
  \hfill
  \includegraphics[width=0.24\textwidth]{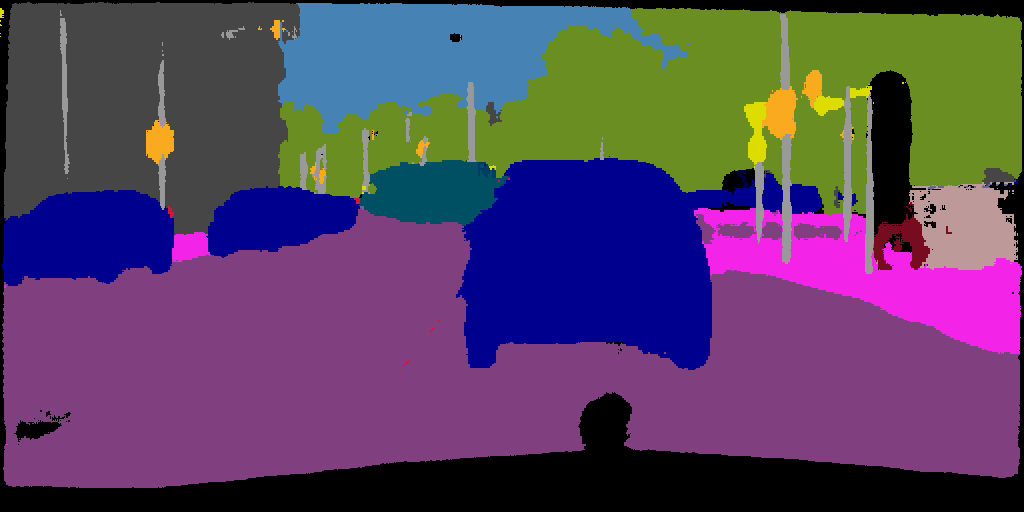}
  \includegraphics[width=0.24\textwidth]{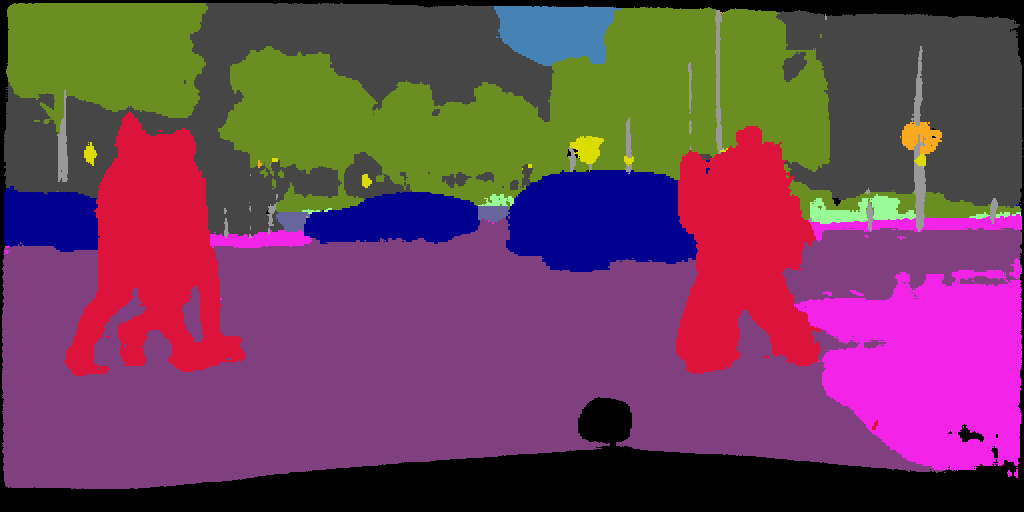}
  \includegraphics[width=0.24\textwidth]{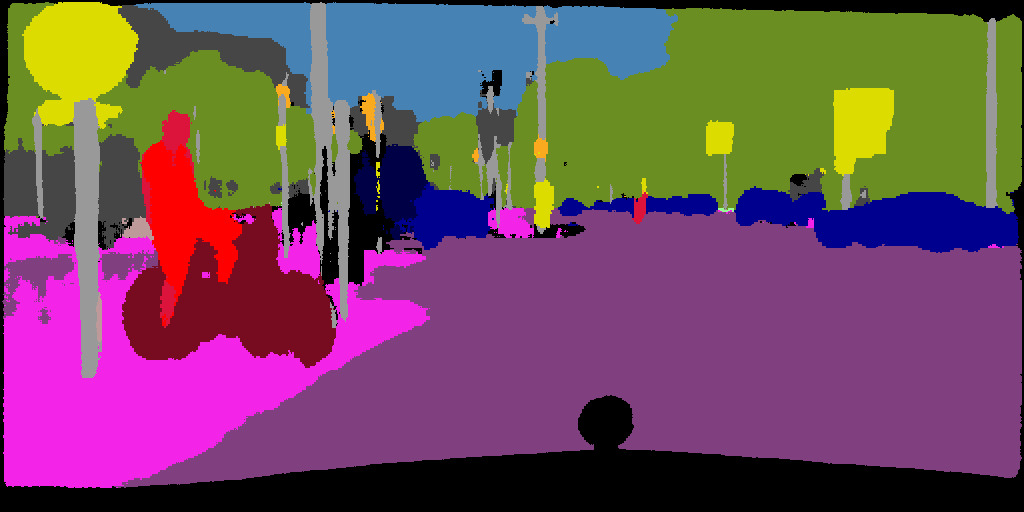}
  \includegraphics[width=0.24\textwidth]{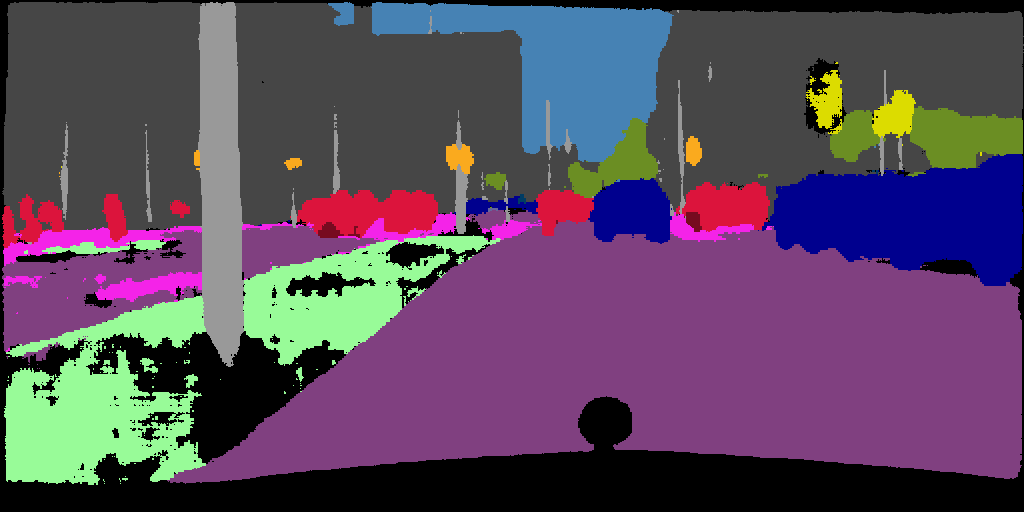}
  
  \hfill
  \includegraphics[width=0.24\textwidth]{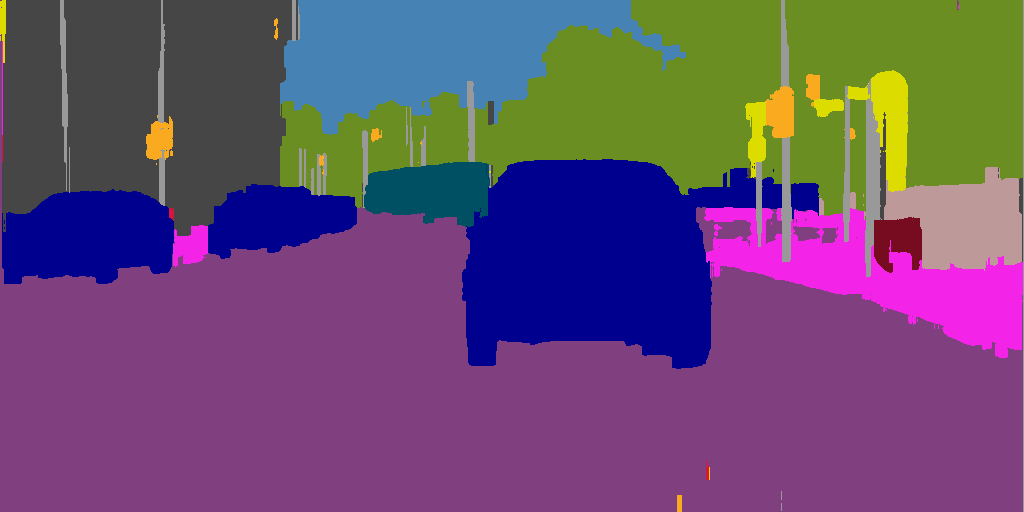}
  \includegraphics[width=0.24\textwidth]{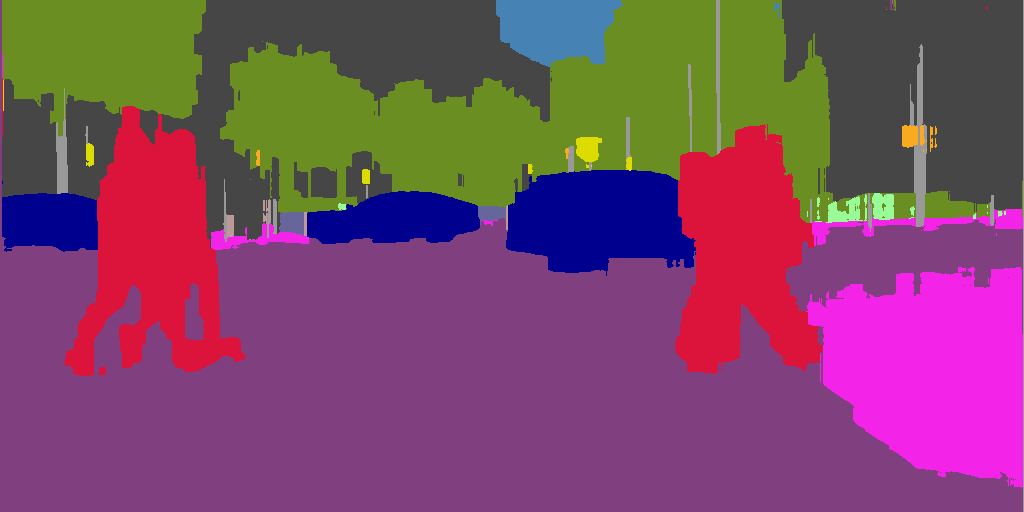}
  \includegraphics[width=0.24\textwidth]{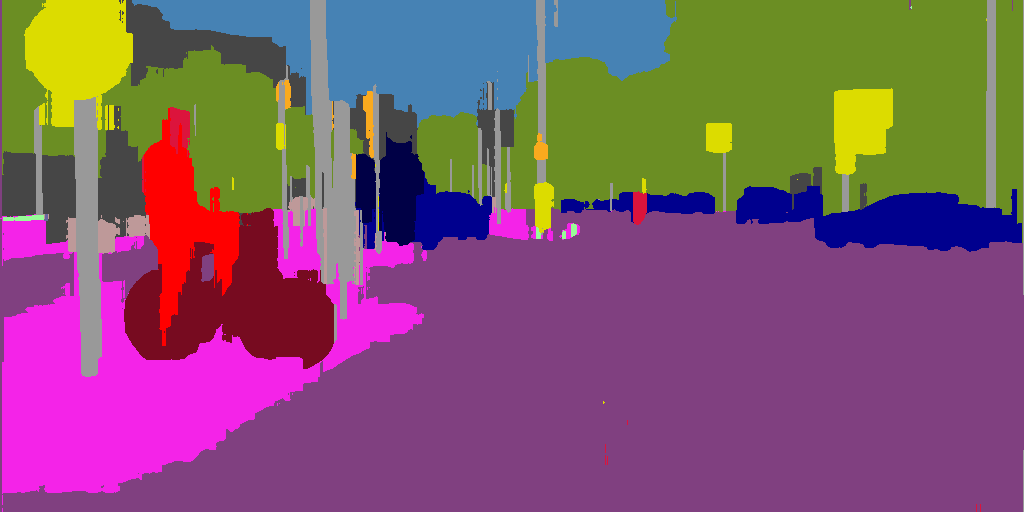}
  \includegraphics[width=0.24\textwidth]{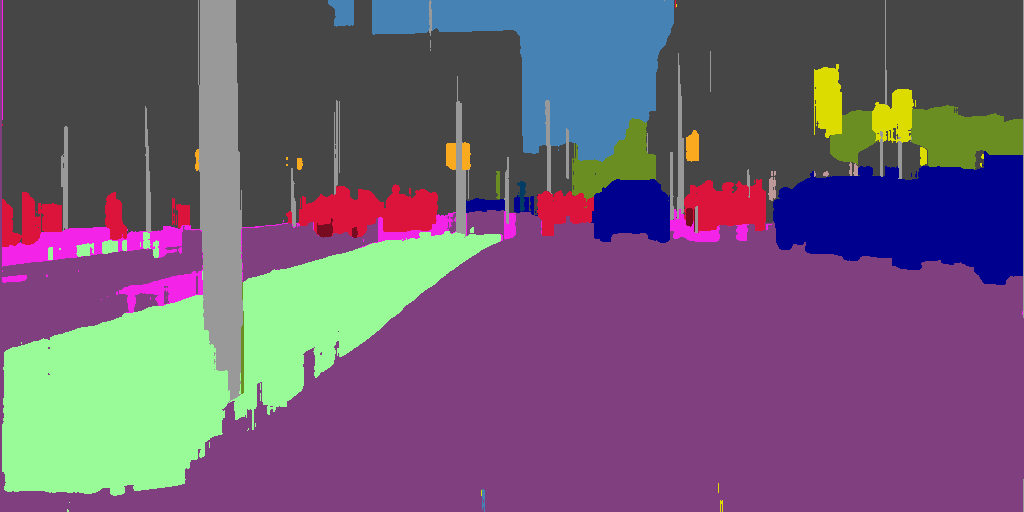}
  
  \hfill
  \includegraphics[width=0.24\textwidth]{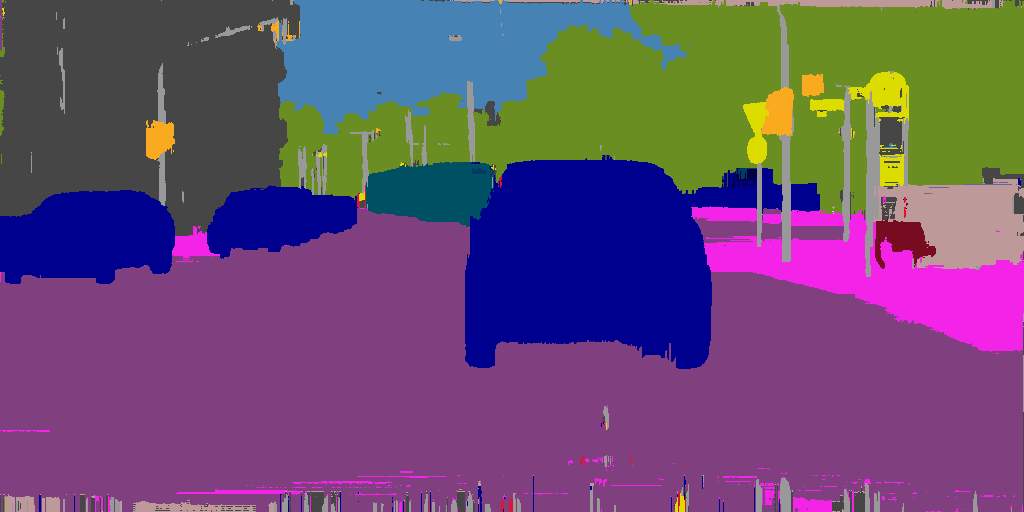}
  \includegraphics[width=0.24\textwidth]{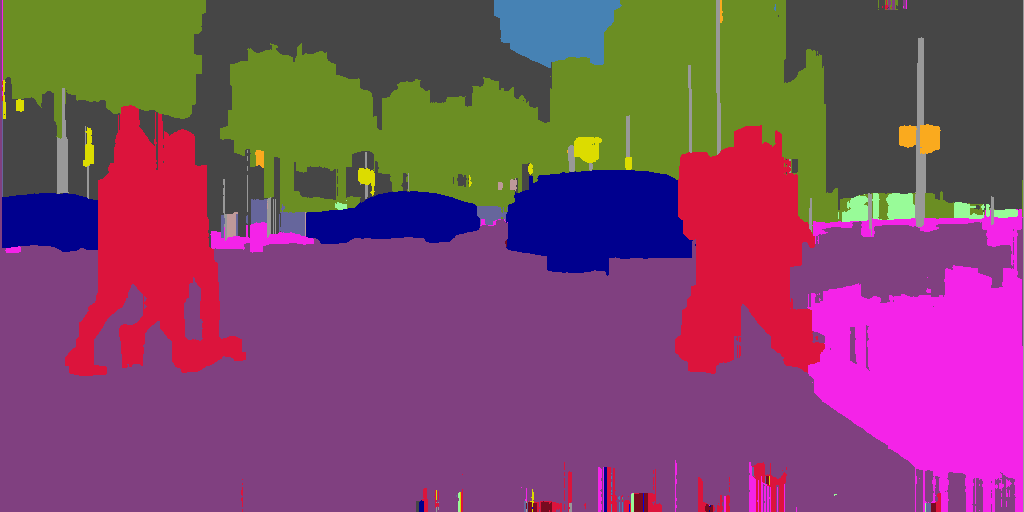}
  \includegraphics[width=0.24\textwidth]{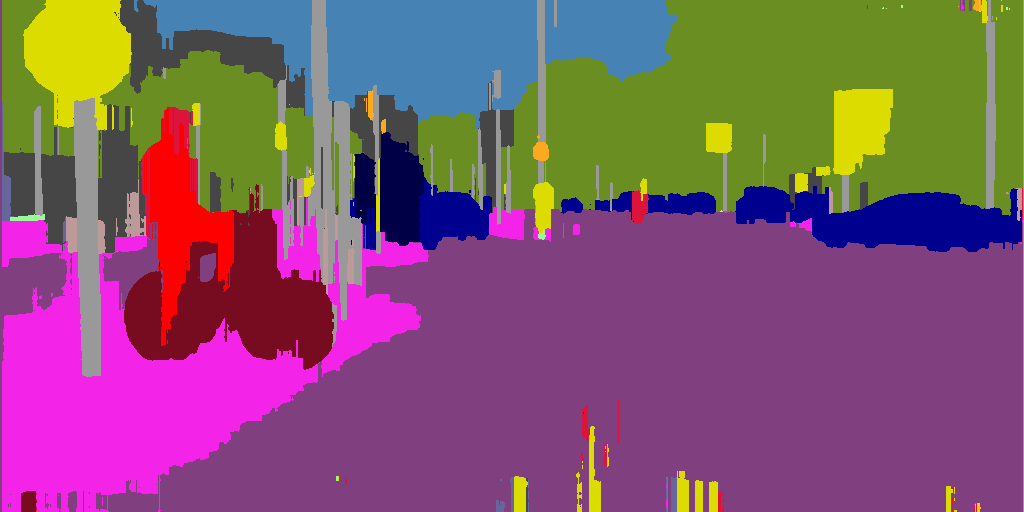}
  \includegraphics[width=0.24\textwidth]{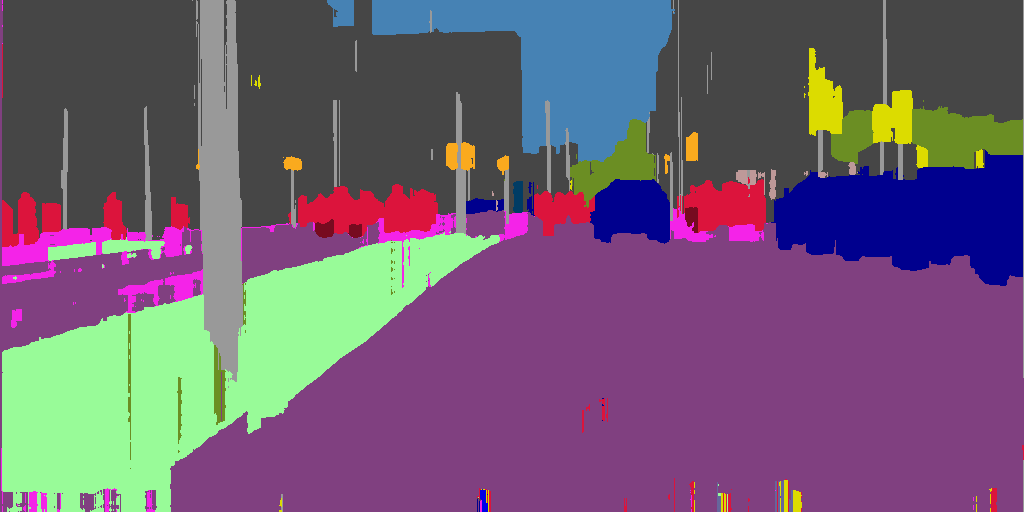}
  
  \hfill
  \includegraphics[width=0.24\textwidth]{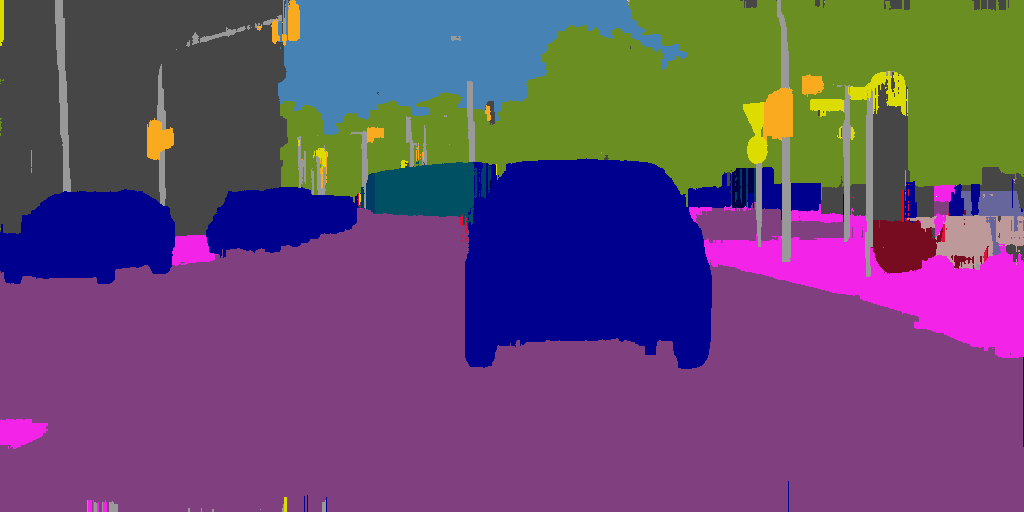} 
  \includegraphics[width=0.24\textwidth]{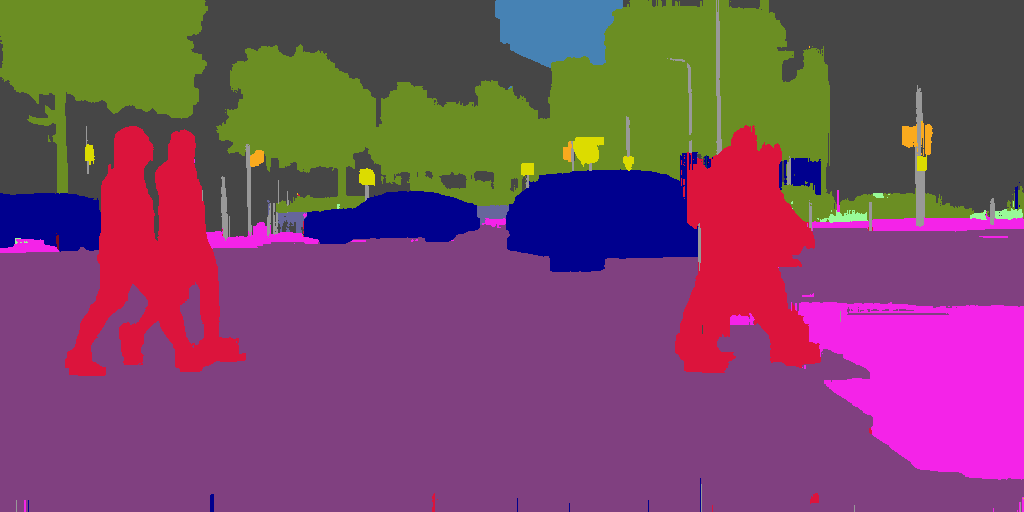} 
  \includegraphics[width=0.24\textwidth]{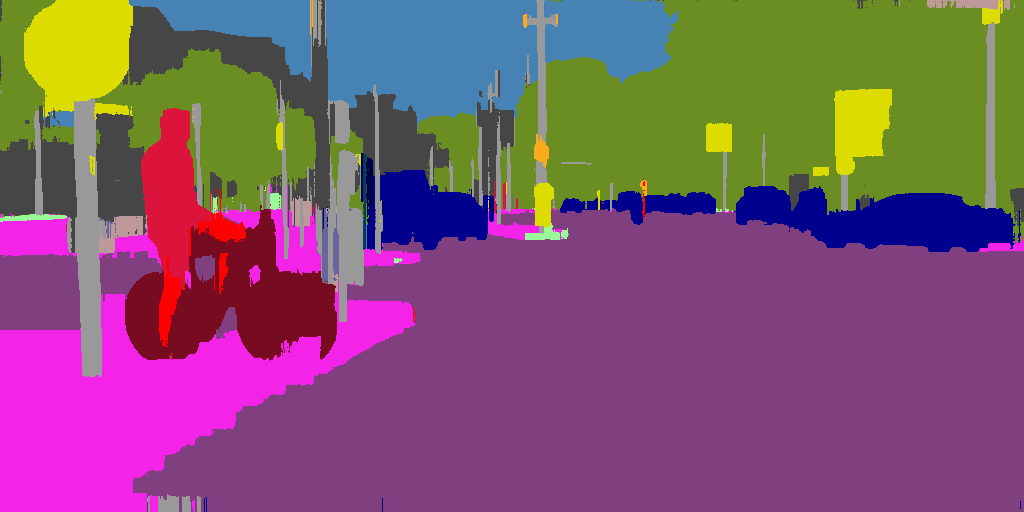}
  \includegraphics[width=0.24\textwidth]{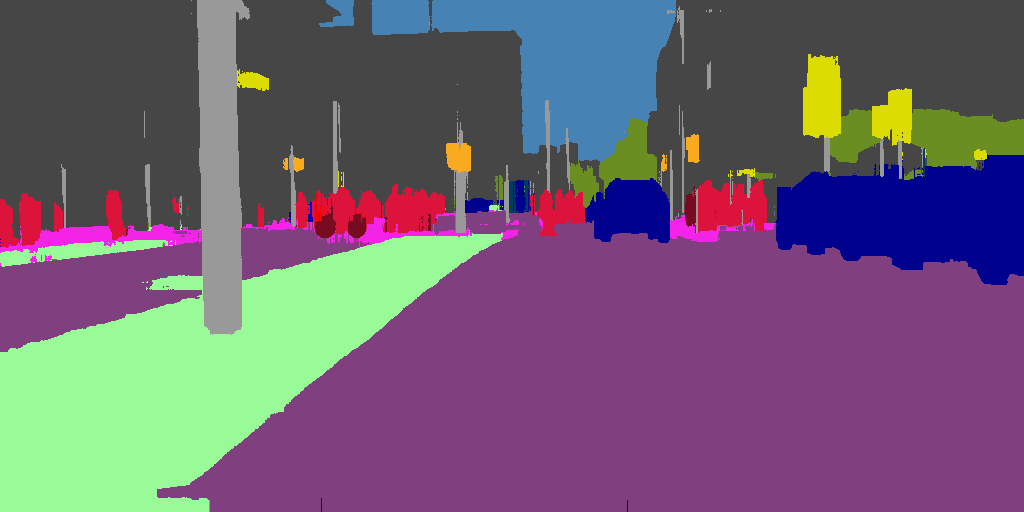}
  
  \hfill
  \includegraphics[width=0.24\textwidth]{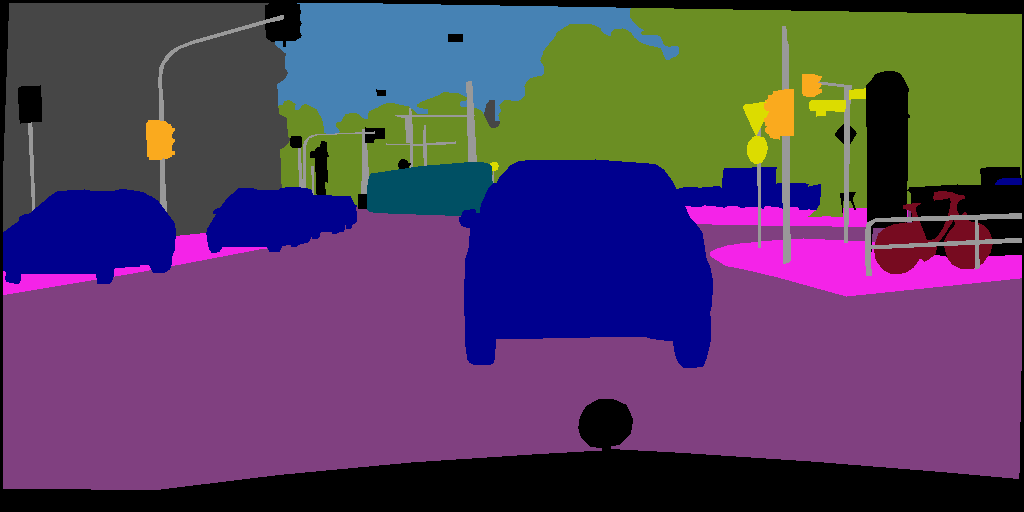}
  \includegraphics[width=0.24\textwidth]{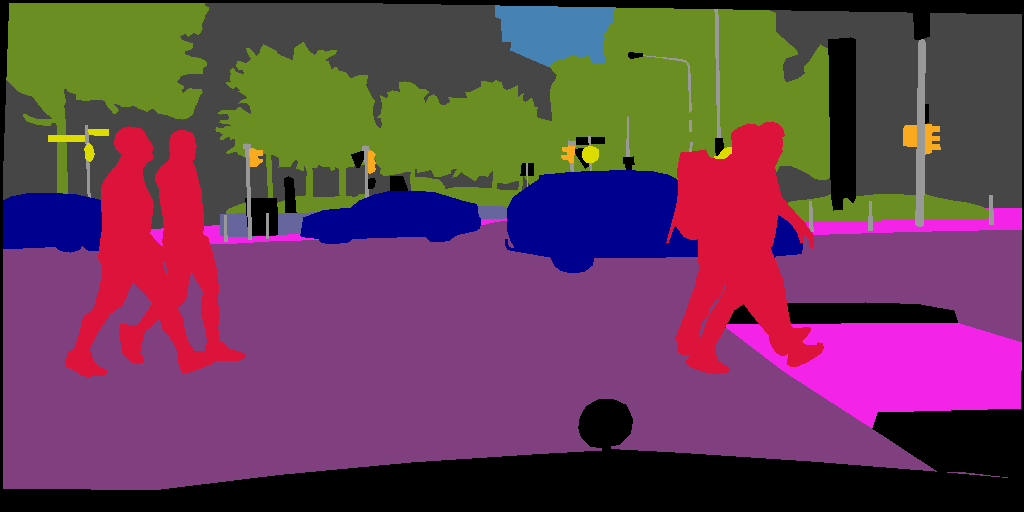}
  \includegraphics[width=0.24\textwidth]{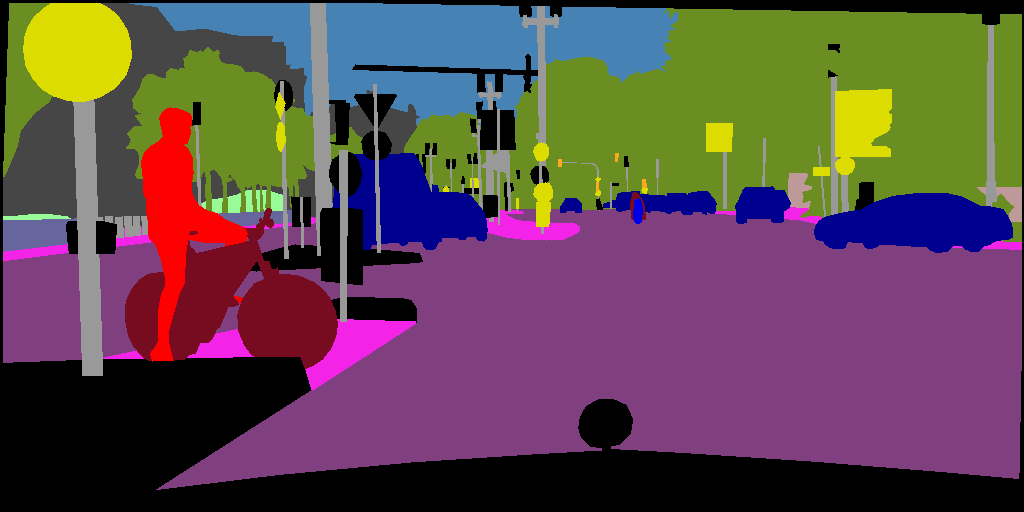}
  \includegraphics[width=0.24\textwidth]{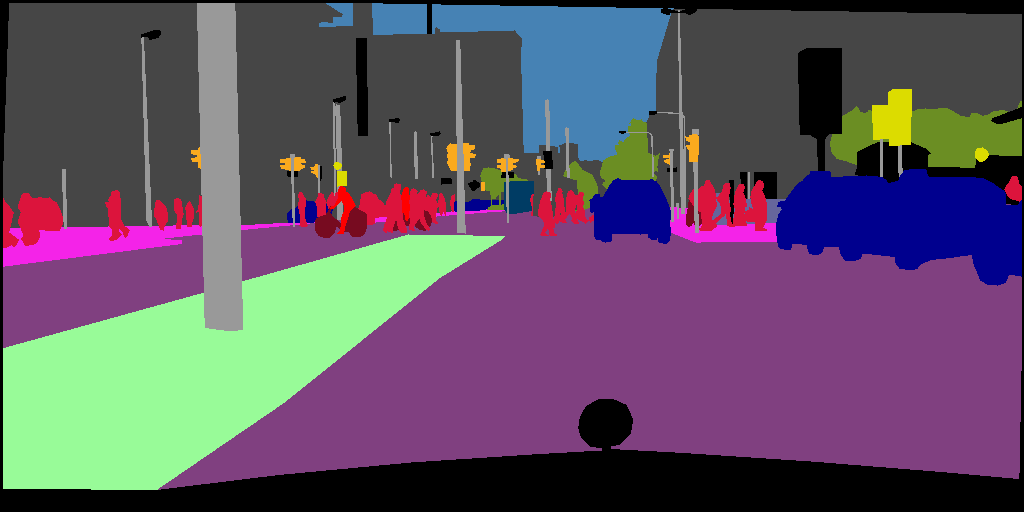}
  
  \hfill
  \begin{tikzpicture}[overlay, anchor=center]
  \draw[cyan, thick, scale=0.8, xshift=-4.5cm, yshift=1cm] (0,0) -- (0,1) -- (1,1) -- (1,0) --(0,0);
  \draw[cyan, thick, scale=0.8, xshift=-4.5cm, yshift=3.6cm] (0,0) -- (0,1) -- (1,1) -- (1,0) --(0,0);
  \draw[cyan, thick, scale=0.8, xshift=-4.5cm, yshift=6.2cm] (0,0) -- (0,1) -- (1,1) -- (1,0) --(0,0);
  \draw[cyan, thick, scale=0.8, xshift=-4.5cm, yshift=8.8cm] (0,0) -- (0,1) -- (1,1) -- (1,0) --(0,0);
  \draw[cyan, thick, scale=0.8, xshift=-4.5cm, yshift=11.4cm] (0,0) -- (0,1) -- (1,1) -- (1,0) --(0,0);
  \draw[cyan, thick, scale=0.6, xshift=-13.3cm, yshift=1.5cm] (0,0) -- (0,2) -- (1.4,2) -- (1.4,0) --(0,0);
  \draw[cyan, thick, scale=0.6, xshift=-13.3cm, yshift=5.1cm] (0,0) -- (0,2) -- (1.4,2) -- (1.4,0) --(0,0);
   \draw[cyan, thick, scale=0.6, xshift=-13.3cm, yshift=8.7cm] (0,0) -- (0,2) -- (1.4,2) -- (1.4,0) --(0,0);
  \draw[cyan, thick, scale=0.6, xshift=-13.3cm, yshift=12.3cm] (0,0) -- (0,2) -- (1.4,2) -- (1.4,0) --(0,0);
   \draw[cyan, thick, scale=0.6, xshift=-13.3cm, yshift=15.9cm] (0,0) -- (0,2) -- (1.4,2) -- (1.4,0) --(0,0);
  \draw[cyan, thick, scale=0.6, xshift=-21cm, yshift=1.5cm] (0,0) -- (0,2) -- (2,2) -- (2,0)--(0,0);
  \draw[cyan, thick, scale=0.6, xshift=-21cm, yshift=5.1cm] (0,0) -- (0,2) -- (2,2) -- (2,0)--(0,0);
  \draw[cyan, thick, scale=0.6, xshift=-21cm, yshift=8.7cm] (0,0) -- (0,2) -- (2,2) -- (2,0)--(0,0);
  \draw[cyan, thick, scale=0.6, xshift=-21cm, yshift=12.2cm] (0,0) -- (0,2) -- (2,2) -- (2,0)--(0,0);
  \draw[cyan, thick, scale=0.6, xshift=-21cm, yshift=15.8cm] (0,0) -- (0,2) -- (2,2) -- (2,0)--(0,0);
  \draw[cyan, thick, scale=0.6, xshift=-27.45cm, yshift=2.2cm] (0,0) -- (0,2) -- (1.3,2) -- (1.3,0)--(0,0);
  \draw[cyan, thick, scale=0.6, xshift=-27.45cm, yshift=5.7cm] (0,0) -- (0,2) -- (1.3,2) -- (1.3,0)--(0,0);
  \draw[cyan, thick, scale=0.6, xshift=-27.45cm, yshift=9.3cm] (0,0) -- (0,2) -- (1.3,2) -- (1.3,0)--(0,0);
  \draw[cyan, thick, scale=0.6, xshift=-27.45cm, yshift=12.9cm] (0,0) -- (0,2) -- (1.3,2) -- (1.3,0)--(0,0);
   \draw[cyan, thick, scale=0.6, xshift=-27.45cm, yshift=16.5cm] (0,0) -- (0,2) -- (1.3,2) -- (1.3,0)--(0,0);
  \end{tikzpicture}
  \caption{Visual ablation study for semantic segmentation on the Cityscapes~\cite{Cordts16} validation set. The results in the first column show that the BP-Layer can recover fine details such as the thin structures of the traffic light. In the second column one can observe that the legs and heads of the pedestrians are recovered and do not appear as a single blob-like structure. This can also be seen when looking at the bike in the third column. The fourt column shows that the BP-Layer can regularize over inconsistencies in the initial estimation from ESPNet~\cite{Mehta18} as seen on the sidewalk. }
  \label{fig:vis:ablation:semantic}
\end{figure*}
\begin{table*}
  \centering
  \small
  \setlength\tabcolsep{4pt}
  \begin{tabular}{lccccccccc}
    \toprule
    \textbf{Method} & \textbf{avg} & \textbf{flat} & \textbf{nature} & \textbf{object} & \textbf{sky} & \textbf{construction} & \textbf{human} & \textbf{vehicle} \\
     \hline
     ESPNet~\cite{Mehta18} & 82.18 & 95.49 & 89.46 & 52.94 & 92.47 & 86.67 & 69.76 & 88.45  \\
     \shortstack{LBPSS pixel-wise joint} & \textbf{84.31} & \textbf{97.90} & \textbf{90.01} & \textbf{58.89} & \textbf{93.10} & \textbf{88.08} & \textbf{72.79} & \textbf{89.43}   \\ %
  \end{tabular} 
  \caption{Benchmark results for categories on the Cityscapes~\cite{Cordts16} test set} %
  \label{tab:ablation:ss:categories}
  \vspace{1em}
  \centering
  \setlength\tabcolsep{3pt}
  \resizebox{\textwidth}{!}{
  \begin{tabular}{lcccccccccccccccccccccc}
    \toprule
    \textbf{Method} & \textbf{avg} & road & side. & build. & wall & fen. & pole & tr. light & tr. sign & veg. & terr. & sky & person & rider & car & truck & bus & train & motorc. & bic.  \\
     \hline
     ESPNet~\cite{Mehta18} & 60.34 & 95.68 & 73.29 & 86.60 & \textbf{32.79} & 36.43 & 47.06 & 46.92 & 55.41 & 89.83 & \textbf{65.96} & 92.47 & 68.48 & \textbf{45.84} & 89.90 & \textbf{40.00} & 47.73 & \textbf{40.70} & \textbf{36.40} & 54.89   \\
     \shortstack{LBPSS pw joint} & \textbf{61.00} & \textbf{97.00} & \textbf{76.88} & \textbf{87.38} & 31.29 & \textbf{37.99} & \textbf{53.60} & \textbf{53.84} & \textbf{60.85} & \textbf{90.41} & 65.85 & \textbf{93.10} & \textbf{70.34} & 43.27 & \textbf{90.93} & 31.59 & \textbf{50.32} & 33.93 & 31.77 & \textbf{58.67}   \\
     \hline
  \end{tabular} }
  \caption{Benchmark results with respect to the mIOU metric for each class on the Cityscapes~\cite{Cordts16} test set.} 
  \label{tab:ablation:ss:classes}
\end{table*}
\begin{figure*}[t]
  \centering
  \includegraphics[width=0.6\textwidth]{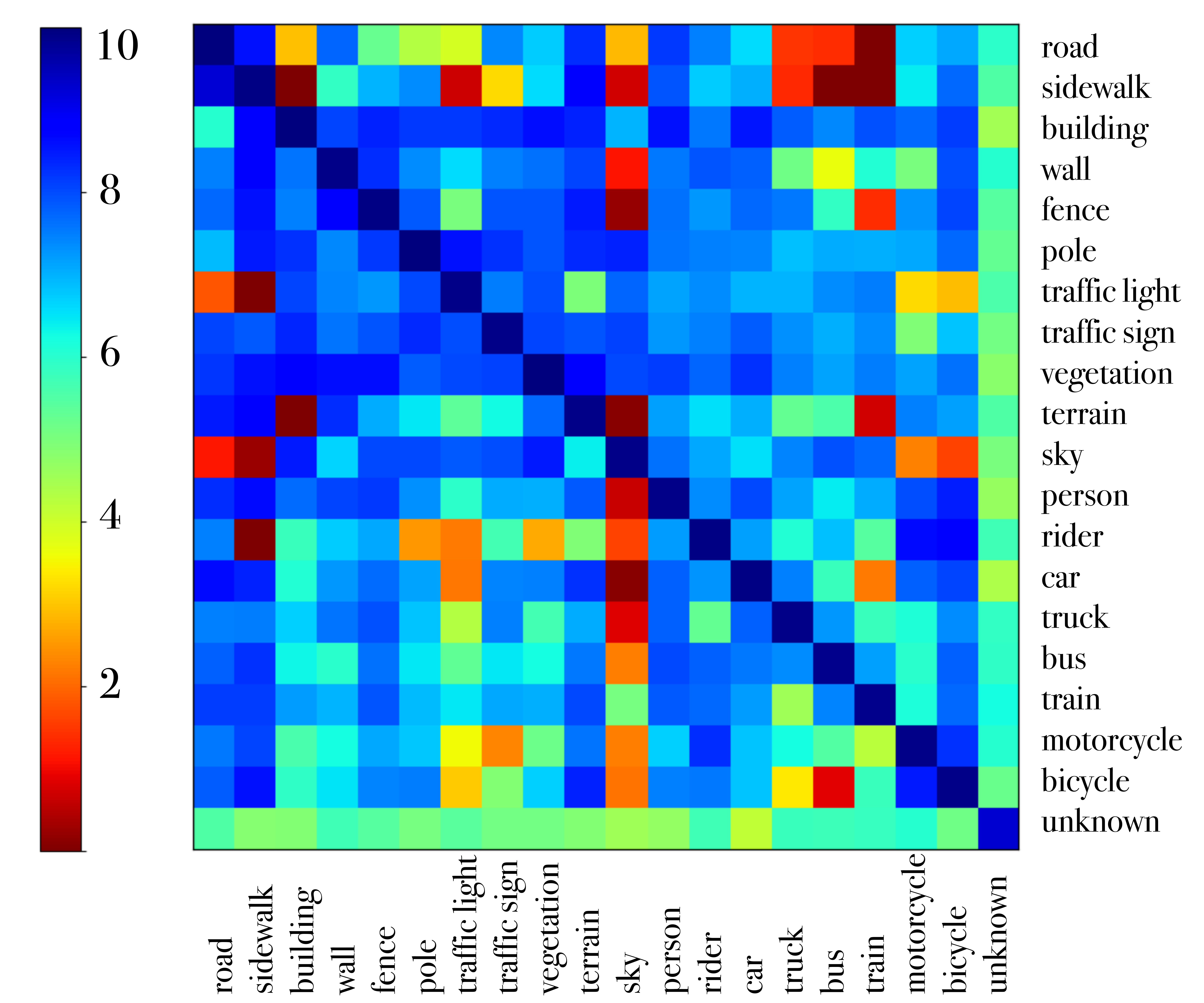}
  \caption{Vertical transition score matrix for all classes, where the upper triangular matrix encodes upwards transitions and the lower triangular matrix encodes downwards transitions.}
  \label{fig:suppl:transition:semantic}
\end{figure*}

\subsection{Semantic Segmentation}
We show here additional evaluation metrics provided by the Cityscapes benchmark.
In Table~\ref{tab:ablation:ss:categories}, we show the category mIOU score for each invidual category. 
It can be observed, that the BP-Layer improves this metric for every category and thus the average score for all categories is also improved. 
The BP-layer also improves the average class mIOU, as seen in Table~\ref{tab:ablation:ss:classes}. For this metric, the BP-layer improves the results for most classes. 
However, the mIOU is slightly decreased for the classes truck, train and motorcycle. This is due to the fact that a confusion between these classes in the result from ESPNet~\cite{Mehta18} can be propagated by the BP-Layer leading to larger patches of incorrect semantic labels. Figure~\ref{fig:vis:ablation:semantic} shows a visual ablation study of the different models for semantic segmentation. It can be seen that all of the models utilizing the BP-Layer are able to regularize over inconsistencies in the original result from ESPNet~\cite{Mehta18}. Furthermore, the pixel wise models are able to better preserve fine structures like traffic lights. If we use the BP-Layer without jointly training the ESPNet, we get some line artifacts in the global and pixel results. 
These artefacts are easily removed by jointly training both networks as seen in the pixel joint result.

In Figure~\ref{fig:suppl:qual:test:semantic}, we show qualitative results from the LBPSS pixel joint model on the test set of Cityscapes~\cite{Cordts16}. 
It can be seen that the detail on the boundaries of the segmentation masks for scene elements such as cars and pedestrians is preserved, as transition scores are predicted from the input image. 
We can also show the full vertical transition score matrix for all classes, which we do in Figure~\ref{fig:suppl:transition:semantic}. As described in the paper, the matrix is not symmetric which allows for different scores when transitioning upwards and downwards. 
If we investigate this matrix in more detail, we are actually able to interpret the learned results.
An interesting observation can \eg be seen when looking at the column for the sky class. 
It encodes that downward label transitions from car, truck or train to sky are very expensive and upwards transitions from \eg. car to sky are comparably cheap. 
This is very intuitive and encodes that the sky is always above the car and not below. 
Another example is that traffic lights and vegetation are often surrounded by sky and thus these  scores are higher. 
Also the scores for the unknown class very intuitive. The very similar scores to all other classes can be interpreted as a uniform distribution. This makes totally sense, because the class ``unknown'' has interactions with all other classes.

\end{document}